\documentclass{article}

\usepackage[final]{neurips_2025}

\usepackage[utf8]{inputenc} 
\usepackage[T1]{fontenc}    
\usepackage{hyperref}       
\usepackage{adjustbox}
\usepackage{url}            
\usepackage{booktabs}       
\usepackage{amsfonts}       
\usepackage{nicefrac}       
\usepackage{microtype}      
\usepackage{xcolor}         
\usepackage{amsmath}
\usepackage{amssymb}
\usepackage{subcaption}
\usepackage{siunitx}
\usepackage{multirow}
\usepackage{wrapfig}
\usepackage{array}
\usepackage{amsthm}
\usepackage{enumitem}
\usepackage{amsthm}
\usepackage{booktabs}   
\usepackage{makecell}   
\newtheorem{remark}{Remark}

\usepackage[linesnumbered,ruled,vlined]{algorithm2e}
\usepackage[ruled,vlined]{algorithm2e}
\SetKwProg{Procedure}{Procedure}{}{}  
\SetKw{KwRet}{Return}

\newtheorem{lemma}{Lemma}
\newtheorem{thm}{Theorem}

\def \u{\textbf{u}}
\def \x{\textbf{x}}

\def \E{\mathcal{E}}
\def \eE{\mathbb{E}}
\def \opt{{\mathsf{OPT}}}

\def \C{{\mathcal{C}}}
\def \a{{\mathsf{a}}}

\newcommand{\PIMMA}{\nolinkurl{Hephaestus}\xspace}

\title{Hephaestus: Mixture Generative Modeling with Energy Guidance for Large-scale QoS Degradation}

\author{Nguyen Do$^{1, \dagger}$, Bach Ngo$^{2,\dagger}$, Youval Kashuv$^{1}$, Canh V. Pham$^{3}$, Hanghang Tong$^{4}$, My T. Thai$^{1,*}$ \\
$^1$University of Florida, FL, USA \qquad  $^2$The Frazer School, FL, USA\\
$^3$ORLab, Phenikaa University, Viet Nam \quad $^4$University of Illinois at Urbana-Champaign, IL, USA\\
$^\dagger$Equal contribution \qquad  $^{*}$Correspondence to: mythai@cise.ufl.edu\\
}

\begin{document}

\maketitle

\begin{abstract}

We study the Quality of Service Degradation (QoSD) problem, in which an adversary perturbs edge weights to degrade network performance. This setting arises in both network infrastructures and distributed ML systems, where communication quality, not just connectivity, determines functionality. 
While classical methods rely on combinatorial optimization, and recent ML approaches address only restricted linear variants with small-size networks, no prior model directly tackles the QoSD problem under nonlinear edge-weight functions. This work proposes \PIMMA, a self-reinforcing generative framework that synthesizes feasible solutions in latent space, to fill this gap. Our method includes three phases: (1) Forge: a Predictive Path-Stressing (PPS) algorithm that uses graph learning and approximation to produce feasible solutions with performance guarantee, (2) Morph: a new theoretically grounded training paradigm for Mixture of Conditional VAEs guided by an energy-based model to capture solution feature distributions, and (3) Refine: a reinforcement learning agent that explores this space to generate progressively near-optimal solutions using our designed differentiable reward function. Experiments on both synthetic and real-world networks show that our approach consistently outperforms classical and ML baselines, particularly in scenarios with nonlinear cost functions where traditional methods fail to generalize.

\end{abstract}


\section{Introduction}

We consider the problem of degrading path-based system performance through minimal, stealthy perturbations over a set of weighted connections. Formally, let a directed graph $G=(V, E)$ with $|V| = n$ nodes and $|E| = m$ edges, represent a system of interacting components, where each $i$-th edge is associated with a non-decreasing weight function $f_i: \mathbb{N}^0 \rightarrow \mathbb{R}^{+}$. A perturbation $\mathbf{x} =(x_1, x_2, \ldots, x_m)\in \mathbb{N}^{|E|}$ is used to increase edge weights where the new weight $i$-th edge becomes $f_i\left(x_i\right)\forall i \in [m]$. We also refer to $x_e$ as the budget allocated to increase the weight of edge $e$ to $f_e(x_e)$.
Given a set of critical source-target pairs $\mathcal{K}=\left\{\left(s_1, t_1\right), \ldots,\left(s_k, t_k\right)\right\}$,  a maximum perturbation budgets (box constraints) $\mathbf{b}=(b_1, b_2, \ldots, b_m) \in \mathbb{N}^{m}$ and a threshold $T \in \mathbb{R}^{+}$, the goal is to find the lowest-cost perturbation $\mathbf{x}$ such that every shortest path between $\left(s_i, t_i\right) \in \mathcal{K}$ exceeds $T$. We formulate this problem, also called Quality of Service Degradation (QoSD) \cite{nguyen2019network} as the following constrained optimization:
\begin{equation}
\begin{aligned}
\min _{\mathbf{x} \in \mathbb{N}^{m}} & \|\mathbf{x}\|_1 \\
\text { subject to: } & \operatorname{SP}_G\left(s_i, t_i ; \mathbf{x}\right) \geq T \quad \forall\left(s_i, t_i\right) \in \mathcal{K} \\ 
& 0 \leq x_i \leq b_i, x_i \in \mathbb{Z}^+ \cup \{0\} \quad \forall i \in [m]
\end{aligned}
\end{equation}
where $\mathrm{SP}_G(s, t ; \mathbf{x}) = \sum_{e \in \ddot{\rho}_{s,t}} f_e(x_e)$ denotes the length of the shortest path $\ddot{\rho}_{s,t}$ from $s$ to $t$,  under perturbation $\mathbf{x}$. QoSD is an NP-complete problem \cite{nguyen2019network}. Note that the total path cost in $\mathrm{SP}_G(s, t; \mathbf{x})$ is determined by the edge functions $f_i(x_i)$, rather than directly by $x_i$. Although $x_i \in \mathbb{Z}^+ \cup \{0\}$, the problem is not an Integer Program (IP) in general, since the cost functions $f_i(x_i)$ may be nonlinear (e.g., quadratic convex or log-concave). Intuitively, the QoSD problem seeks perturbation $\mathbf{x}$ that ``stress'' global system paths beyond their functional thresholds, not only connectivity while remaining subtle at the local level.

This problem arises in many real-world systems, from networked infrastructures to machine learning, where performance depends not just on connectivity but on effective communication. In blockchain, for example, transactions must reach all miners promptly to maintain consensus; targeted delays can disrupt synchronization and increase the risk of forks \cite{lengbounded, 7958588, pinzon2016double, dennis2016temporal}. The same story as with traffic control systems \cite{lengbounded, chen2018exposing, checkoway2011comprehensive, 5504804, mazloom2016security}, attackers can manipulate signal timing to induce congestion and delay key routes. In Graph Neural Networks (GNN) \cite{zhou2020graph, velivckovic2017graph}, steathly perturbing edge weights or node features can distort message passing and harm predictive performance. 
In all these settings, (i) the system functionality is compromised, regardless of its connectivity, making all the state of the art (SOTA) methods focusing on structural failures become ineffective; 
and (ii) attacks are stealthy distributed across the network, making local defenses ineffective. These characteristics make QoSD a natural and general framework for modeling such vulnerabilities.

Despite its compact formulation, this problem presents several challenges: First, the objective is non-submodular and becomes more complex under nonlinear edge functions such as quadratic or log-concave costs, making greedy or relaxation-based methods ineffective. Second, feasibility checking is expensive: every update to $\mathbf{x}$ requires recomputing global shortest paths for all pairs in $\mathcal{K}$—a bottleneck for large graphs or many constraints. 
Third, the solution space is exponentially  large 
as the number of perturbation vectors grows exponentially with the number of edges, making the search for quality solutions intractable. These challenges limit the applicability of classical optimization and learning methods; thus, it calls for scalable alternatives that can handle a large number of path-based constraints and complex edge dynamics.

To address these challenges, we propose \PIMMA, a three-phase Forge-Morph-Refine generative framework for solving QoSD at scale under both linear and nonlinear edge functions. The framework begins with Forge, generating diverse feasible solutions using our shortest-path attention-based Predictive Path-Stressing (PPS) algorithm, which efficiently approximates global path constraints and \underline{\textit{addresses both scalability and feasibility checking}}. 
We  next employ Morph, a mixture-of-generative-experts model guided by an Energy-Based Model (EBM), which allows deterministic expert expansion to approximate arbitrarily complex multimodal distributions over combinatorial solution spaces. Finally at Refine, a reinforcement learner refines the generated candidates in the latent space using a differentiable reward function, enabling smooth and efficient optimization. As a result, for any unseen graph that shares structural characteristics with those observed during training, the RL agent can quickly select appropriate latent variables and decode them into high-quality solutions—effectively addressing \underline{\textit{the challenge of intractable over large and combinatorial spaces}}.

Overall, our key contributions are summarized as follows:

\begin{itemize}[itemsep=0pt, parsep=0pt, topsep=0pt, leftmargin=1.5em]
  \item We propose \PIMMA, the first end-to-end, generative self-reinforcing framework for QoSD that unifies feasibility search, solution modeling, and optimization into a scalable pipeline capable of handling non-linear costs and large path constraints.
  \item Theoretically, we provide (i) an approximation guarantee for PPS, a corner stone of Forge,  that bounds the cost of the generated solutions relative to the optimum, (ii) a convergence analysis for Energy-Based Model guided mixture training that avoids intractable normalizing constants for Morph, and (iii) a proof that policy refinement in the latent space can be directly performed via gradient ascent, enabling near-optimal solution tuning in Refine with RL. 
    \item Extensive experiments on both synthetic and real-world networks demonstrate that \PIMMA consistently outperforms classical and ML baselines. 
    Ablation studies further validate the complementary role of each phase in improving overall performance, especially under nonlinear cost regimes where traditional methods fail to generalize.
    
\end{itemize}
\section{Related Work}


\textbf{Network Interdiction.}  The QoSD problem was first introduced by Nguyen and Thai \cite{nguyen2019network}. It models a novel form of soft network interdiction, where attackers degrade end-to-end performance by increasing edge weights without explicitly disrupting topological connectivity. This contrasts sharply with classical hard interdiction \cite{israeli2002shortest, smith2013network, smith2020survey, lengbounded}, which assume complete removal of nodes or edges to disrupt flow or connectivity under budget constraints. These classical settings are 
less realistic in stealthy or infrastructure-constrained environments. 
Until recently, Stackelberg routing games under logit-based attacker response models \cite{mai2024stackelberg} adopt a bounded-rational view of adversaries and optimize defender strategies probabilistically over sensitive nodes. While conceptually related to soft disruption, these models differ significantly from QoSD: they focus on node-based protection, rely on attacker stochasticity, and do not generalize to continuous edge-based disruptions.

\textbf{Possible Solving Methods.} Nguyen and Thai \cite{nguyen2019network} proposed three approximation algorithms—Adaptive Trading (AT), Iterative Greedy (IG), and Sampling Algorithm (SA)—for solving the QoSD problem. These methods support both linear and nonlinear $f_e$ but guarantee performance only in the linear case and treat source-target pairs independently, leading to less efficient budget spending. In contrast, \PIMMA leverages a PPS algorithm with guarantees for both settings. This foundation enables the training of self-reinforcing generative model that captures structural correlations across source-target pairs, resulting in more globally effective degradation strategies.

On the other hand, QoSD can be reformulated as an ILP in the special case  when weight function $f_e(.)$ is linear, making recent ML techniques for integer and mixed-integer linear programs (ILPs/MILPs) \cite{BENGIO2021405, li2023from, ZHANG2023205, li2024machinelearninginsidesoptverse} applicable to this setting. One direction integrates ML into solver heuristics \cite{NIPS2014_533d190f, baltean2019scoring, Presolve}, such as branching \cite{NEURIPS2019_d14c2267}, separation \cite{li2023learning, ye2024lightmilpopt}, and cut selection \cite{wang2023learning, NEURIPS2024_71008846}. Another focuses on learning to generate heuristic solutions \cite{nair2021solvingmixedintegerprograms, yoon2022confidencethresholdneuraldiving, Khalil2022, Ye2023GBDT, Zeng2024, ngo2025charmechainbasedreinforcementlearning}.   Notably, the Predict-and-Search (PS) framework \cite{han2023gnn, huang2024contrastive} leverages solvers such as Gurobi \cite{gurobi2021} or SCIP \cite{Achterberg2009} to generate label solutions during training. At inference time, it predicts initial solutions, which are then refined using the same solvers, following the  predict-then-optimize paradigm \cite{Elmachtoub2022, Ferber2020, Zharmagambetov2024}. In contrast, DiffILO \cite{geng2025differentiable} relaxes ILPs into differentiable surrogate objectives and optimizing them end-to-end via gradient descent, enabling unsupervised training without solver-generated labels but only works effectively with small to medium ILP. \PIMMA~ fundamentally differs from these approaches by directly handling both linear and non-linear weight functions, including quadratic-convex and log-concave—without relying on solver and can be scalable to the large graph and number of constraints.


\section{Hephaestus}


\begin{figure*}[htp]
    \centering
\includegraphics[width=0.9\linewidth]{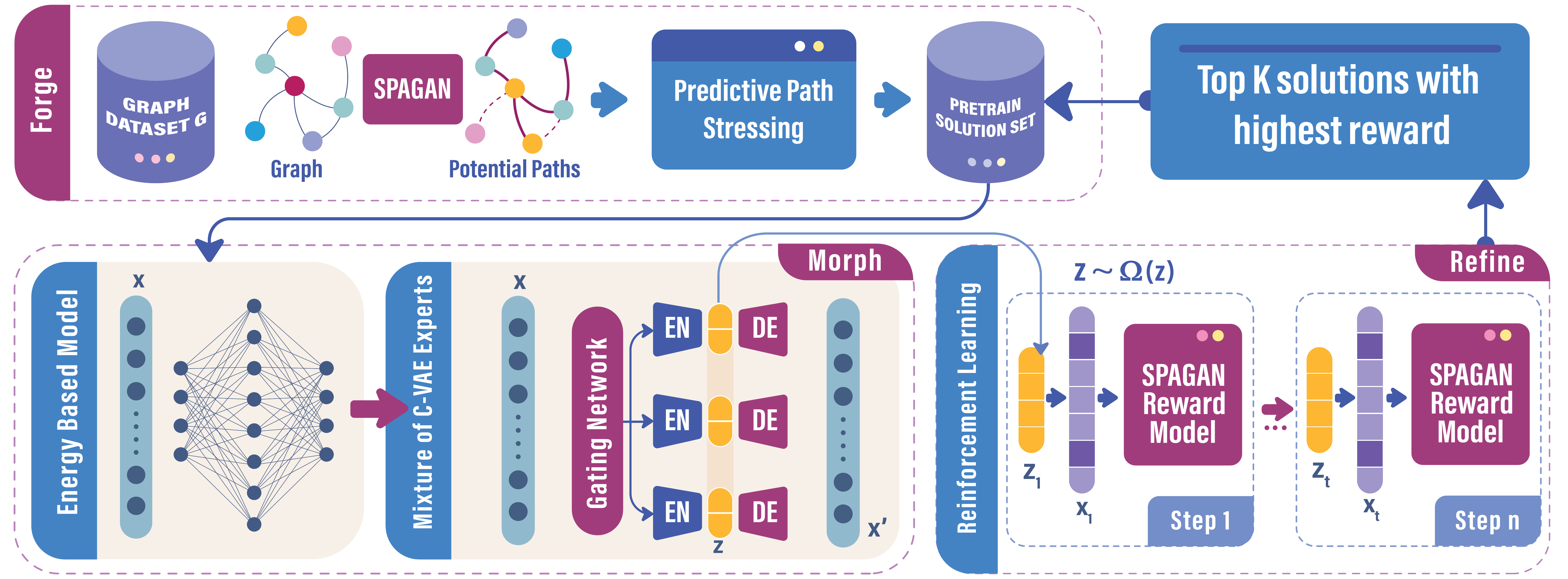}    \caption{Given a graph dataset, we train SPAGAN to predict shortest path costs and guide PPS in generating feasible solutions across varying thresholds, forming a Pretrained Solution Dataset. Then, EBM approximates the underlying solution distribution, while a Mix-CVAE learns to closely match this distribution. Once the Mix-CVAE converges, an RL agent modifies latent samples, which are decoded into lower-cost solutions. The top-$k$ high-reward samples are added back to the Solution Dataset to retrain both the EBM and Mix-CVAE in future episodes, enabling continual improvement.}
    \label{fig:overall-framework}
\end{figure*}

We present our \PIMMA~framework, illustrated in Figure~\ref{fig:overall-framework}, which consists of three phases: (1) {\bf Forge}: Feasible Solution Generation, (2) {\bf Morph}: Energy-Guided Generative Modeling, and (3) {\bf Refine}: Latent Policy Optimization. In Forge, we use SPAGAN \cite{ijcai2019p569} to estimate shortest-path costs and guide our Predictive Path Stressing (PPS) algorithm, which generates feasible solutions (or perturbations) to QoSD and forms a pre-trained solution set $\mathfrak{D}^{\text{sol}}$. The Morph phase trains an Energy-Based Model (EBM) \cite{du2019implicit, lecun2006tutorial} on $\mathfrak{D}^{\text{sol}}$ for guiding a Mixture-of-Conditional VAEs (Mix-CVAE) to approximate the underlying solution feature distribution. To improve coverage, new CVAE \cite{sohn2015learning, Nguyen_Dam_Do_Tran_Pham_2025, do2025swift} experts are dynamically added in regions of EBM where Mix-CVAE shows high modeling error. In the final phase Refine, a reinforcement learning agent operates in the latent space of Mix-CVAE to refine latent variables and synthesize better solutions. These new samples are iteratively added back into $\mathfrak{D}^{\text{sol}}$, allowing both EBM and Mix-CVAE to improve over time.

\subsection{Forge: Feasible Solution Generation}
Training strong generative model in Morph requires a high-quality dataset of feasible solutions  $\mathbf{x}$, as the model can only learn to produce effective solution if the training data itself covers a wide range of valid perturbations with low budget. 
However, generating such data is challenging: for each pair $(s,t)$, the QoSD objective is often non-submodular and the number of feasible paths having cost shorter than $T$ can be extremely large, making it computationally expensive to enumerate or verify them all. To tackle this challenge, we design a hybrid ML-approximation algorithm that leverages a trained shortest-path attention model to guide the approximation search. We begin with a dataset of graphs \(\mathfrak{D}^{\text{graph}} = \{G_i = (V_i, E_i)\}_{i=1}^{N}\) representing diverse sampled network topologies. For each \(G_i\), we define a collection of target node pairs \(\mathcal{K}_i = \{(s_j, t_j)\}_j\), from which shortest path estimations will be drawn. To this end, we train a Shortest Path Graph Attention Network (SPAGAN) \cite{ijcai2019p569}, denoted by $\mathfrak{F}_{\boldsymbol{\theta}} : (G, s, t, \mathbf{x}) \mapsto \widehat{d}_{s,t}$, 
where \(\mathfrak{F}_{\boldsymbol{\theta}}\) takes as input a graph \(G = (V, E)\) and a source-target pair \((s,t)\), and outputs a real-valued estimate \(\widehat{d}_{s,t} \in \mathbb{R}_{+}\) of the shortest-path distance between \(s\) and \(t\) under baseline weight function $f_e$. The model \(\mathfrak{F}_{\boldsymbol{\theta}}\) is trained using a supervised regression loss (i.e Huber Loss \cite{NEURIPS2024_ead8e195, hubercvpr}) over mini-batches of path distances from subgraphs, where ground truth distances are obtained via exact computation (i.e Dijkstra algorithm \cite{dijkstra1959note}). By training over a diverse set of subgraphs extracted from \(\mathfrak{\text{graph}}\), the model learns transferable embeddings that capture structural priors and efficiently generalize to large graphs, enabling efficient solution verification in unseen instances.

Once \(\mathfrak{F}_{\boldsymbol{\theta}}\) is trained, we use it to guide the searching of diverse solutions $\mathbf{x}$. 
Specifically, we propose a PPS algorithm to iteratively search adversarial perturbations that elevate the predicted path length \(\widehat{d}_{s,t}(\mathbf{x})\) of each pair $(s, t)$ beyond a given threshold \(T\). Unlike traditional approximation methods~\cite{nguyen2019network} that require costly repeat exact path computation, PPS leverages the efficiency of \(\mathfrak{F}_{\boldsymbol{\theta}}\) to estimate marginal gains in violation potential and guides the perturbation accordingly. Let \(P\) denote the current set of shortest paths under perturbation \(\mathbf{x}\), i.e., $P = \left\{ \ddot{\rho}_{s,t} \mid (s,t) \in \mathcal{K},\ \mathfrak{F}_{\boldsymbol{\theta}}(G, s, t; \mathbf{x}) < T \right\}$
where $\mathfrak{F}_{\boldsymbol{\theta}}(G, s, t, \mathbf{x}) \approx \sum_{e \in \ddot{\rho}_{s,t}} f_e(x_e)$ returns the predicted shortest path length under \(\mathbf{x}\), if $\mathfrak{F}_{\boldsymbol{\theta}}$ is well trained. To quantify the progress of \(\mathbf{x}\), we define the potential function $\mathcal{C}(P, \mathbf{x}) = \sum_{\ddot{\rho}_{s,t} \in P} \min\left(T, \mathfrak{F}_{\boldsymbol{\theta}}(G, s, t, \mathbf{x})\right)$. Each iteration of PPS selects an edge \(e \in \mathcal{E}\) and an increment \(\Delta x_e \in \mathbb{N}\) that maximizes the predicted gain-to-cost ratio:
\begin{equation}
    (e^{\ast}, \Delta^{\ast}) = \arg\max_{e \in \mathcal{E},\ \Delta \in [1, b_e - x_e]} \frac{\mathcal{C}(P, \mathbf{x} + \Delta \cdot \mathbf{1}_e) - \mathcal{C}(P, \mathbf{x})}{\Delta}
\end{equation}
where \(\mathbf{1}_e\) is the standard basis vector corresponding to edge \(e\), and \(b_e\) is the upper budget bound for that edge. After determining $\left(e^*, \Delta^*\right)$, perturbation is updated via $\mathbf{x} \leftarrow \mathbf{x}+\Delta^* \cdot \mathbf{1}_{e^*}$, and the path set $P$ is recomputed via $\mathfrak{F}_{\boldsymbol{\theta}}$. This process is repeated until almost all node pairs $(s, t) \in \mathcal{K}$ are predicted to have path lengths exceeding the target threshold, i.e., $\C(P, \x)\geq |P|T-\bar{\epsilon}$, where $\bar{\epsilon}$ is an input parameter. For each graph $G_i \sim \mathfrak{D}^{graph}$, the PPS algorithm is executed independently across a set of QoS thresholds $\mathcal{T}=\{T_1, \ldots, T_M\}$ and pairs $\mathbf{K}=\{\mathcal{K}_1, \ldots, \mathcal{K}_M\}$, producing a corresponding set of feasible perturbations $\{\mathbf{x}_{\mathcal{K},T}^{(i)} | \mathcal{K} \in \mathbf{K}, T \in \mathcal{T}\}$. The resulting pretrained solution dataset is thus formalized as $\mathfrak{D}^{sol}=\{(G_i, \mathcal{K}, T, \mathbf{x}_{\mathcal{K},T}^{(i)}) | G_i \in \mathfrak{D}^{graph}, \mathcal{K} \in \mathbf{K}, T \in \mathcal{T}\}$ which forms the basis of the Morph phase (Section~\ref{subsec:latent-fragility-morphing}). For notational simplicity, we refer to each element of $\mathfrak{D}^{sol}$ as a training instance $(G, \mathcal{K}, T, \mathbf{x})$, representing a specific degradation scenario and having a guarantee:

\begin{thm}(PPS Ratio)
\label{thm:pps-ratio}
Let \( h = \lceil T / w_{\min} \rceil \), where \( w_{\min} = \min_{e \in E} w_e \). 
Assume that the set $\E$ is chosen from $E$ such that $\Pr[\E^* \subseteq \E]=\a $, where $\E^*$ is the set of edges of the optimal solution and $\a \in (0, 1)$ is a constant. Given a parameter $\bar{\epsilon}>0$, then running PPS on $\E$ yields a solution \( \x \) such that $\C(P, \x) \geq |P|T-\bar{\epsilon}$ and $\eE[\|\x\|_1] \le (1+h\ln(n)+\ln T +\ln(1/\bar{\epsilon})) \opt/\a$.  (Proof in Appx \textbf{B.1})
\end{thm}
Theorem~\ref{thm:pps-ratio} ensures that every solution generated by PPS achieves a provably bounded cost relative to the optimal. Such quality assurance is critical: the generative model introduced in Morph will be trained purely on \(\mathfrak{D}^{sol}\), thus its quality directly determines the generative capacity and generalization performance. Without such a theoretical grounding, the learned latent distribution could diverge or collapse around suboptimal patterns. 
\subsection{Morph: Energy-Guided Generative Modeling}
\label{subsec:latent-fragility-morphing}

Given the pre-trained solution set $\mathfrak{D}^{\text{sol}}$ from the Forge phase, Morph aims to model the underlying pattern of high-quality solutions through the conditional distribution $p(\mathbf{x} \mid \mathbf{c})$, where the condition $\mathbf{c} = [G, \mathcal{K}, T]$ encodes the graph, the set of critical node pairs, and the threshold context. Since the true distribution $p$ is unknown, we approximate it using a conditional generative model $\Omega(\mathbf{x} \mid \textbf{c})$ that can generate feasible solutions for arbitrary $[G, \mathcal{K},T]$ instance. To achieve this, we implement $\Omega$ as a Mixture of Conditional VAEs (Mix-CVAEs), denoted $\Omega = [\Omega_0, \dots, \Omega_N]$, where each expert $\Omega_i$ consists of an encoder $\mathcal{P}_\psi$ and a decoder $\mathcal{M}_\phi$. For a given training pair $(\mathbf{x}, \textbf{c})$, the encoder produces a latent posterior $\tilde{q}_\psi(\mathbf{z} \mid \mathbf{x}, \textbf{c})$, and the decoder reconstructs the input via $\tilde{p}_\phi(\mathbf{x} \mid \mathbf{z}, \textbf{c})$, forming the mapping $\Omega_i(\mathbf{x}, \textbf{c}) = \mathcal{M}_\phi(\mathcal{P}_\psi(\mathbf{x}, \textbf{c}))=\mathcal{M}_\phi \circ \mathcal{P}_\psi$. We train each CVAE expert by maximizing the evidence lower bound (ELBO) \cite{Nguyen_Dam_Do_Tran_Pham_2025} on samples $(G, \mathcal{K},T, \mathbf{x}) \sim \mathfrak{D}^{\text{sol}}$:
\begin{equation}
    \mathcal{L}^{ELBO}_{\Omega_i} = \mathbb{E}_{\tilde{q}_\psi}\log \tilde{p}_\phi(\mathbf{x} \mid \mathbf{z}, \textbf{c}) - \mathrm{KL}[\tilde{q}_\psi(\mathbf{z} \mid \mathbf{x}, \textbf{c}) \,\|\, \tilde{p}_\phi(\mathbf{z} \mid \textbf{c})]
    \label{eqn:ELBO}
\end{equation}
where the first term encourages accurate reconstruction of $\mathbf{x}$, and the second regularizes the latent representation $\mathbf{z}\in \mathbb{R}^d$ to align with a context-aware prior. After training, new solutions can be generated by sampling $\mathbf{z} \sim \tilde{p}_\phi(\mathbf{z} \mid \textbf{c})$ and decoding via $\tilde{\mathbf{x}} = \mathcal{M}_\phi(\mathbf{z}, \textbf{c})$.

\textbf{Mode-Seeking Problem.} Since optimizing the ELBO is merely an indirect way to minimize the reverse KL divergence, this objective often encourages the learned distribution $\tilde{p}_\phi(\mathbf{x} \mid \textbf{c})$ to focus on the dominant modes that appear most frequently in the dataset $\{\mathbf{x}_i\}$. However, if $p(\mathbf{x})$ is truly multimodal \cite{nguyen2024expert}, a single expert $\Omega_i$ can exhibit mode-seeking behavior \cite{dang2024neural}—i.e., it may ignore rare but critical regions of $p(\mathbf{x} \mid \textbf{c})$. More severely, we do not even know which regions of $p$ are not being captured by the CVAE $\Omega_i$, since the true form of $p$ is unknown. 

\textbf{Energy-Based Guidance for Generative Modeling.} To address the above limitation, we introduce an Energy-Based Model (EBM) \cite{lecun2006tutorial} defined as \(q(\mathbf{x}) = \frac{1}{Z} \exp\left(-E(\mathbf{x})/\tau\right)\), where \(E(\mathbf{x})\) is a learnable energy function, \(\tau > 0\) is a temperature parameter controlling the sharpness of the distribution and the normalizing constant function \(Z = \int_{\mathcal{X}} \exp\left(-E(\mathbf{x})/\tau\right) d\mathbf{x}\) ensures \(q(\mathbf{x})\) is a valid probability distribution. Unlike CVAE, EBM makes no strong parametric assumptions about the form of the true distribution \(p(\mathbf{x})\); it instead defines a flexible energy landscape shaped directly by the energy values. We thus employ EBM \(q(\mathbf{x})\) as a \textit{surrogate} for the true—but unknown—distribution \(p(\mathbf{x})\), and aim to learn \(q\) by minimizing $\min_{q \in \mathcal{Q}} \; \mathrm{KL}\left(p(\mathbf{x}) \,\|\, q(\mathbf{x})\right)$, where \(\mathcal{Q}\) denotes the feasible family of EBMs parameterized by \(E(\mathbf{x})\). Simultaneously, we train the generative model \(\Omega(\mathbf{x})\) (e.g., a Mix-CVAEs) to match \(q(\mathbf{x})\) by minimizing the KL divergence $\min_{\Omega \in \mathfrak{E}} \; \mathrm{KL}\left(q(\mathbf{x}) \,\|\, \Omega(\mathbf{x})\right)$, where \(\mathfrak{E}\) represents the feasible set of generative distributions realizable by our model class. Ideally, when the two KL divergences become zero, then $p(x)=q(x)=\Omega(x)$ or indirectly $\Omega(x) = p(x)$.

\begin{figure*}[htp]
    \centering
    \includegraphics[width=1.0\linewidth]{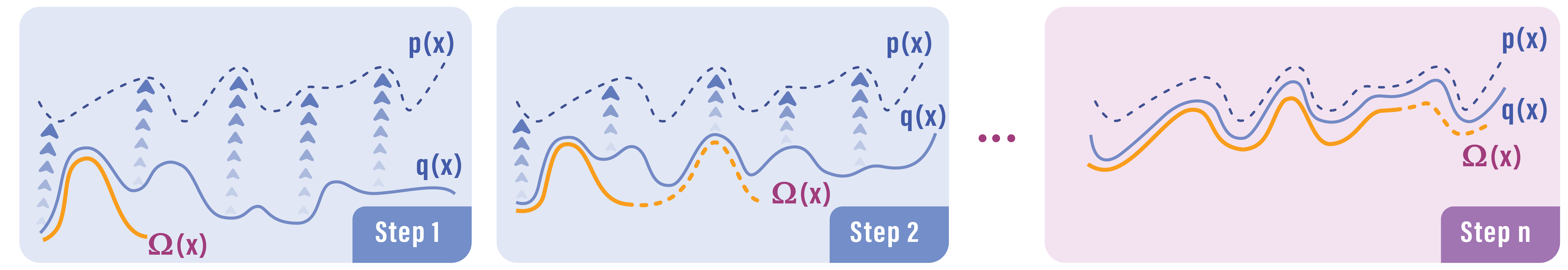}
    \caption{At each iteration, the generative model $\Omega(x)$ adds an expert in regions where the log-ratio $\chi(x) = \log \left(q(x) / \Omega(x) \right)$ exceeds $\delta$, thereby covering more of $q(x)$. Meanwhile, $q(x)$ adapts to align with $p(x)$. As the steps progress, $q(x)$ increasingly approximates $p(x)$, and $\Omega(x)$, augmented by new experts, converges to $q(x)$, ultimately yielding $p(x) \approx q(x) \approx \Omega(x)$.}
    \label{fig:indirectly-approx}
\end{figure*}

However, a single C-VAE may only capture a dominant mode of the distribution $q(x)$ as we mentioned earlier. To ensure comprehensive coverage, we monitor the density gap between the EBM and the current generative model via the log-ratio function $\chi(\mathbf{x})=\log q(\mathbf{x})/\Omega(\mathbf{x})$. Whenever $\chi(\mathbf{x})>\delta$ for some threshold $\delta>0$, we interpret this region as underrepresented by the current C-VAE model $\Omega$, and dynamically add a new C-VAE expert $\Omega_{N+1}(\mathbf{x})$ specialized to this region forming new Mixture of C-VAE $\Omega'$. This expert is integrated into the mixture through a gating mechanism, ensuring that the overall generative model incrementally improves its coverage of high-density regions in $q(\mathbf{x})$, and thereby indirectly approximates the true data distribution $p(\mathbf{x})$ (See Figure \ref{fig:indirectly-approx}) . 

\begin{thm}(Expert Augmentation Efficiency): Suppose there exists a constant \(\epsilon>0\) such that \(\mathbb{P}_{p(x)}\{\chi(x)>\delta\}\ge \epsilon\); on the region \(\{x: \chi(x)>\delta\}\), a new expert \(\Omega_{N+1}(x)\) is added with gating weight \(\alpha(x)=\mathbf{1}\{\chi(x)>\delta\}a_0\) (  \(0<a_0\le\bar{a}<1\)) and trained to satisfy \(\Omega_{N+1}(x)\ge c\,q(x)\) for some \(c\in(0,1)\); then the updated mixture \(\Omega'(x)=\alpha(x)\,\Omega_{N+1}(x)+(1-\alpha(x))\,\Omega(x)\) satisfies \(\mathrm{KL}(q\|\Omega')\le \mathrm{KL}(q\|\Omega)-\gamma(\delta,\epsilon)\), where \(\gamma(\delta,\epsilon)=a_0(\delta+\log c)\,\epsilon_0>0\) and \(\epsilon_0>0\) is a lower bound on \(\int_{\chi(x)>\delta}q(x)dx\). (Proof in Appendix \textbf{B.2})
\label{thm:expert-aug}
\end{thm}

From Theorem \ref{thm:expert-aug}, we observe that each time a new expert is added, $\mathrm{KL}(q\|\Omega')$ is reduced by at least \(\gamma(\delta,\epsilon)\) relative to the previous $\mathrm{KL}(q\|\Omega)$. 
Since the initial KL divergence with only one expert is finite, repeatedly augmenting the mixture with new experts allows us, in the limit of infinitely many experts without sacrificing the computational cost \cite{nguyen2024statistical, NEURIPS2022_91edff07}, to drive the $\mathrm{KL}(q\|\Omega')$ to zero on 
\(\mathfrak{D}^{sol}\). However, directly computing 
$\mathrm{KL}(p(\mathbf{x}) \| q(\mathbf{x}))$ and $\mathrm{KL}( q(\mathbf{x})\|\Omega(\mathbf{x}) )$ is highly challenging in practice, as it requires evaluating the normalization constant $Z$ of the EBM. Estimating $Z$ typically involves Markov Chain Monte Carlo (MCMC) \cite{gilks1995markov}, which becomes extremely expansive during the EBM training. 
Fortunately, as we will show next, there is no need to compute \(Z\). 

\begin{thm}(Normalization Free Function)
The objective function $\min_{q}\,\max_{\Omega}\{\mathrm{KL}(p\|q)-\mathrm{KL}(\Omega\|q)\},$ is normalizing free which is independent with \(Z\). (Proof in Appendix \textbf{B.3})
\label{thm:normalizing-free-minmax}
\end{thm}

Per Theorem~\ref{thm:normalizing-free-minmax}, we completely avoid the need to compute $Z$ by reformulating the problem as a minimax objective, where the normalizing constant cancels out in the difference of KL divergences:
\begin{equation}
    \min _{q\in\mathcal{Q}} \max _{\Omega \in \mathfrak{E}} \{\mathrm{KL}(p \| q)-\mathrm{KL}(\Omega \| q) \}
    \label{eqn:minimax}
\end{equation}
Any terms involving $\log Z$ appear in both $\mathrm{KL}(p \| q)$ and $\mathrm{KL}(\Omega \| q)$ and hence cancel each other exactly. This cancellation frees us from having to perform expensive MCMC estimates of $Z$, yet still allows the EBM $q$ to guide the generative model $\Omega$ effectively. 

\textbf{Minimax Optimization.} From Equation \ref{eqn:minimax}, the EBM is optimized by minimizing the difference between the expected energy under the data distribution and the expected energy under the Mix-CVAE model distribution $\min_\theta \mathbb{E}_{p(\mathbf{x})}[E_\theta(\mathbf{x})] - \mathbb{E}_{\Omega(\mathbf{x})}[E_\theta(\mathbf{x})]$ (minimax derivation in Appendix \textbf{B.3}). To stabilize training and ensure that energy values remain bounded, we incorporate a commonly-used regularization term \cite{du2019implicit} based on the squared energies $\gamma \left( \mathbb{E}_{p(\mathbf{x})}[E_\theta(\mathbf{x})^2] + \mathbb{E}_{\Omega(\mathbf{x})}[E_\theta(\mathbf{x})^2] \right)$ where \(\gamma > 0\) is a regularization coefficient. For Mix-CVAE approximating EBM, we modify the standard ELBO to incorporate EBM guidance during expert training. Specifically, based on Equation~\ref{eqn:ELBO}, we define a guided objective
$\mathcal{L}^{guide}_{\Omega_i} = \mathcal{L}^{ELBO}_{\Omega_i} + \lambda \cdot \mathbb{E}_{\tilde{p}_\phi(\mathbf{z} \mid c)}\bigl[E_\theta\bigl(\tilde{p}_\phi(\mathbf{x} \mid \mathbf{z}, c)\bigr)\bigr]$,
where the added term penalizes low-quality samples in high-energy (i.e., low-density) regions under the energy-based model $q_\phi(\mathbf{x}) \propto \exp(-E_\theta(\mathbf{x}))$.

\subsection{Refine: Latent Policy Optimization}

Once the Mix-CVAE \(\Omega\) has successfully captured the true solution distribution, similar inputs (consisting of $\mathbf{x}$, $G$, $\mathcal{K}$, and $T$) that lead to similar-quality solutions are mapped to nearby points in the continuous latent space. This property makes the search and optimization process more efficient. In Refine, we train an RL agent with a policy $\pi$ to blend and explore that latent space, aiming to generate solutions that are better than those in the original pretrained dataset $\mathfrak{D}^{sol}$. These newly discovered solutions are then added back into $\mathfrak{D}^{sol}$, reinforcing their feature patterns and making it easier to retrain the generative model in future iterations. The optimization of policy $\pi$ over the latent space is formulated as a Markov Decision Process (MDP) \cite{WHITE19891}, $\mathcal{M} \stackrel{\text { def }}{=}(\mathcal{S}, \mathcal{A}, \Gamma, \mathcal{R})$. This includes (i) a finite sets of states $\mathcal{S}$, (ii) a finite set of actions $\mathcal{A}$, (iii) a transition distribution $\Gamma\left(s^{\prime} \mid s, a\right)$ where $s, s^{\prime} \in \mathcal{S}, a \in \mathcal{A}$ and (iv) a reward function $\mathcal{R}: \mathcal{S} \times \mathcal{A} \rightarrow \mathbb{R}$. We specify each component as follows:

\textbf{State} $(s \in \mathcal{S})$ : A state is defined by latent representations $s_i=\left(\boldsymbol{z}_i, \textbf{c}\right)$ for $\boldsymbol{z}_i=\mathcal{P}_\psi\left(\mathbf{x}_i, \textbf{c}\right)$ and $\mathbf{x}_i \in \mathfrak{D}^{sol}_{\text {ep}}$, where $\boldsymbol{z}_i \in \mathbb{R}^d$ is the latent vector produced by the encoder $\mathcal{P}_\psi$ from the input data $\mathbf{x}_i$.

\textbf{Action} ($a \in \mathcal{A}$): An action $a_i=\left(\mu_i, \sigma_i\right)$ is a vector of two components: $\mu_i \in \mathbb{R}^d$ (predicted mean) and $\sigma_i \in \mathbb{R}^d$ (predicted scale). The modification vector $\delta_i=\sigma_i \cdot \epsilon_{noise}+\mu_i$, where $\epsilon_{noise} \sim \mathcal{N}(0, I)$, and the latent vector is updated as $\hat{\boldsymbol{z}}_i=\boldsymbol{z}_i+\delta_i$.

\textbf{Transition Dynamics (\(\Gamma\))}: When an action \( a_i = (\mu_i, \sigma_i) \) is taken, a new state \( s_{i+1} = (\boldsymbol{z}_{i+1}, \textbf{c}) \), where  $ \boldsymbol{z}_{i+1} = \hat{\boldsymbol{z}}_i$, is then formed.

\textbf{Reward Function} ($\mathcal{R}$): Given $z_{i+1}$, the decoder $\mathcal{M}_\phi$ decodes it into a solution $\hat{\mathbf{x}}_{i+1} = \{x_1, \dots, x_m\}$. To make optimization smoother, we apply a soft transformation to each element: $\bar{x}_j = \log(1 + e^{x_j})$ where $x_j \in \hat{\mathbf{x}}_{i+1}$ to produce a smooth vector $\bar{\mathbf{x}}_{i+1}$.  Moreover, instead of using a binary feasibility label (feasible or not) for each pair $(s,t) \in S$, we define a smooth total feasibility score over all pairs:
\begin{equation}
    \digamma(G, \mathcal{K}, \hat{\mathbf{x}}_{i+1}) = \sum_{(s, t) \in \mathcal{K}} \frac{1}{1 + \exp\left(-\zeta \cdot \left(\mathfrak{F}_{\boldsymbol{\theta}}(G, s, t; round(\hat{\mathbf{x}}_{i+1})) - T\right)\right)}.
    \label{eqn:smooth-feasibility}
\end{equation}
Here, for each pair $(s,t)$, $\zeta$ controls how sharply the score changes near the threshold $T$. If the predicted path cost slightly exceeds $T$, the sigmoid quickly pushes the score closer to 1. This formulation ensures that $\digamma(G, S, \hat{\mathbf{x}}_{i+1})$ provides a continuous and differentiable reward signal, which is more suitable for optimizing the RL policy. Finally, our reward function is defined as: 
\begin{equation}
\mathcal{R}(\mathbf{x}_{i+1})= \digamma(G, \mathcal{K}, \hat{\mathbf{x}}_{i+1}) - \varkappa \cdot \log (1+ \|\bar{\mathbf{x}}_{i+1}\|_1).
\label{eqn: reward_function}
\end{equation}
This reward formulation balances two objectives: the first term encourages the generation of solution vectors 
by pushing the predicted path cost above $T$, while the second term, weighted by $\varkappa$, penalizes excessive budget usage, thus promoting stealth and cost-effectiveness. We intentionally keep the cost penalty relatively small so that the RL agent focuses primarily on finding a valid solution (i.e, only increasing weights) before optimizing cost. In what follows, we show several properties of our dense reward function which make the RL training process much easier.

\begin{lemma} (Differentiable Reward Function) Let $\mathfrak{F}_\theta: \mathcal{X} \rightarrow \mathbb{R}$ be differentiable on an open set $\mathcal{X} \subset \mathbb{R}^m$. The reward function
$
\mathcal{R}(\mathbf{x})$ is differentiable on $\mathcal{X}$. (Proof in Appendix \textbf{B.4})
\label{lm: differentiable}
\end{lemma}
\begin{thm} (Reward Estimation Consistency)
Assume that $\Omega$ has converged properly, for any perturbed latent vector $\hat{z}_i := z_i + \hat{\epsilon} \cdot \nabla_{z_i} \mathcal{R}(\mathcal{M}_\phi(z_i, \textbf{c}))$ with small $\hat{\epsilon} > 0$, we have  $\mathcal{R}\left(\hat{\mathbf{x}}_{i}\right) > \mathcal{R}\left(\mathbf{x}_{i}\right)$, where $\hat{\mathbf{x}}_{i} = \mathcal{M}_\phi(\hat{z}_i, \textbf{c})$. (Proof in Appendix \textbf{B.5})
\label{thm: Reward Estimation Consistency}
\end{thm}

Specifically, by Theorem \ref{thm: Reward Estimation Consistency}, small perturbations in the latent code \(z_i\) following the gradient of the reward yield strictly higher returns. Since the reward \(R(x)\) is differentiable (Lemma \ref{lm: differentiable}), we can perform gradient ascent \cite{ruder2016overview} directly in latent space as in Equation \ref {eqn: reward_function}. Concretely, starting from \(z_i\), we iterate  $\hat z_i \leftarrow  z_i + \hat{\epsilon} \cdot \nabla_{z_i} \mathcal{R}(\mathcal{M}_\phi(z_i, \textbf{c}))$
until convergence to $z_i^*$ leading to next state \( s_{i+1} = (\boldsymbol{z}_{i+1} = \hat z_i^*, \textbf{c}) \).  The resulting latent offset \(\hat\delta_i = \hat z_i - z_i\) together with \(\sigma_i\) defines a feasible action \(\tilde a_i = (\hat\delta_i,\sigma_i)\). This warm-start procedure mitigates inefficient random exploration in the early stages and accelerates convergence by guiding policy $\pi$ towards locally optimal regions. Once the model reaches sufficient reward levels, we allow it to autonomously explore and exploit the latent space using standard reinforcement learning framework, leveraging efficient exploration strategies from the RL literature \cite{ecoffet1901go, pathak2017curiosity, silver2017mastering}. See Appendices \textbf{A} and \textbf{C} for the full algorithm and its 
analysis.

{\bf Inference Process.} Given a new instance problem, policy $\pi$ iteratively modifies a random latent vector $z$ in latent space to maximize the reward defined in Equation~\ref{eqn: reward_function}. However, it is not guaranteed that the decoded solution $\hat{\mathbf{x}}$ from $\mathcal{M}_\phi(z)$ is valid. To address this, we introduce a feasibility refinement algorithm called PPS-I, a variant of the PPS algorithm. Unlike PPS, which relies on the learned estimator $\mathfrak{F}_{\boldsymbol{\theta}}(\cdot; \hat{\mathbf{x}})$ for the approximation of the shortest path, PPS-I replaces it with the exact calculation of the shortest path using Dijkstra's algorithm to ensure 100\% feasibility. This yields $\a = 1$ in the ratio stated in Theorem~\ref{thm:pps-ratio}. Moreover, with strong Mix-CVAE $\Omega$, the initial solution $\hat{\mathbf{x}}$ produced by $\pi$ is often close to valid, making the refinement overhead of PPS-I significantly minimized. Implementation details and analysis of PPS-I are provided in Appendix \textbf{A4}.


\section{Experiments}

\textbf{Settings.} We evaluate the effectiveness and scalability of Hephaestus on synthetic and real-world networks under varying QoSD thresholds. For synthetic graphs, we generate Erdős–Rényi topologies with $n = 1024$ nodes and vary edge density $l$, fixed threshold $T = 20$, $|\mathcal{K}|=10$ critical pairs. Real-world datasets include Email \cite{yin2017local}, Gnutella \cite{leskovec2007graph}, RoadCA \cite{leskovec2009community}, and Skitter \cite{leskovec2005graphs}, covering diverse scales and domains. For each, we sample $50$ critical pairs, compute their maximum baseline shortest-path length, and set $T$ to 140\%–260\% of this value to normalize degradation difficulty across topologies. Dataset statistics are in Appendix \textbf{C.1}. For training the entire \PIMMA framework in both synthetic and real networks, we generate a large corpus of graphs with varying architectures (e.g., Barabási-Albert, Erds-Rényi, and Watts-Strogatz), as well as diverse edge densities and thresholds, while keeping the number of nodes consistent with those in the testing graphs.

\textbf{Baselines.} We compare \PIMMA~with a range of baselines, including classical approximation methods and learning-based IP solvers. Approximation baselines include Adaptive Trading (AT), Iterative Greedy (IG), and Sampling Algorithm (SA) from \cite{nguyen2019network}, which offer varying trade-offs between cost-effectiveness, concavity handling, and scalability. For learning-based IP methods, we evaluate three SOTA methods for ILP: DIFFILO \cite{geng2025differentiable}, Predict-and-Search \cite{han2023gnn}, and L-MILPOPT \cite{ye2024lightmilpopt}, which combine neural predictors with ILP solvers like Gurobi. Baseline solutions are refined using Gurobi with a 3000s timeout—tripled from the original setup. For large networks (e.g., RoadCA, Skitter), we extend this to 6000s and enable `MIPFocus=1` \cite{geng2025differentiable, bertsimas2016best, tjeng2017evaluating} with heuristics to prioritize feasibility. This setup is crucial, as exact solving often leads to long runtime or memory errors in large networks. 
In constrast, our method employs PPS-I to ensure 100\% feasible refinement without relying on exact solvers, while still achieving significantly faster performance due to its approximate nature. Note that we excluded a diffusion-based generative solver (e.g., the Gurobi-guided diffusion model \cite{Gurobi-guided-diffusion}) due to its scalability issues in large-scale constrained combinatorial optimization problems, particularly over graphs with millions of nodes. In particular, this approach requires costly iterative sampling using a series of VAE models, with 100 denoising steps for Denoising Diffusion Implicit Model (DDIM) \cite{song2021denoising} and 1000 steps for Denoising Diffusion Probabilistic Model (DDPM) \cite{10.5555/3495724.3496298}, making them computationally impractical to evaluate in large-scale setups such as RoadCA or Skitter.

\textbf{Edge Weight Function Settings.}
We evaluate methods under three types of $f_e(x)$ representing different degradation dynamics: (i) \underline{\textit{Linear}}: $f_e(x) = \aleph(x)$, modeling uniform delay per unit cost; (ii) \underline{\textit{Quadratic Convex}}: $f_e(x) = \aleph(x^2)$, capturing congestion effects with rapidly increasing cost; and (iii) \underline{\textit{Log Concave}}: $f_e(x) = \aleph(\ln x)$, modeling diminishing returns as in error-rate degradation. 
ML-based IP solvers operate only under linear functions, so we compare all methods in this setting, including the exact solver. For convex and concave settings, only the approximation methods and \PIMMA~are applicable. Notably, Gurobi fails on the Log Concave case due to the non-convex feasible region induced by logarithmic constraints, and is excluded from that evaluation.

\begin{figure*}[htp]
    \centering
    \captionsetup[subfigure]{skip=0.1pt}  
    \begin{subfigure}[t]{0.31\linewidth}
        \centering
         \includegraphics[width=\linewidth]{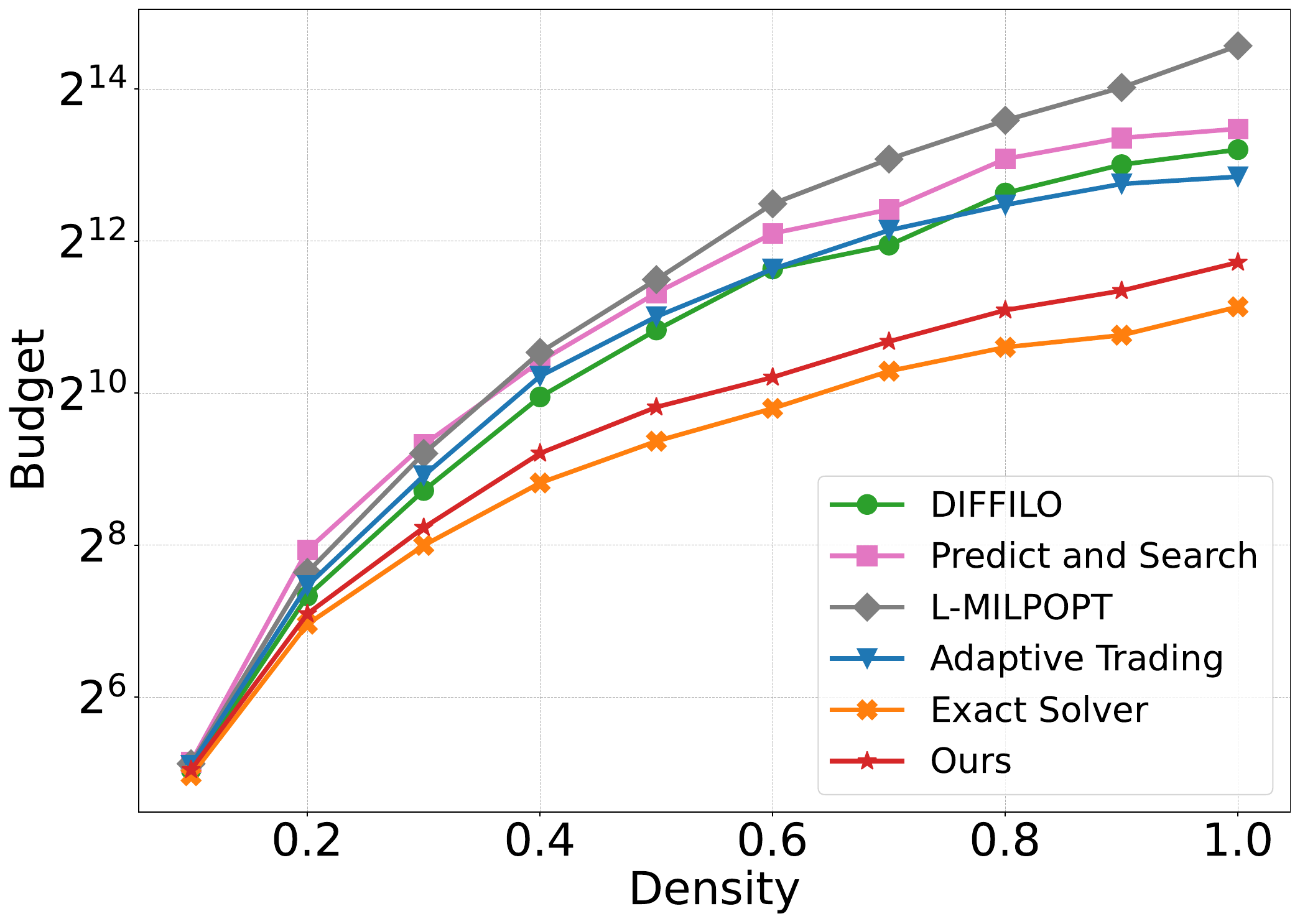}
        \caption{Linear}
    \end{subfigure}
    \hfill
    \begin{subfigure}[t]{0.31\linewidth}
        \centering
        \includegraphics[width=\linewidth]{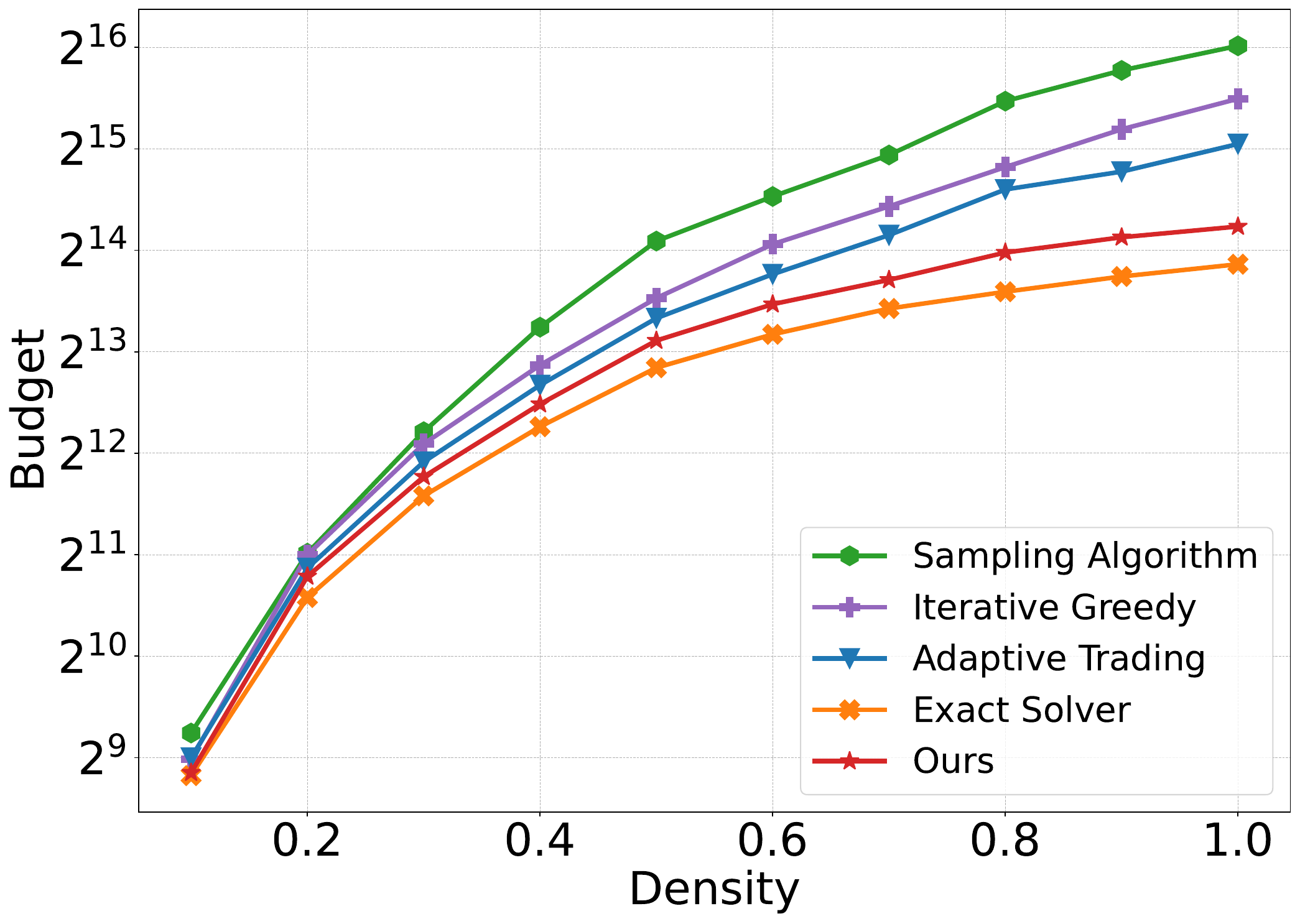}
        \caption{Quadratic Convex}
    \end{subfigure}
    \hfill
    \begin{subfigure}[t]{0.31\linewidth}
        \centering
        \includegraphics[width=\linewidth]{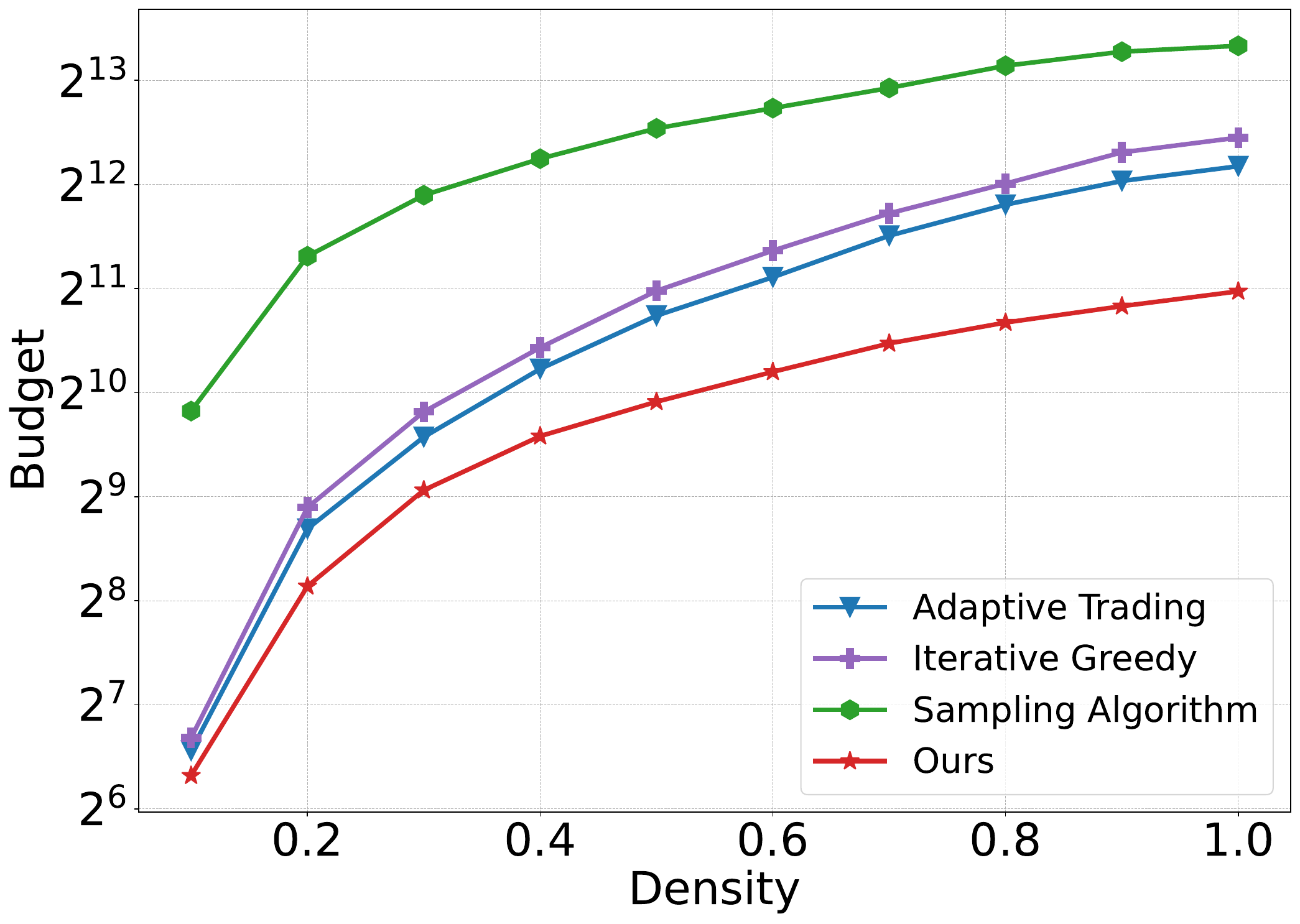}
        \caption{Log Concave}
    \end{subfigure}
    \vspace{-5pt}  
    \caption{Our solution quality on the testing synthetic graph with varying density is evaluated under Linear, Quadratic Convex, and Log-Concave functions. ML-based ILPs are compared with the best-performing approximation baseline AT for Linear weights but excluded for the other two due to incompatibility; exact solver also does not support Log-Concave.}
    \label{fig:synthetic Solution quality of algorithms}
\end{figure*}

\begin{table}[htp]
\centering
\scriptsize 
\setlength{\tabcolsep}{1.4pt} 
\renewcommand{\arraystretch}{0.6}
\begin{tabular}{l *{4}{r} *{4}{r} *{4}{r} *{4}{r}}
\toprule
& \multicolumn{4}{c}{Email} & \multicolumn{4}{c}{Gnutella} & \multicolumn{4}{c}{RoadCA} & \multicolumn{4}{c}{Skitter} \\
\cmidrule(lr){2-5} \cmidrule(lr){6-9} \cmidrule(lr){10-13} \cmidrule(lr){14-17}
Method & 140\% & 180\% & 220\% & 260\% & 140\% & 180\% & 220\% & 260\% & 140\% & 180\% & 220\% & 260\% & 140\% & 180\% & 220\% & 260\% \\
\midrule
Adaptive Trading
        & 2653 & 5460  & 8226  & 9675
        & 3528 & 6613  & 8528  & 10656
        & {12651}  & {16972}   & {23504}   & {32935}
        & {367026}  & {903278}   & {1739299}   & {3018467}    \\
      Iterative Greedy
        & 2673 & 5458  & 8258  & 9713
        & 3537 & 6626  & 8564  & 10674
        & {13782}  & {17830}   & {24076}   & {33127}
        & {389415}  & {915682}   & {1983011}   & {3125008}  \\
      Sampling Alg.
        & 4363  & 9249   & 14665  & 17290
        & 4200  & 7775   & 14920  & 16427
        & {23742}  & {43822}   & {70823}   & {90166}
        & {597239}  & {1789237}   & {2871248}   & {6294921}    \\
      DIFFILO
        & \textbf{2621} & \textbf{5308}  & 8303  & 9695
        & \textbf{3490} & 6662  & 8619  & 10987
        & {11043}  & {17161}   & {24031}   & {32976}
        & {350397}  & {902366}   & {1724336}   & {3012839}   \\
      Predict \& Search
        & 2654 & 5421  & 8458  & 9701
        & 3645 & 6808  & 8839  & 11364
        & {11196}  & {18084}   & {24967}   & {33954}
        & {358937}  & {929115}   & {1849725}   & {3197237}   \\
      L-MILPOPT
        & 2651 & 5462  & 8486  & 9811
        & 3615 & 6922  & 9185  & 11772
        & {12069}  & {19302}   & {25310}   & {34866}
        & {360994}  & {931678}   & {1901972}   & {3220155}   \\
      Exact Solver
        & 2570 & 5268  & 7968  & 9318
        & 3383 & 6402  & 8211  & 10073
        & {---}  & {---}   & {---}   & {---}
        & {---}  & {---}   & {---}   & {---}   \\
      \textbf{Ours}
        & 2647 & 5329 & \textbf{ 8206} & \textbf{ 9601}
        & 3497 & \textbf{ 6511} & \textbf{ 8376} & \textbf{ 10495}
        & {\textbf{ 9276}} & {\textbf{ 14415}} & {\textbf{ 20186}} & {\textbf{ 27699}}
        & {\textbf{ 255789}} & {\textbf{ 658727}} & {\textbf{1258765}} & {\textbf {2199372}} \\
\bottomrule
\end{tabular}
\caption{Performance of \PIMMA~and baselines on four real datasets at different thresholds $T$ under Linear edge weight function setting. The best is highlighted in bold excluding exact solution. For RoadCA and Skitter dataset, ML-based methods can only use Gurobi with heuristic mode.}
\label{tab:performance-real-linear}
\end{table}

\textbf{Solution Quality Evaluation.} On the synthetic dataset, \PIMMA~achieves the lowest total cost in all graph densities, outperforming AT and closely matching the exact solver (Figure \ref{fig:synthetic Solution quality of algorithms}-Linear). This is because \PIMMA~uses various solutions from the PPS algorithm to train $\Omega_{\Theta}$ which learns where the budget should be spent effectively. As a result, \PIMMA~often spends budgets on common edges that affect multiple source-target pairs. In contrast, even the best approximation algorithm in baselines, AT, cannot provide better solution than \PIMMA~as it handles each pair independently, which can lead to spending on edges that do not help other pairs. 

Learning-based methods also exhibit the same behavior, resulting in solutions that are worse than ours. DiffILO spreads budget too widely to fix constraint violations because it relaxes binary decisions into soft probabilistic variables, while Light-MILPopt splits the graph, which breaks global structure and leads to redundant spending. Predict-and-Search depends on Gurobi-generated labels: if these are poor, it cannot refine them due to a narrow search space. For the Quadratic Convex (QC) and Log Concave scenarios, all IP-based ML methods become inapplicable, and even the exact solver (Gurobi) is only able to handle the QC case. Our method continues to outperform the approximation algorithms in both settings. Appendix C.8 shows that \PIMMA with PPS-I consistently achieves higher solution quality and faster runtime compared to ML-based baselines refined with Gurobi.


Across the four real-world networks under thresholds $T \in \{140\%, 180\%, 220\%, 260\%\}$, the advantage of \PIMMA~becomes more evident, especially on larger graphs. As shown in Table \ref{tab:performance-real-linear}, ours consistently achieves the lowest total budget at the highest threshold $T = 260\%$. On large-scale networks where exact solvers are no longer applicable, \PIMMA~achieves the most significant improvements. Specifically, it reduces total cost by 28.1\% on Skitter and 16.8\% on RoadCA compared to the second-best method, DIFFILO. On medium-scale graphs like Gnutella, \PIMMA~outperforms DIFFILO by 4.5\% at the highest threshold, and is only marginally behind it (by 7 units) at the lowest threshold. On the smaller Email network, ours achieves a 1.0\% improvement over DIFFILO at the highest threshold and is only 1\% worse at the lowest. Due to space limit, we encourage readers to Appendix \textbf{C.7} for a more detailed comparison of \PIMMA~against other baselines in the Quadratic Convex and Log Concave cost functions in real dataset.

\begin{figure*}[htp]
    \centering
    \includegraphics[width=1.0\linewidth]{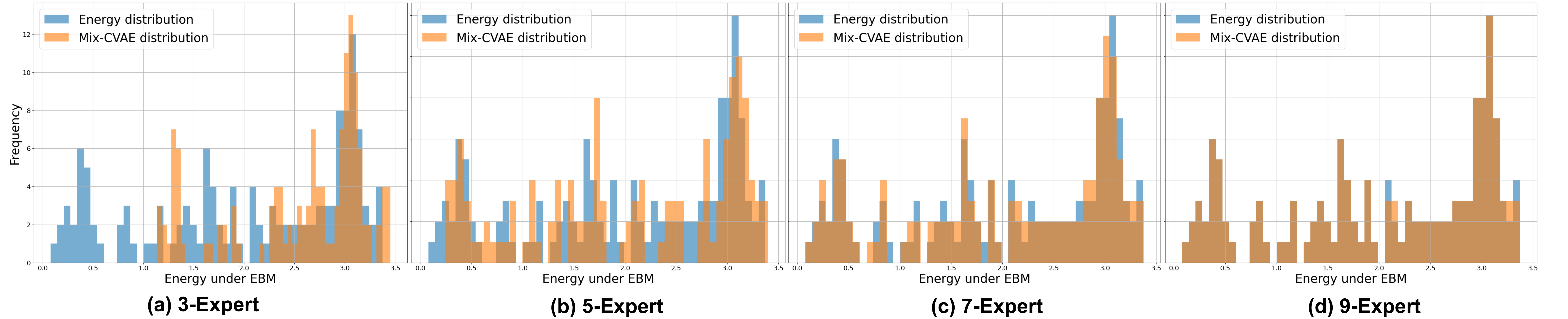}
    \caption{Comparison between the distribution from the EBM and the distribution from the Mix-CVAE with varying numbers of experts on a synthetic dataset with maximum density. Increasing the number of experts improves mode coverage and alignment with the target EBM distribution.}
    \label{fig:Comparison1}
\end{figure*}

\begin{figure*}[htp]
    \centering
    \includegraphics[width=1.0\linewidth]{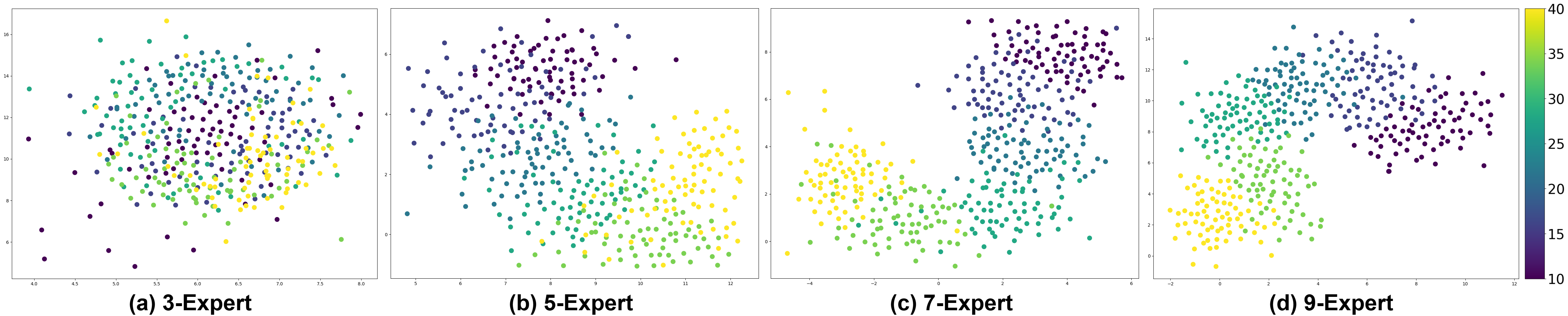} 
    \caption{An example of conditional latent space visualization via UMAP \cite{mcinnes2018umap} for the trained Mix-CVAE on the same synthetic graph and pairs but different thresholds. Points represent latent vectors $\mathbf{z}$, colored by threshold $T$. Clear clustering shows the latent space captures meaningful patterns.}
    \label{fig:latent_space}
\end{figure*}


\textbf{Impact of Expert Addition in Mix-CVAE guided by EBM.} We seek to determine whether Mix-CVAE, when guided by the EBM and augmented with additional experts via our method, can fully capture the distribution induced by the EBM. Figure 4 empirically demonstrates the effectiveness of our expert expansion strategy. As the number of experts in $\Omega$ increases from 3 to 9, the distribution of generated samples (orange curves) increasingly aligns with the energy-based target distribution (blue curves). Notably, the generated distribution progressively captures more modes and exhibits a better approximation of both the shape and spread of the underlying the target EBM. This improvement verifies that expert addition enables Mix-CVAE to overcome mode collapse and better reflect the true multimodal structure of the solution space $p(x \mid c)$. Figure 5 further corroborates this benefit via latent space visualization. Using UMAP projections of latent variables $z$ sampled across different thresholds in the same synthetic network, we observe clear cluster formation corresponding to structural and contextual variations. These clusters become more evident as more experts are added, implying that the augmented latent space encodes richer semantic distinctions. This structured separation is critical for downstream reinforcement learning in Phase 3, where the agent explores the latent space to synthesize stronger attacks. Readers are referred to Appendix \textbf{C} for more ablation experiments.

\section{Conclusion}
In this paper, we introduced \PIMMA, a generative mixture model for assessing network QoS under adversarial budget constraints. By combining predictive path stressing with self-improving generative learning, it captures structural vulnerabilities and generalizes across network scales. Experiments on synthetic and real-world graphs show that our framework outperforms both approximation and learning-based baselines in cost, feasibility, and scalability. These results highlight the potential of generative modeling with structured latent search for scalable network topology optimization.

\section{Discussion}
\label{app:checklist}

\textbf{Limitations.} While the proposed learning-based framework performs well on QoSD problems, it still has some key limitations. In theory, the overall architecture can be extended to other combinatorial optimization tasks on graphs, as long as suitable solvers or estimators are available to support gradient-free training and iterative refinement. However, its success relies on the quality of the initial solutions, which are used to train the generative model. Specifically, the framework depends on having a reasonably good approximation algorithm to produce the initial dataset in Forge. 
In addition, the model’s ability to generalize to unseen network instances is influenced by two important factors. First, it depends on how well the GNN-based estimator (such as SPAGAN) can generalize when predicting shortest-path costs on new graphs. Second, it relies on the quality of the graph representation that is used as input to the conditional generative model. If the estimator or the encoder fails to capture meaningful structural features, the overall performance both in terms of feasibility and cost, may degrade. Future research could improve generalization by incorporating graph-invariant features, training on more diverse graph distributions, or using more expressive graph-based foundation models.

\textbf{Broader Impact.} This work proposes a learning-based framework for solving Quality of Service Degradation (QoSD) problems in large-scale networks, with the potential to influence both algorithmic research and practical applications in infrastructure security, communication planning, and critical network vulnerability assessment. By combining generative modeling, energy-based guidance, and reinforcement learning, the framework offers a scalable alternative to traditional combinatorial solvers, enabling approximate yet effective solutions for computationally intractable tasks. This approach can assist network designers and operators in stress test systems, proactively identifying fragile components, and evaluating resilience under adversarial or high-load scenarios. It also provides a reusable architecture for structured optimization over graphs, which could benefit domains such as transportation, power grids, and supply chain logistics. 

\section{Acknowledgments}

This work was partially supported by National Science Foundation grant IIS-2416606. Any opinions, findings, and conclusions or recommendations expressed in this paper, however, are those of the authors and
do not necessarily reflect the views of the funding agency.





\clearpage

\bibliography{neurips_2025}
\bibliographystyle{unsrt}

\medskip

{
\small

}

\newpage
\section*{NeurIPS Paper Checklist}
\label{app:checklist}


\begin{enumerate}

\item {\bf Claims}
    \item[] Question: Do the main claims made in the abstract and introduction accurately reflect the paper's contributions and scope?
    \item[] Answer: \answerYes{} 
    \item[] Justification: The claims made in the abstract and introduction (Lines 1–18, 67–78) accurately reflect the contributions of the paper: (i) proposing a new generative framework Hephaestus for QoSD with both linear and nonlinear edge weights, (ii) introducing a predictive path-stressing algorithm with provable guarantees, (iii) theoritically modeling solution distributions with an energy-guided mixture of CVAEs, and (iv) refining solutions via latent-space RL.

\item {\bf Limitations}
    \item[] Question: Does the paper discuss the limitations of the work performed by the authors?
    \item[] Answer: \answerYes{} 
    \item[] Justification: Limitations of our framework are discussed in the Limitation Section (Appendix/Supplementary file).

\item {\bf Theory assumptions and proofs}
    \item[] Question: For each theoretical result, does the paper provide the full set of assumptions and a complete (and correct) proof?
    \item[] Answer: \answerYes{} 
    \item[] Justification: The paper provides theoretical guarantees in Theorem 1 (Line 159), Theorem 2 (Line 209), Theorem 3 (Line 223), Lemma 1 (Line 272), and Theorem 4 (Line 274), with explicit assumptions involving constants such as $\delta$, $\varepsilon$, and bounds on energy/log-ratio terms. Complete proofs are deferred to Appendix B. Due to space limitations, additional corollaries, theorems, and lemmas that support and extend the main results are also included in the appendix.

    \item {\bf Experimental result reproducibility}
    \item[] Question: Does the paper fully disclose all the information needed to reproduce the main experimental results of the paper to the extent that it affects the main claims and/or conclusions of the paper (regardless of whether the code and data are provided or not)?
    \item[] Answer: \answerYes{} 
    \item[] Justification: The paper provides sufficient detail to reproduce the main experimental results. Section 4 outlines datasets, baselines, evaluation protocols, and QoSD settings. Appendix C includes ablations, dataset statistics, and details on thresholds, graph generation, and tuning parameters for all baselines.

\item {\bf Open access to data and code}
    \item[] Question: Does the paper provide open access to the data and code, with sufficient instructions to faithfully reproduce the main experimental results, as described in supplemental material?
    \item[] Answer: \answerYes{} 
    \item[] Justification: An anonymized code is in the supplemental material with scripts for training SPAGAN, Mix-CVAE, EBM, and the RL refinement phase, along with instructions and preprocessed data for both synthetic and real datasets.

\item {\bf Experimental setting/details}
    \item[] Question: Does the paper specify all the training and test details (e.g., data splits, hyperparameters, how they were chosen, type of optimizer, etc.) necessary to understand the results?
    \item[] Answer: \answerYes{} 
    \item[] Justification: Section 4 and Appendix C detail dataset, thresholds, sampling strategy, SPAGAN training, CVAE settings, and reinforcement learning parameters.

\item {\bf Experiment statistical significance}
    \item[] Question: Does the paper report error bars suitably and correctly defined or other appropriate information about the statistical significance of the experiments?
    \item[] Answer: \answerNo{} 
    \item[] Justification: While results are consistently benchmarked across multiple thresholds and densities, the paper reports the average performance over 20 independent trials per instance, as stated in Appendix C. However, it does not include error bars, standard deviations, or confidence intervals.

\item {\bf Experiments compute resources}
    \item[] Question: For each experiment, does the paper provide sufficient information on the computer resources (type of compute workers, memory, time of execution) needed to reproduce the experiments?
    \item[] Answer: \answerYes{} 
    \item[] Justification: Section 4 provides timeout settings for baselines (e.g., 3000s–6000s for Gurobi), and Appendix C includes compute specifications and other individual experiment run for training and inference phases.

    
\item {\bf Code of ethics}
    \item[] Question: Does the research conducted in the paper conform, in every respect, with the NeurIPS Code of Ethics \url{https://neurips.cc/public/EthicsGuidelines}?
    \item[] Answer: \answerYes{} 
    \item[] Justification: The research complies with the NeurIPS Code of Ethics. All experiments are conducted on publicly available or synthetic datasets. No human subjects or personally identifiable data were involved.


\item {\bf Broader impacts}
    \item[] Question: Does the paper discuss both potential positive societal impacts and negative societal impacts of the work performed?
    \item[] Answer: \answerYes{} 
    \item[] Justification: The paper explicitly discusses social impacts of the model studied, and the advancement of our solutions. The broader impact statement is in Appendix.
    
\item {\bf Safeguards}
    \item[] Question: Does the paper describe safeguards that have been put in place for responsible release of data or models that have a high risk for misuse (e.g., pretrained language models, image generators, or scraped datasets)?
    \item[] Answer: \answerNA{} 
    \item[] Justification: The models developed are not of high misuse risk (e.g., they are not language or image generators), and datasets used are standard graph datasets or synthetic ones without sensitive content.

\item {\bf Licenses for existing assets}
    \item[] Question: Are the creators or original owners of assets (e.g., code, data, models), used in the paper, properly credited and are the license and terms of use explicitly mentioned and properly respected?
    \item[] Answer: \answerYes{} 
    \item[] Justification: All datasets (e.g., Email, Gnutella, RoadCA) and tools (e.g., Gurobi, SCIP) are properly cited with references [35], [41], [60–63], and their usage complies with open academic licensing terms.

\item {\bf New assets}
    \item[] Question: Are new assets introduced in the paper well documented and is the documentation provided alongside the assets?
    \item[] Answer: \answerYes{} 
    \item[] Justification: They are provided with documentation and instructions in the supplemental material.

\item {\bf Crowdsourcing and research with human subjects}
    \item[] Question: For crowdsourcing experiments and research with human subjects, does the paper include the full text of instructions given to participants and screenshots, if applicable, as well as details about compensation (if any)? 
    \item[] Answer: \answerNA{} 
    \item[] Justification: The research does not involve any crowdsourcing or human subject experiments.

\item {\bf Institutional review board (IRB) approvals or equivalent for research with human subjects}
    \item[] Question: Does the paper describe potential risks incurred by study participants, whether such risks were disclosed to the subjects, and whether Institutional Review Board (IRB) approvals (or an equivalent approval/review based on the requirements of your country or institution) were obtained?
    \item[] Answer: \answerNA{} 
    \item[] Justification: The paper does not involve human subjects; therefore, IRB approval is not applicable.

\item {\bf Declaration of LLM usage}
    \item[] Question: Does the paper describe the usage of LLMs if it is an important, original, or non-standard component of the core methods in this research? Note that if the LLM is used only for writing, editing, or formatting purposes and does not impact the core methodology, scientific rigorousness, or originality of the research, declaration is not required.
    \item[] Answer: \answerNA{} 
    \item[] Justification: No large language models (LLMs) are used in our core methodologies, lemmas and theorems.

\end{enumerate}

\clearpage

\noindent\rule{\textwidth}{1.5pt}  

\begin{center}
    {\LARGE\bfseries Supplementary for Hephaestus: Mixture Generative Modeling with Energy Guidance for Large-scale QoS Degradation}
\end{center}

\noindent\rule{\textwidth}{1.5pt}  

\addcontentsline{toc}{section}{Appendix}

\textbf{This supplementary material serves as the appendix to the main paper to provide full algorithmic pseudocode for all components of the framework, detailed derivations and proofs of key theoretical results, comprehensive ablation studies and hyperparameter settings, as well as expanded discussions on implementation to reproduce results, scalability, and generalization. It is organized as follows:}

\begin{itemize}[leftmargin=*, label=]
    \item \phantomsection \textbf{\ref{algo:algo_details}. \nameref{algo:algo_details}} \dotfill \pageref{algo:algo_details}
    \begin{itemize}
        \item[] \ref{algo:hephaestus_framework}. \nameref{algo:hephaestus_framework} \dotfill \pageref{algo:hephaestus_framework}
        \item[] \ref{algo:forge_phase}. \nameref{algo:forge_phase} \dotfill \pageref{algo:forge_phase}
        \item[] \ref{algo:train_spagan}. \nameref{algo:train_spagan} \dotfill \pageref{algo:train_spagan}
        \item[] \ref{app:predictive_path_stressing}. \nameref{algo:pps} \dotfill 
        \pageref{algo:pps}
        \item[] \ref{algo:morph_phase}. \nameref{algo:morph_phase} \dotfill \pageref{algo:morph_phase}
        \item[] \ref{algo:refine_rl_episode}. \nameref{algo:refine_rl_episode} \dotfill \pageref{algo:refine_rl_episode}
        \item[] \ref{algo:hephaestus_inference}. \nameref{algo:hephaestus_inference} \dotfill \pageref{algo:hephaestus_inference}
        \item[] \ref{algo:pps_i}. \nameref{algo:pps_i} \dotfill \pageref{algo:pps_i}
    \end{itemize}
    \item \phantomsection \textbf{\ref{app:proofs}. \nameref{app:proofs}} \dotfill \pageref{app:proofs}
    \begin{itemize}
        \item[] \ref{app:proof_thm1}. \nameref{app:proof_thm1} \dotfill \pageref{app:proof_thm1}
        \item[] \ref{app:proof_thm2}. \nameref{app:proof_thm2} \dotfill \pageref{app:proof_thm2}
        \item[] \ref{app:proof_thm3}. \nameref{app:proof_thm3} \dotfill \pageref{app:proof_thm3}
        \item[] \ref{app:proof_lemma1}. \nameref{app:proof_lemma1} \dotfill \pageref{app:proof_lemma1}
        \item[] \ref{app:proof_thm4}. \nameref{app:proof_thm4} \dotfill \pageref{app:proof_thm4}
    \end{itemize}
    \item \phantomsection \textbf{\ref{app:experiments}. \nameref{app:experiments}} \dotfill \pageref{app:experiments}
    \begin{itemize}
        \item[] \ref{app:dataset_details}. \nameref{app:dataset_details} \dotfill \pageref{app:dataset_details}
        \item[] \ref{app:hyperparameters}. \nameref{app:hyperparameters} \dotfill \pageref{app:hyperparameters}
        \item[] \ref{app:spagan_validation}. \nameref{app:spagan_validation} \dotfill \pageref{app:spagan_validation}
        \item[] \ref{app:spagan-robust}. \nameref{app:spagan-robust} \dotfill \pageref{app:spagan-robust}
        \item[] \ref{app:weak-data}. \nameref{app:weak-data} \dotfill \pageref{app:weak-data}
        \item[] \ref{app:latent-optimization}. \nameref{app:latent-optimization} \dotfill \pageref{app:latent-optimization}
        \item[] \ref{appendix:ebm_mixcvae_training}. \nameref{appendix:ebm_mixcvae_training}
        \dotfill \pageref{appendix:ebm_mixcvae_training}
        \item[] \ref{appendix:latent_space}. \nameref{appendix:latent_space} \dotfill \pageref{appendix:latent_space}
        \item[] \ref{appendix:expert_addition}. \nameref{appendix:expert_addition}
        \dotfill \pageref{appendix:expert_addition}
        \item[] \ref{app:nonlinear_performance}. \nameref{app:nonlinear_performance} 
        \dotfill \pageref{app:nonlinear_performance}
        \item[] \ref{app:Soundness of the Reward Function}. \nameref{app:Soundness of the Reward Function} 
        \dotfill \pageref{app:Soundness of the Reward Function}
        \item[] \ref{app:Relative optimal gap convergence}. \nameref{app:Relative optimal gap convergence} 
        \dotfill \pageref{app:Relative optimal gap convergence}
        \item[] \ref{app:HEPHAESTUS_pps_at}. \nameref{app:HEPHAESTUS_pps_at} 
        \dotfill \pageref{app:HEPHAESTUS_pps_at}
    \end{itemize}
\end{itemize}

\vspace{1em}
\hrule
\vspace{1em}

\pagebreak

\section{DETAILS OF HEPHAESTUS}
\label{algo:algo_details}
\subsection{Hephaestus Main Framework}
\label{algo:hephaestus_framework}

\SetKwInput{KwInput}{Input}
\SetKwInput{KwOutput}{Output}

\begin{algorithm}[H]
\DontPrintSemicolon
\KwInput{Initial graph dataset $\mathfrak{D}^{\text{graph}}$, set of critical pairs $\mathbf{K}$, set of thresholds $\mathcal{T}$, hyperparameters for all components.}
\KwOutput{Trained \PIMMA model (SPAGAN $\mathfrak{F}_{\boldsymbol{\theta}}$, EBM $q$, Mix-CVAE $\Omega$, RL agent $\pi$), and ability to generate near-optimal solutions $\mathbf{x}^*$ for new QoSD instances.}

\tcp{Initialize storage for solutions}
$\mathfrak{D}^{\text{sol}} \leftarrow \emptyset$ \;

\tcp{--- Phase 1: Forge ---}
$(\mathfrak{F}_{\boldsymbol{\theta}}, \mathfrak{D}^{\text{sol}}_{\text{initial}}) \leftarrow \text{Forge}(\mathfrak{D}^{\text{graph}}, \mathbf{K}, \mathcal{T})$ \;
$\mathfrak{D}^{\text{sol}} \leftarrow \mathfrak{D}^{\text{sol}} \cup \mathfrak{D}^{\text{sol}}_{\text{initial}}$\;

\tcp{--- Phase 2: Morph ---}
$(q, \Omega) \leftarrow \text{Morph}(\mathfrak{D}^{\text{sol}})$ \;

\tcp{--- Phase 3: Refine (Iterative Self-Reinforcement) ---}
\For{each episode $e = 1, \dots, E_{\text{max\_episodes}}$}{
    $\text{Top-K-Solutions} \leftarrow \text{Refine}(q, \Omega, \pi, \mathfrak{D}^{\text{sol}}, \mathfrak{F}_{\boldsymbol{\theta}})$ \;
    $\mathfrak{D}^{\text{sol}} \leftarrow \mathfrak{D}^{\text{sol}} \cup \text{Top-K-Solutions}$ \tcp{Augment solution dataset}
    
    \If{$e \mod{E_{\text{retrain\_freq}}} == 0$}{
        Periodically retrain Morph with augmented data
        $(q, \Omega) \leftarrow \text{Morph}(\mathfrak{D}^{\text{sol}})$ \;
    }
}
$\pi \leftarrow \text{Finalize RL Agent Training from collected experiences}$ \;

\KwRet{$\mathfrak{F}_{\boldsymbol{\theta}}, q, \Omega, \pi$}
\caption{Hephaestus Main Framework}
\end{algorithm}

The pseudocode for the main framework of \PIMMA  encapsulates a full pipeline for solving the QoS Degradation (QoSD) problem through a self-reinforcing generative approach. The algorithm begins by initializing an empty solution set (line 1) and entering the Forge phase (line 2), where it trains a Shortest Path Graph Attention Network (SPAGAN) to approximate shortest-path costs, and then runs the Predictive Path Stressing (PPS) algorithm to generate diverse, feasible perturbation solutions across multiple graphs, thresholds, and critical source-target pairs. These solutions are collected into a pretrained dataset $\mathfrak{D}^{\text{sol}}$ (line 3), which is then assigned as the foundation training set for the next phase, Morph (line 4).

In Morph (line 4), an Energy-Based Model (EBM) is trained to estimate the underlying solution density, and a Mixture of Conditional VAEs (Mix-CVAE) is optimized to match this density, with new experts dynamically added to cover poorly modeled regions. Finally, the Refine phase (starting from line 6) trains an RL agent to explore and optimize in the latent space of the Mix-CVAE, improving solution quality while maintaining feasibility. Furthermore, after each episode, the best performing (highest reward) solutions are added back to $\mathfrak{D}^{\text{sol}}$, and for all $E_{\text{retrain\_freq}}$ episodes, the generative model $(q, \Omega)$ is periodically re-trained on this augmented dataset (lines 10-11), allowing continual improvement and adaptation. The framework outputs all trained components: SPAGAN, EBM, Mix-CVAE, and RL policy—to synthesize near-optimal QoSD solutions on new graphs (line 12).

\subsection{Forge}
\label{algo:forge_phase}

The Forge phase is responsible for constructing an initial dataset of feasible QoSD perturbation solutions using a graph-learned approximation method. It begins by training an SPAGAN, denoted $\mathfrak{F}_{\boldsymbol{\theta}}$, on the provided set of input graphs $\mathfrak{D}^{\text{graph}}$ (Line 2). This model learns to efficiently approximate the shortest-path costs under varying edge perturbations. An empty set $\mathfrak{D}^{\text{sol}}_{\text{new}}$ is initialized to store solutions (Line 3). Then, for every graph $G_i$ in the dataset (Line 4), and for every configuration of critical source-target pairs $\mathcal{K} \in \mathbf{K}$ and thresholds $T \in \mathcal{T}$ (Lines 5–6), the PPS is executed to produce a perturbation vector $\mathbf{x}^{(i)}_{\mathcal{K},T}$ (Line 7). This vector is expected to increase all relevant shortest path costs exceed the threshold $T$, using SPAGAN as an efficient surrogate for shortest path cost estimation. Each resulting solution instance is then added to the solution set $\mathfrak{D}^{\text{sol}}_{\text{new}}$ (Line 8), preserving its associated graph, set of critical pairs, and threshold. Once all iterations are complete, the function returns both the trained SPAGAN model and the newly constructed dataset of feasible perturbations (Line 10), which will serve as the training base for the Morph phase.

\SetKwInput{KwInput}{Input}
\SetKwInput{KwOutput}{Output}

\begin{algorithm}[H]
\Procedure{Forge($\mathfrak{D}^{\text{graph}}, \mathbf{K}, \mathcal{T}$)}{
    $\mathfrak{F}_{\boldsymbol{\theta}} \leftarrow \text{Train\_SPAGAN}(\mathfrak{D}^{\text{graph}})$ \tcp{Train SPAGAN for path cost estimation}
    $\mathfrak{D}^{\text{sol}}_{\text{new}} \leftarrow \emptyset$ \;
    \For{each graph $G_i \in \mathfrak{D}^{\text{graph}}$}{
        \For{each critical pair configuration $\mathcal{K} \in \mathbf{K}$}{
            \For{each threshold $T \in \mathcal{T}$}{
                $\mathbf{x}_{\mathcal{K},T}^{(i)} \leftarrow \text{PPS}(G_i, \mathcal{K}, T, \mathfrak{F}_{\boldsymbol{\theta}}, \{f_e\}, \mathbf{b})$ \;
                $\mathfrak{D}^{\text{sol}}_{\text{new}} \leftarrow \mathfrak{D}^{\text{sol}}_{\text{new}} \cup \{(G_i, \mathcal{K}, T, \mathbf{x}_{\mathcal{K},T}^{(i)})\}$ \;
            }
        }
    }
    \KwRet $(\mathfrak{F}_{\boldsymbol{\theta}}, \mathfrak{D}^{\text{sol}}_{\text{new}})$
}
\caption{Forge}
\end{algorithm}

\subsection{SPAGAN Training}
\label{algo:train_spagan}

This procedure describes the supervised training for SPAGAN, used to approximate shortest-path distances between node pairs in a graph. The function begins by initializing the SPAGAN model $\mathfrak{F}_{\boldsymbol{\theta}}$ (Line 2), which is parameterized to learn over graph-structured input. For a fixed number of training epochs, the model is iteratively updated (starting from line 3). During each epoch, the algorithm samples mini-batches of data—each consisting of a graph $G$, a source-target pair $(s, t)$, and the corresponding ground-truth shortest-path distance $\text{true\_dist}$, typically computed via Dijkstra’s algorithm (Line 5). For each instance, the model predicts the baseline shortest-path cost $\widehat{d}_{s,t}$ using the sub-graph (Line 6). The prediction error is measured using the Huber loss function, which provides robustness to outliers and smooth gradients (Line 7). The model parameters are then updated using backpropagation (Line 8), allowing the network to gradually learn a transferable representation of graph structure and path dynamics. Once training converges, the fully trained SPAGAN model $\mathfrak{F}_{\boldsymbol{\theta}}$ is returned (Line 9) to be used in Forge for shortest path estimation.

\SetKwInput{KwInput}{Input}
\SetKwInput{KwOutput}{Output}

\begin{algorithm}[H]
\Procedure{Train\_SPAGAN($\mathfrak{D}^{\text{graph}}$)}{
    Initialize SPAGAN model $\mathfrak{F}_{\boldsymbol{\theta}}$ \;
    $\mathfrak{D}^{\text{subgraph}} \leftarrow \text{ExtractSubgraphs}(\mathfrak{D}^{\text{graph}})$ \tcp*{Generate training subgraphs from full graphs}
    \For{each training epoch}{
        \For{each batch $(G_{\text{sub}}, s, t, \text{true\_dist})$ from $\mathfrak{D}^{\text{subgraph}}$}{
            $\widehat{d}_{s,t} \leftarrow \mathfrak{F}_{\boldsymbol{\theta}}(G_{\text{sub}}, s, t, \mathbf{0})$ \tcp{Predict baseline shortest path on subgraph}
            $\text{loss} \leftarrow \text{HuberLoss}(\widehat{d}_{s,t}, \text{true\_dist})$ \;
            Update $\mathfrak{F}_{\boldsymbol{\theta}}$ parameters using backpropagation \;
        }
    }
    \KwRet Trained $\mathfrak{F}_{\boldsymbol{\theta}}$
}
\caption{SPAGAN Training with Subgraph Sampling}
\end{algorithm}

\subsection{Predictive Path Stressing (PPS)}
\label{app:predictive_path_stressing}

Solving the QoS Degradation (QoSD) problem is computationally challenging due to the exponential number of feasible paths and the non-submodular nature of the objective function under nonlinear edge costs. To mitigate these challenges, PPS avoids enumerating all feasible paths and instead incrementally constructs a feasible perturbation vector $\mathbf{x} \in \mathbb{N}^{|E|}$, using shortest-path predictions from a pretrained SPAGAN model $\mathfrak{F}_{\boldsymbol{\theta}}$. Unlike exact methods, PPS relies entirely on SPAGAN to both determine shortest paths and estimate their costs under perturbation. The algorithm starts from an initial perturbation $\mathbf{x}_{\text{initial}}$ (Line 2), and constructs a set $\mathcal{K}_{\text{violate}}$ of source-target pairs whose predicted path costs given by $\mathfrak{F}_{\boldsymbol{\theta}}(G, s, t; \mathbf{x})$—fall below threshold $T$ (Line 3). For each violating pair, it obtains the corresponding shortest path $\rho_{s,t}$ using the SPAGAN shortest-path predictor (Lines 6–8). A soft potential function is defined as:
$\mathcal{C}(P, \mathbf{x}) = \sum_{\rho_{s,t} \in P} \min(T, \mathfrak{F}_{\boldsymbol{\theta}}(G, s, t; \mathbf{x}))
$, which serves as a proxy to measure gap to feasibility (Lines 13–14). While $\mathcal{C}(P, \mathbf{x})$ remains less than the relaxed threshold $|P| \cdot T - \bar{\epsilon}$, the algorithm performs updates by selecting the edge $e^\ast$ and increment $\Delta^\ast$ that lead to the greatest increase in potential per budget unit (Lines 10–24). This is computed by evaluating each candidate update $\mathbf{x}'$ and recomputing the potential function $\mathcal{C}(P, \mathbf{x}')$ using SPAGAN predictions (Lines 17–20), without requiring explicit path enumeration or cost function evaluations. Once the optimal update is applied (Line 24), the algorithm checks whether the potential function $\mathcal{C}(P, \mathbf{x})$ has exceeded the soft feasibility threshold $|P| \cdot T - \bar{\epsilon}$. If so, the set of violating pairs is refreshed by recomputing their shortest paths and corresponding predicted costs using SPAGAN (Line 25). The process repeats until all pairs satisfy the constraint $\mathfrak{F}_{\boldsymbol{\theta}}(G, s, t; \mathbf{x}) \ge T$, at which point the final perturbation vector $\mathbf{x}$ is returned (Line 27).

\begin{algorithm}[H]
\Procedure{PPS($G = (V, E), \mathcal{K}, T, \{f_e\}, \mathbf{b}, \mathbf{x}_{\text{initial}}, \mathfrak{F}_{\boldsymbol{\theta}}$)}{
    \KwInput{Graph $G = (V, E)$, target pairs $\mathcal{K}$, threshold $T$, edge cost functions $\{f_e\}$, budget box $\mathbf{b}$, initial vector $\mathbf{x}_{\text{initial}}$, trained SPAGAN $\mathfrak{F}_{\boldsymbol{\theta}}$}
    \KwOutput{Feasible adversarial budget vector $\mathbf{x}$ such that estimated path cost $\sum_{e \in \rho_{s,t}} f_e(x_e) \ge T$ for all $(s,t) \in \mathcal{K}$}

    $\mathbf{x} \leftarrow \mathbf{x}_{\text{initial}}$ \;
    $\mathcal{K}_{\text{violate}} \leftarrow \{(s, t) \in \mathcal{K} \mid \sum_{e \in \rho_{s,t}} f_e(x_e) < T,\ \rho_{s,t} = \texttt{SPAGANPath}(\mathfrak{F}_{\boldsymbol{\theta}},G, s, t; \mathbf{x}) \}$ \;

    \While{$\mathcal{K}_{\text{violate}} \neq \emptyset$}{
        $P \leftarrow \emptyset$ \;
        \ForEach{$(s, t) \in \mathcal{K}_{\text{violate}}$}{
            $\rho_{s,t} \leftarrow \texttt{SPAGANPath}(\mathfrak{F}_{\boldsymbol{\theta}},G, s, t; \mathbf{x})$ \;
            $P \leftarrow P \cup \{\rho_{s,t}\}$ \;
        }

        \tcp{Evaluate soft potential function}
        $\mathcal{C}(P, \mathbf{x}) \leftarrow 0$ \;
        \While{$\mathcal{C}(P, \mathbf{x}) < |P| \cdot T - \bar{\epsilon}$}{
            
            $(e^\ast, \Delta^\ast, \delta_{\max}) \leftarrow (\text{None}, \text{None}, -\infty)$ \;
            $\mathcal{E}_P \leftarrow \bigcup_{\rho \in P} \rho$ \;
            \ForEach{$\rho_{s,t} \in P$}{
                $\mathcal{C}(P, \mathbf{x}) \leftarrow \mathcal{C}(P, \mathbf{x}) + \min(T, \mathfrak{F}_{\boldsymbol{\theta}}(G, s, t, \mathbf{x}))$ \;
            }
            \ForEach{$e \in \mathcal{E}_P$}{
                \For{$\Delta = 1$ \KwTo $b_e - x_e$}{
                    $\mathbf{x}' \leftarrow \mathbf{x} + \Delta \cdot \mathbf{1}_e$ \;
                    $\mathcal{C}(P, \mathbf{x}') \leftarrow 0$ \;
                    \ForEach{$\rho_{s,t} \in P$}{
                        $\mathcal{C}(P, \mathbf{x}') \leftarrow \mathcal{C}(P, \mathbf{x}') + \min(T, \mathfrak{F}_{\boldsymbol{\theta}}(G, s, t, \mathbf{x}'))$
                    }
                    $\delta \leftarrow \frac{\mathcal{C}(P, \mathbf{x}') - \mathcal{C}(P, \mathbf{x})}{\Delta}$ \;
                    \If{$\delta > \delta_{\max}$}{
                        $(e^\ast, \Delta^\ast, \delta_{\max}) \leftarrow (e, \Delta, \delta)$ \;
                    }
                }
            }

            \tcp{Apply optimal update}
            $\mathbf{x} \leftarrow \mathbf{x} + \Delta^\ast \cdot \mathbf{1}_{e^\ast}$ \;
        }

        \tcp{Update violating pairs using SPAGAN}
        $\mathcal{K}_{\text{violate}} \leftarrow \{(s, t) \in \mathcal{K} \mid \sum_{e \in \rho_{s,t}} f_e(x_e) < T,\ \rho_{s,t} = \texttt{SPAGANPath}(G, s, t; \mathbf{x}) \}$ \;
    }

    \KwRet{$\mathbf{x}$}
}
\caption{Predictive Path Stressing (PPS)}
\label{algo:pps}
\end{algorithm}

\subsection{Morph}
\label{algo:morph_phase}

The Morph phase is designed to model the distribution of high-quality QoSD solutions using a generative framework guided by energy-based learning. Intuitively, the goal is to make the Energy-Based Model (EBM) $q_\theta$ approximate the true—but unknown—solution distribution as closely as possible. To do this, the EBM is trained to assign \textit{low energy} (i.e., high likelihood) to real solutions $\mathbf{x}_{\text{real}} \sim \mathfrak{D}^{\text{sol}}$, effectively pulling its density toward regions with meaningful feasible solutions. At the same time, it is encouraged to assign \textit{high energy} (i.e., low likelihood) to generated (fake) samples $\mathbf{x}_{\text{fake}} \sim \Omega$, which come from the current generative model $\Omega$. This adversarial learning setup drives the EBM away from areas the generator covers poorly, creating a pressure that helps both models evolve: the EBM becomes more selective, and the generator learns to cover harder regions.

\SetKwInput{KwInput}{Input}
\SetKwInput{KwOutput}{Output}

\begin{algorithm}[H]
\Procedure{Morph($\mathfrak{D}^{\text{sol}}$)}{
    Initialize EBM $q_\theta$ and Mix-CVAE $\Omega = [\Omega_0, \dots, \Omega_N]$ (initially $N=0$ for first expert) \;
    \For{each minimax training iteration $k = 1, \dots, K_{\text{max\_morph}}$}{
        \tcp{Update EBM $q_\theta$}
        Sample $(\mathbf{x}_{\text{real}}, \mathbf{c})$ from $\mathfrak{D}^{\text{sol}}$ \;
        Sample $\mathbf{z}_{\text{fake}} \sim \tilde{p}_\phi(\mathbf{z} | \mathbf{c})$ (from any expert $\Omega_i$)\; 
        $\mathbf{x}_{\text{fake}} \leftarrow \mathcal{M}_\phi(\mathbf{z}_{\text{fake}}, \mathbf{c})$ (from current $\Omega$'s decoder $\mathcal{M}_\phi$);\\
        $L_{q} \leftarrow \mathbb{E}_{(\mathbf{x}_{\text{real}}, \mathbf{c}) \sim \mathfrak{D}^{\text{sol}}}[E_\theta(\mathbf{x}_{\text{real}})] - \mathbb{E}_{\mathbf{x}_{\text{fake}} \sim \Omega}[E_\theta(\mathbf{x}_{\text{fake}})] + \gamma (\mathbb{E}[E_\theta(\mathbf{x}_{\text{real}})^2] + \mathbb{E}[E_\theta(\mathbf{x}_{\text{fake}})^2])$;
        Update parameters $\theta$ of EBM to minimize $L_{q}$;

        \tcp{Update Mix-CVAE $\Omega$}
        Sample $(\mathbf{x}_{\text{real}}, \mathbf{c})$ from $\mathfrak{D}^{\text{sol}}$ \;
        $L_{\Omega} \leftarrow 0$\;
        \For{each expert $\Omega_i = (\mathcal{P}_\psi^{(i)}, \mathcal{M}_\phi^{(i)}) \in \Omega$}{
             $\mathbf{z}_{\text{encoded}} \leftarrow \mathcal{P}_\psi^{(i)}(\mathbf{x}_{\text{real}}, \mathbf{c})$ \;
             $L^{ELBO}_{\Omega_i} \leftarrow \mathbb{E}_{\mathbf{z} \sim \tilde{q}_{\psi^{(i)}}(\mathbf{z} | \mathbf{x}_{\text{real}}, \mathbf{c})}[\log \tilde{p}_{\phi^{(i)}}(\mathbf{x}_{\text{real}} \mid \mathbf{z}, \mathbf{c})] - \mathrm{KL}[\tilde{q}_{\psi^{(i)}}(\mathbf{z} \mid \mathbf{x}_{\text{real}}, \mathbf{c}) \,\|\, \tilde{p}_{\phi^{(i)}}(\mathbf{z} \mid \mathbf{c})]$ 
             
             Sample $\mathbf{z}_{\text{prior}} \sim \tilde{p}_{\phi^{(i)}}(\mathbf{z} \mid \mathbf{c})$ \;
             $\mathbf{x}_{\text{generated}} \leftarrow \mathcal{M}_\phi^{(i)}(\mathbf{z}_{\text{prior}}, \mathbf{c})$ \;
             $L^{guide}_{\Omega_i} \leftarrow L^{ELBO}_{\Omega_i} + \lambda \cdot E_\theta(\mathbf{x}_{\text{generated}})$ \tcp{Penalize high energy generations}
             $L_{\Omega} \leftarrow L_{\Omega} + L^{guide}_{\Omega_i}$ \tcp{Potentially use gating weights here}
        }
        Update Mix-CVAE parameters $\psi, \phi$ for all experts to minimize $L_{\Omega}$ \;

        \tcp{Expert Addition Strategy}
        \If{$k \pmod{K_{\text{check\_expert}}} == 0$}{
            Sample $\mathbf{x}_{\text{check}}$ from $\mathfrak{D}^{\text{sol}}$ or generate from current $\Omega$\;
            $\chi(\mathbf{x}_{\text{check}}) \leftarrow \log (q_\theta(\mathbf{x}_{\text{check}}) / \Omega(\mathbf{x}_{\text{check}} | \mathbf{c}_{\text{check}}))$ \tcp{Density ratio}
            \If{$\chi(\mathbf{x}_{\text{check}}) > \delta_{\text{expert\_add}}$ and current number of experts $< N_{\text{max}}$}{
                Add a new CVAE expert $\Omega_{N+1}$ to $\Omega$ \;
                Initialize/Train $\Omega_{N+1}$ (e.g., focused on data from regions where $\chi > \delta_{\text{expert\_add}}$ or re-train mixture) \;
                $N \leftarrow N+1$\;
            }
        }
    }
    \KwRet (Trained $q_\theta$, Trained $\Omega$)
}
\caption{Morph}
\end{algorithm}

Formally, the algorithm starts by initializing both the EBM $q_\theta$ and a mixture of Conditional VAEs $\Omega = [\Omega_0, \dots, \Omega_N]$, starting with one single expert (Line 2). Each minimax training iteration proceeds in two stages. First, the EBM is updated by contrasting energy scores between real samples and generated ones. Fake samples are produced by sampling a latent vector $\mathbf{z}_{\text{fake}} \sim \tilde{p}_\phi(\mathbf{z} \mid \mathbf{c})$ from the prior of any expert and decoding it via decoder $\mathcal{M}_\phi$ of $\Omega$ (Lines 4–6). The EBM loss pushes energy lower on real samples and higher on generated ones, with variance-based regularization to stabilize training (Line 7), and the parameters $\theta$ are updated accordingly (Line 8).

Next, the generative model $\Omega$ is updated (Lines 10–17). For each expert $\Omega_i$, the encoder maps real inputs to latent space, and the decoder reconstructs the solution. The training objective combines the standard ELBO which promotes good reconstruction and posterior–prior alignment—with an energy penalty term (Line 15). This penalty uses the EBM to discourage high-energy generations, i.e., samples that lie in unrealistic or undersampled regions. The total loss $L_\Omega$ aggregates across all experts and is minimized to improve the generative model’s coverage of low-energy regions (Line 16). To adaptively expand model capacity, the algorithm includes a periodic \textit{expert addition strategy} (Lines 21–28). Every $K_{\text{check\_expert}}$ steps, it evaluates whether the current mixture $\Omega$ underfits any region of the solution space using a density ratio test: $\chi(\mathbf{x}) = \log (q_\theta(\mathbf{x}) / \Omega(\mathbf{x} \mid \mathbf{c}))$. If this score exceeds a threshold $\delta_{\text{expert\_add}}$, indicating insufficient generative density, a new CVAE expert is added and trained specifically on that difficult region. This enables the generator to incrementally cover diverse and potentially multimodal distributions. The process continues until the maximum number of experts is reached or training ends. Finally, Morph returns both the trained EBM and the mixture-based generative model.

\subsection{Refine}
\label{algo:refine_rl_episode}

\SetKwInput{KwInput}{Input}
\SetKwInput{KwOutput}{Output}

\begin{algorithm}[H]
\Procedure{Refine($q, \Omega, \pi, \mathfrak{D}^{\text{sol}}, \mathfrak{F}_{\boldsymbol{\theta}}$)}{
    \KwInput{Energy model $q$, generative model $\Omega$, RL policy $\pi$, solution dataset $\mathfrak{D}^{\text{sol}}$, SPAGAN estimator $\mathfrak{F}_{\boldsymbol{\theta}}$}
    \KwOutput{Top-K refined feasible solutions with low cost}

    Initialize $\bar{\mathcal{S}}_{\text{new}} \leftarrow \emptyset$ \tcp{Storage for solutions generated in this episode}

    \For{each RL training step $s = 1, \dots, S_{\text{max}}$}{
        Sample instance $(G, \mathcal{K}, T, \mathbf{x})$ from $\mathfrak{D}^{\text{sol}}$ \;
        $\mathbf{c} \leftarrow [G, \mathcal{K}, T]$ \tcp{Context input}
        $\mathbf{z}_{\text{current}} \leftarrow \mathcal{P}_\psi(\mathbf{x}, \mathbf{c})$ \tcp{Encode current solution into latent space}

        \For{each step $t = 1, \dots, \mathfrak{T}_{\text{max}}$ }{
            $(\mu_t, \sigma_t) \leftarrow \pi(\text{state}(\mathbf{z}_{\text{current}}, \mathbf{c}))$ \tcp{Sample action}
            $\epsilon \sim \mathcal{N}(0, I)$ \;
            $\delta_t \leftarrow \mu_t + \sigma_t \cdot \epsilon$ \tcp{Perturb latent}
            $\mathbf{z}_{\text{next}} \leftarrow \mathbf{z}_{\text{current}} + \delta_t$ \;

            $\hat{\mathbf{x}} \leftarrow \mathcal{M}_\phi(\mathbf{z}_{\text{next}}, \mathbf{c})$ \tcp{Decode to solution}
            $\bar{\mathbf{x}} \leftarrow \log(1 + e^{\hat{\mathbf{x}}})$ \tcp{Soft transform for reward}

            \tcp{Evaluate feasibility score via SPAGAN approximation}
            $f\text{score} \leftarrow \sum_{(s_p, t_p) \in \mathcal{K}} \frac{1}{1 + \exp(-\zeta (\mathfrak{F}_{\boldsymbol{\theta}}(G, s_p, t_p; \text{round}(\hat{\mathbf{x}})) - T))}$

            \tcp{Compute reward (Eq. 6)}
            $R_t \leftarrow f\text{score} - \varkappa \cdot \log(1 + \|\bar{\mathbf{x}}\|_1)$

            Store transition $((\mathbf{z}_{\text{current}}, \mathbf{c}), (\mu_t, \sigma_t), R_t, (\mathbf{z}_{\text{next}}, \mathbf{c}))$ in replay buffer \;
            Update policy $\pi$ using replay buffer (e.g., PPO, DDPG, gradient-ascent, etc.) \;
            $\mathbf{z}_{\text{current}} \leftarrow \mathbf{z}_{\text{next}}$ \;

            \If{$R_t \geq R_{\text{thresh}}$ or $t = \mathfrak{T}_{\text{max}}$}{
                $\mathbf{x}_{\text{refined}} \leftarrow \text{PPS-I}(G, \mathcal{K}, T, \hat{\mathbf{x}})$ \tcp{Ensure 100\% feasibility}
                $\bar{\mathcal{S}}_{\text{new}} \leftarrow \bar{\mathcal{S}}_{\text{new}} \cup \{(G, \mathcal{K}, T, \mathbf{x}_{\text{refined}})\}$ \;
                \textbf{Break}
            }
        }
    }

    \tcp{Select best K solutions based on true cost}
    Sort $\bar{\mathcal{S}}_{\text{new}}$ in ascending order of $\|\mathbf{x}_{\text{refined}}\|_1$ \;
    Select top $K$ entries as $\bar{\mathcal{S}}_{\text{topK}}$ \;
    \KwRet $\bar{\mathcal{S}}_{\text{topK}}$
}
\caption{Refine}
\end{algorithm}

This algorithm implements a single episode of the Refine phase, which performs latent-space optimization via reinforcement learning to improve solution quality for the QoSD problem. The goal is to generate low-cost, feasible perturbation vectors by guiding a latent policy network using a differentiable reward structure. The algorithm begins by initializing an empty buffer $\bar{\mathcal{S}}_{\text{new}}$ to store high-quality solutions generated during the episode (Line 2). At each RL training step (Line 3), an instance $(G, \mathcal{K}, T, \mathbf{x})$ is sampled from the solution dataset $\mathfrak{D}^{\text{sol}}$, and the corresponding context vector $\mathbf{c} = [G, \mathcal{K}, T]$ is constructed (Line 5). The current solution $\mathbf{x}$ is encoded into latent space via the encoder $\mathcal{P}_\psi$ of any CVAE expert $\Omega_i \in \Omega$ to yield $\mathbf{z}_{\text{current}}$ (Line 6). An RL trajectory is then simulated over $\mathfrak{T}_{\text{max}}$ steps (Line 7). At each step $t$, the RL agent samples an action $a_t = (\mu_t, \sigma_t)$ from the policy network given the current state (Line 8), and applies a stochastic perturbation to the latent vector using Gaussian noise $\epsilon \sim \mathcal{N}(0, I)$ (Lines 9–10). The next latent state $\mathbf{z}_{\text{next}}$ is computed and decoded into a candidate perturbation vector $\hat{\mathbf{x}}$ using the decoder $\mathcal{M}_\phi$ (Lines 11–12). A soft transformation $\bar{\mathbf{x}} = \log(1 + e^{\hat{\mathbf{x}}})$ is then applied to allow stable reward evaluation (Line 13). To assess the quality of the decoded solution, a feasibility score is computed based on SPAGAN’s predictions of shortest path costs for each critical pair (Line 14), followed by a differentiable reward $R_t$ that penalizes excessive cost via a logarithmic budget term (Line 16). The resulting transition is stored in a replay buffer, and the RL policy is updated using any standard algorithm such as PPO, DDPG, or curiosity-driven exploration (Lines 17). The latent vector is then updated for the next step (Line 18). If the reward exceeds a predefined threshold $R_{\text{thresh}}$ or the trajectory reaches the time limit (Line 19), the decoded solution is passed through PPS-I (Line 20) to continue refine and ensure exact feasibility. The resulting solution is stored in $\bar{\mathcal{S}}_{\text{new}}$ (Line 21), and the trajectory is terminated (Line 22). After all episodes are completed, the algorithm ranks the refined solutions by their true cost $\|\mathbf{x}_{\text{refined}}\|_1$ (Line 23), selects the top-K best ones (Line 24), and returns them as the output of the refinement episode (Line 25). These solutions are later fed back into the self-reinforcement loop, improving both the generative model and the energy function in subsequent iterations.

\subsection{Inference Process}
\label{algo:hephaestus_inference}

\SetKwInput{KwInput}{Input}
\SetKwInput{KwOutput}{Output}

\begin{algorithm}[H]
\Procedure{Inference($G_{\text{new}}, \mathcal{K}_{\text{new}}, T_{\text{new}}, \Omega, \pi, \mathfrak{F}_{\boldsymbol{\theta}}$)}{
    \KwInput{New instance $(G_{\text{new}}, \mathcal{K}_{\text{new}}, T_{\text{new}})$, trained $\Omega, \pi, \mathfrak{F}_{\boldsymbol{\theta}}$}
    \KwOutput{Near-optimal feasible solution $\mathbf{x}^*_{\text{final}}$}
    
    $\mathbf{c}_{\text{new}} \leftarrow [G_{\text{new}}, \mathcal{K}_{\text{new}}, T_{\text{new}}]$ \;
    Sample initial latent vector $\mathbf{z}_{\text{init}}$ (e.g., from $\Omega$'s prior $\tilde{p}_\phi(\mathbf{z} | \mathbf{c}_{\text{new}})$ or encode a heuristic solution) \;
    $\mathbf{z}^* \leftarrow \mathbf{z}_{\text{init}}$ \;
    
    \tcp{RL agent refines latent vector for the new instance}
    \For{$k=1, \dots, K_{\text{inference\_steps}}$}{
        Action $a = (\mu, \sigma) \leftarrow \pi(\text{state}(\mathbf{z}^*, \mathbf{c}_{\text{new}}))$ \tcp{RL policy acts on current latent state}
        $\epsilon_{\text{noise}} \sim \mathcal{N}(0, I)$ \tcp{Exploration or deterministic if $\sigma$ is small}
        $\delta \leftarrow \sigma \cdot \epsilon_{\text{noise}} + \mu$ \;
        $\mathbf{z}^* \leftarrow \mathbf{z}^* + \delta$ \;
    }
    
    $\mathbf{x}_{\text{raw}} \leftarrow \mathcal{M}_\phi(\mathbf{z}^*, \mathbf{c}_{\text{new}})$ (Decoder from $\Omega$) \;
    $\mathbf{x}^*_{\text{final}} \leftarrow \text{PPS-I}(G_{\text{new}}, \mathcal{K}_{\text{new}}, T_{\text{new}}, \mathbf{x}_{\text{raw}})$ \tcp{Final refinement for feasibility}
    \KwRet $\mathbf{x}^*_{\text{final}}$
}
\caption{Inference Process}
\end{algorithm}

In Algorithm 7, the inference process of the Hephaestus framework aims to generate near-optimal, feasible solutions on a new unseen QoSD instance $(G_{\text{new}}, \mathcal{K}_{\text{new}}, T_{\text{new}})$ using the trained models $\Omega$, $\pi$, and $\mathfrak{F}_{\boldsymbol{\theta}}$. The goal is to leverage the learned latent-space generator and RL policy to efficiently synthesize a high-quality perturbation vector without needing to re-run the full training pipeline. The process begins by constructing the context vector $\mathbf{c}_{\text{new}} = [G_{\text{new}}, \mathcal{K}_{\text{new}}, T_{\text{new}}]$ (Line 2). An initial latent vector $\mathbf{z}_{\text{init}}$ is then obtained either by sampling from the prior of the trained Mix-CVAE $\tilde{p}_\phi(\mathbf{z} \mid \mathbf{c}_{\text{new}})$ (Line 3). This serves as the starting point for iterative improvement. The current latent solution is set to $\mathbf{z}^* = \mathbf{z}_{\text{init}}$ (Line 4). To refine this latent vector, the trained RL policy $\pi$ is applied iteratively (Lines 5–9). At each step $k$, the policy takes the current latent state and produces an action $a = (\mu, \sigma)$, which defines a mean and uncertainty over latent perturbations (Line 6). A Gaussian perturbation $\delta$ is sampled and applied to the latent vector $\mathbf{z}^*$, gradually steering the solution toward more feasible and lower-cost regions (Lines 9–10). This process continues either for a fixed number of inference steps $K_{\text{inference\_steps}}$ or until the reward (implicitly computed within the policy) stabilizes. After refinement, the latent code $\mathbf{z}^*$ is decoded into a raw perturbation vector $\mathbf{x}_{\text{raw}}$ using the decoder $\mathcal{M}_\phi$ of the generative model $\Omega$ (Line 10). To ensure full feasibility, this vector is passed through the PPS-I post-processing module (Line 11), which guarantees that the final solution $\mathbf{x}^*_{\text{final}}$ satisfies the QoSD constraints. The final output is then returned (Line 12).

\subsection{Predictive Path Stressing - Inference (PPS-I)}
\label{algo:pps_i}

This procedure describes PPS-I, a variant of the Predictive Path Stressing (PPS) algorithm that retains the same iterative update mechanism over shortest-path constraints. However, PPS-I differs from PPS in two key aspects. First, instead of relying on SPAGAN predictions, it uses Dijkstra’s algorithm to compute exact shortest paths $\rho_{s,t}$ and evaluates their true costs using the nonlinear edge functions $f_e(x_e)$, as shown in Lines 7, 8 and 27. This guarantees correctness under arbitrary cost functions but incurs higher computational overhead. Second, PPS-I accepts a non-zero initial perturbation vector $\mathbf{x}_{\text{initial}}$ as input (Line 3), allowing it to continually refine approximate solutions produced by learning-based modules rather than starting from scratch. The rest of the logic—computing soft potential (Lines 13–15), evaluating marginal gain for candidate updates (Lines 16–23), and selecting optimal edge increments (Line 26)—remains structurally similar to PPS. The set of violating pairs is refreshed using recomputed exact shortest paths (Line 27), and the process continues until full feasibility is achieved.

\begin{algorithm}[H]
\Procedure{PPS-I($G = (V, E), \mathcal{K}, T, \{f_e\}, \mathbf{b}, \mathbf{x}_{\text{initial}}$)}{
    \KwInput{Graph $G = (V, E)$, critical pairs $\mathcal{K}$, threshold $T$, edge cost functions $\{f_e\}$, budget bounds $\mathbf{b}$, initial solution $\mathbf{x}_{\text{initial}}$}
    \KwOutput{Feasible budget vector $\mathbf{x}$ such that $\sum_{e \in \rho_{s,t}} f_e(x_e) \ge T$ for all $(s,t) \in \mathcal{K}$}

    $\mathbf{x} \leftarrow \mathbf{x}_{\text{initial}}$ \;
    $\mathcal{K}_{\text{violate}} \leftarrow \{(s, t) \in \mathcal{K} \mid \sum_{e \in \rho_{s,t}} f_e(x_e) < T,\ \rho_{s,t} = \texttt{DijkstraPath}(G, s, t; \mathbf{x}) \}$ \;

    \While{$\mathcal{K}_{\text{violate}} \neq \emptyset$}{
        $P \leftarrow \emptyset$ \tcp{Shortest paths for current violations}
        \ForEach{$(s, t) \in \mathcal{K}_{\text{violate}}$}{
            $\rho_{s,t} \leftarrow \texttt{DijkstraPath}(G, s, t; \mathbf{x})$ \;
            $P \leftarrow P \cup \{\rho_{s,t}\}$ \;
        }

        \tcp{Evaluate potential function}
        $\mathcal{C}(P, \mathbf{x}) \leftarrow 0$ \;

        \While{$\mathcal{C}(P, \mathbf{x}) < |P| \cdot T - \bar{\epsilon}$}{
            $(e^\ast, \Delta^\ast, \delta_{\max}) \leftarrow (\text{None}, \text{None}, -\infty)$ \;
            $\mathcal{E}_P \leftarrow \bigcup_{\rho \in P} \rho$ \;
            \ForEach{$\rho_{s,t} \in P$}{
                        $c_{\rho} \leftarrow \sum_{e \in \rho_{s,t}} f_e(x_e)$ \;
                        $\mathcal{C}(P, \mathbf{x}) \leftarrow \mathcal{C}(P, \mathbf{x}) + \min(T, c_{\rho})$ \;
                    }
            \ForEach{$e \in \mathcal{E}_P$}{
                \For{$\Delta = 1$ \KwTo $b_e - x_e$}{
                    $\mathbf{x}' \leftarrow \mathbf{x} + \Delta \cdot \mathbf{1}_e$ \;
                    $\mathcal{C}(P, \mathbf{x}') \leftarrow 0$ \;

                    \ForEach{$\rho_{s,t} \in P$}{
                        $c'_{\rho} \leftarrow \sum_{e' \in \rho_{s,t}} f_{e'}(x'_{e'})$ \;
                        $\mathcal{C}(P, \mathbf{x}') \leftarrow \mathcal{C}(P, \mathbf{x}') + \min(T, c'_{\rho})$ \;
                    }

                    $\delta \leftarrow \frac{\mathcal{C}(P, \mathbf{x}') - \mathcal{C}(P, \mathbf{x})}{\Delta}$ \;
                    \If{$\delta > \delta_{\max}$}{
                        $(e^\ast, \Delta^\ast, \delta_{\max}) \leftarrow (e, \Delta, \delta)$ \;
                    }
                }
            }

            \tcp{Apply optimal update}
            $\mathbf{x} \leftarrow \mathbf{x} + \Delta^\ast \cdot \mathbf{1}_{e^\ast}$ \;
        }

        \tcp{Update violating pairs}
        $\mathcal{K}_{\text{violate}} \leftarrow \{(s, t) \in \mathcal{K} \mid \sum_{e \in \rho_{s,t}} f_e(x_e) < T,\ \rho_{s,t} = \texttt{DijkstraPath}(G, s, t; \mathbf{x}) \}$ \;
    }

    \KwRet{$\mathbf{x}$}
}
\caption{Predictive Path Stressing - Inference (PPS-I)}
\label{algo:ppsi}
\end{algorithm}

\section{THEOREMS AND PROOFS}
\label{app:proofs}

\subsection{Predictive Path Stressing Algorithm Ratio}
\label{app:proof_thm1}

\textbf{Theorem 1} (PPS Ratio) Let \( h = \lceil T / w_{\min} \rceil \), where \( w_{\min} = \min_{e \in E} w_e \). 
Assume that the set $\E$ is chosen from $E$ such that $\Pr[\E^* \subseteq \E]=\a $, where $\E^*$ is the set of edges of the optimal solution and $\a \in (0, 1)$ is a constant. Given a parameter $\bar{\epsilon}>0$, then running the Predictive Path Stressing algorithm on $\E$ yields a solution \( \x \) such that $\C(P, \x) \geq |P|T-\bar{\epsilon}$ and $\eE[\|\x\|_1] \le \frac{(1+h\ln(n)+\ln T +\ln(1/\bar{\epsilon}))}{\a} \opt $

{\em Proof.}
	Recall that $\mathcal{C}(P,\x) = \sum_{p\in P} \min\Bigl(T,\,\sum_{e\in p} f_e(x_e)\Bigr)$ and thus $\C(P, \cdot)$ is monotone.
	 Denote $\x^i$ is the partial solution after iteration $i$ of the algorithm, $e_i$ is the edge added to the solution in iteration $i$ with $x_{e_i}=j_i$ and $\u(e_i
	, j_i)$ be a vector we add to the solution $\x^i$ in the $i$-th iteration.  
	
	Since the Predictive Path Stressing algorithm selects, in each iteration 
$i$, the edge $e$ that maximizes the marginal gain per unit cost, given by $ \frac{\mathcal{C}(P,\x^i + \u(e,j_i)) - \mathcal{C}(P,\x^i)}{j_i}$.
If $\E^* \subseteq \E$,  this value is at least as large as the average marginal gain per unit cost in the optimal solution 
 given by $\frac{|P| T-\C\left(P, \x^i\right)}{\text { OPT }}$.
Therefore, 	
	given the solution $\x^{i-1}$, we have
	\begin{align}
	\frac{\C(P, \x^{i})-\C(P, \x^{i-1})}{j_i} &=		 \frac{\C(P, \x^{i-1}+\u(e_i, j_i))-\C(P, \x^{i-1})}{j_i}
	\\
	&\geq 
	\frac{|P|T-\C(P, \x^{i-1})}{\opt} 
	\\
	\Longrightarrow \ \ \  \C(P, \x^{i})-\C(P, \x^{i-1})  & \geq \frac{j_i}{\opt} ( |P|T-\C(P, \x^{i-1}).
	\end{align}
	Therefore,
	\begin{align}
	& \eE \Big[\C(P, \x^{i})-\C(P, \x^{i-1}) | \x^{i-1}\Big]
	\\
	&\geq 	\eE\Big[ \Pr[ \E^* \subseteq \E] \cdot \frac{j_i}{\opt}(|P|T-c(P, \x^{i-1})) + (1-\Pr[ \E^* \subseteq \E])\cdot 0  |  \x^{i-1} \Big]
	\\
	&\geq \eE\Big[    \frac{\a j_i}{\opt} (|P|T-c(P, \x^{i-1})) \mid  \x^{i-1} \Big] .
	\end{align}
	By taking expectation over $\x^{i-1}$, we obtain
	\begin{align}
	\eE \Big[\C(P, \x^{i})-\C(P, \x^{i-1}) \Big]
	&\geq \eE\Big[    \frac{\a j_i}{\opt} (|P|T-\C(P, \x^{i-1}))  \Big]  \label{ineq: iteration}.
	\end{align}
	Re-arranging the above inequality gives: 
	\begin{align}
	\eE\Big[	|P|T - \mathcal{C}(P,\x^{i})\Big] &\le \eE\Big[ |P|T - (\mathcal{C}(P,\x^{i-1}) +  \frac{\a j_i}{\opt}\left(|P|T - \mathcal{C}(P,\x^{i-1})\right)) \Big]
	\\
	&\le \eE\Big[  \left(|P|T - \mathcal{C}(P,\x^{i-1})\right) - \frac{\a j_i}{\opt }\left(|P|T - \mathcal{C}(P,\x^{i-1})\right)  \Big]
	\\
	&= \eE\Big[ (1 - \frac{\a j_i}{\opt})\left(|P|T - \mathcal{C}(P,\x^{i-1})\right) \Big].
	\label{ine:last}
	\end{align}
	Let $t$ is the number iteration. By applying the inequality \eqref{ine:last} iteratively over these \(t\) iterations, we obtain the following:
	\begin{align*}
	\eE\Big[	|P|T - \mathcal{C}(P,\x^{t}) \Big] &\le  \eE\Big[ |P|T \prod_{i=1}^{t} \left(1 - \frac{ \a j_i}{\opt}\right) \Big]
	\\
	&\le \eE\Big[ |P|T \prod_{i=1}^{t} \exp\left(- \frac{\a j_i}{\opt}\right) \Big] \quad \text{(using } 1-z \le e^{-z}\text{)}
	\\
	&\le \eE\Big[ |P|T \exp\left(-\frac{\a \sum_{i=1}^{t} j_i}{\opt} \right) \Big] 
	\\
	&\le \eE\Big[ |P|T \exp\left(-\frac{\a \|\x^t\|_1}{\opt}\right) \Big] .
	\end{align*}
	By the terminal condition of the algorithm, $\C(P, \x^t)\geq |P|T-\bar{\epsilon}$ and $\C(P, \x^{t-1})< |P|T-\bar{\epsilon}$ so we have
		\begin{align}
	& \bar{\epsilon} \leq |P|T-\mathcal{C}\left(P, \x^{t-1}\right) \leq |P|T \cdot \exp\left(-\frac{\a \|\x^{t-1}\|}{\opt}\right)
	\\
	\Longleftrightarrow \ \  & |P|T \cdot \exp\left(-\frac{\a \|\x^{t-1}\|_1}{\opt}\right) \geq \bar{\epsilon} 
	\\
	\Longleftrightarrow \ \  &  \exp\left(-\frac{\a \|\x^{t-1}\|_1}{\opt}\right) \geq \frac{\bar{\epsilon} }{|P|T}
	\\
		\Longleftrightarrow \ \  & \frac{\a \|\x^{t-1}\|_1}{\opt} \leq \ln \Big(\frac{|P|T}{\bar{\epsilon}}\Big)
		\\
			\Longleftrightarrow \ \  &  \|\x^{t-1}\|_1 \leq \frac{\opt}{\a} \ln\Big(\frac{|P|T}{\bar{\epsilon}}\Big).
	\end{align}
	Besides, from the inequality~\eqref{ineq: iteration}, we also have (in expectation)
	\begin{align}
	&\mathcal{C}\left(P, \x^{t-1}\right)+\frac{\a j_t}{\opt}\left(|P| T-\mathcal{C}\left(P, \x^{t-1}\right)\right) \leq \mathcal{C}\left(P, \x^t\right)
	\\
	\Longleftrightarrow  \ \ & \frac{\a j_t}{\opt}\left(|P| T-\mathcal{C}\left(P, \x^{t-1}\right)\right) \leq \mathcal{C}\left(P, \x^t\right)-\mathcal{C}\left(P, \x^{t-1}\right)
	\\
	\Longleftrightarrow  \ \ &\a j_t\left(|P| T-\mathcal{C}\left(P, \x^{t-1}\right)\right) \leq \opt\left(\mathcal{C}\left(P, \x^t\right)-\mathcal{C}\left(P, \x^{t-1}\right)\right)
	\\
	\Longrightarrow  \ \ & \a j_t \leq \opt \ \ (\mbox{Since $|P| T-\mathcal{C}\left(P, \x^{t-1}\right) >\mathcal{C}\left(P, \x^t\right)-\mathcal{C}\left(P, \x^{t-1}\right) $})
	\\
		\Longrightarrow  \ \ &  j_t \leq \frac{\opt }{\a}.
	\end{align}
	The set $P$ of feasible paths can be upper bounded in terms of the maximum path length and the number of nodes. In particular, the number of edges of a feasible path is upper-bounded by
	$h=\left\lceil\frac{T}{w_{\text {min}}}\right\rceil$ (since each edge has weight at least $w_{\min }$), the number of feasible paths of the is upper-bounded by $n^h$. We therefore have:
	\begin{align}
	\eE[	\|\x\|_1] = 	\eE[\|\x^{t-1}\|_1]+ \eE[j_t] &\leq \frac{\opt}{\a} + \frac{\opt\ln\Big(\frac{|P|T}{\bar{\epsilon}}\Big)}{\a}
	\\
	&\leq \frac{\opt}{\a} \Big(1+\ln\Big(\frac{|P|T}{\bar{\epsilon}}\Big)\Big)
	\\
	&\leq  \frac{\opt}{\a} \Big(1+\ln\Big(\frac{n^hT}{\bar{\epsilon}}\Big)\Big)
	\\
	&\leq \frac{\opt}{\a}  \Big(1+h\ln(n)+\ln T + \ln(\frac{1}{\bar{\epsilon}})\Big) 
	\end{align}
	which completes the proof. See Remarks 1 and 2 for how our ratio generalizes to both linear and non-linear cases, and how the performance behaves when model $\mathfrak{J}_{\boldsymbol{\theta}}$ accurately estimates the exact value.

\begin{remark}
    (Approximation ratio when $f_i$ is integer-valued).
In Theorem 1, if $f_i$ is integer-valued for $i\in [m]$ and we set $\bar{\epsilon} \in (0, 1)$, then running the Predictive Path Stressing algorithm on $\E$ yields a feasible solution \( \x \), i.e, $\C(P, \x) = |P|T$ such that $\eE[\|\x\|_1] \le \frac{(1+h\ln(n)+\ln T)}{\a} \opt $
\end{remark}

\begin{remark}
    If PPS selectes all edges in the optimal solution, i.e, $\Pr[\E^*\subseteq \E]=1$, then running the Predictive Path Stressing algorithm on $\E$ yields a solution \( \x \) such that $\C(P, \x) \geq |P|T-\bar{\epsilon}$ and $\eE[\|\x\|_1] \le (1+h\ln(n)+\ln T +\ln(1/\bar{\epsilon})) \opt $
\end{remark}

\subsection{Expert Augmentation Efficiency}
\label{app:proof_thm2}

\textbf{Theorem 2 (Expert Augmentation Efficiency)}: Suppose there exists a constant \(\epsilon>0\) such that \(\mathbb{P}_{p(x)}\{\chi(x)>\delta\}\ge \epsilon\); on the region \(\{x: \chi(x)>\delta\}\), a new expert \(\Omega_{N+1}(x)\) is added with gating weight \(\alpha(x)=\mathbf{1}\{\chi(x)>\delta\}a_0\) (\(0<a_0<1\)) and trained to satisfy \(\Omega_{N+1}(x)\ge c\,q(x)\) for some \(c\in(0,1)\); then the updated mixture \(\Omega'(x)=\alpha(x)\,\Omega_{N+1}(x)+(1-\alpha(x))\,\Omega(x)\) satisfies \(\mathrm{KL}(q\|\Omega')\le \mathrm{KL}(q\|\Omega)-\gamma(\delta,\epsilon)\), where \(\gamma(\delta,\epsilon)=a_0(\delta+\log c)\,\epsilon_0>0\) and \(\epsilon_0>0\) is a lower bound on \(\int_{\chi(x)>\delta}q(x)dx\).

{\em Proof.} In the regions where \(\chi(x) = \log \frac{q(x)}{\Omega(x)} \text{ and } \chi(x) > \delta\), we add a new expert \(\Omega_{N+1}(x)\) and extend the gating network so that its output becomes:

\begin{equation}
    w'_i(x), \quad i = 0,1,\dots,N+1, \quad \text{with } \sum_{i=0}^{N+1} w'_i(x)=1.
\end{equation}

Without loss of generality and for analysis purposes, we form a new mixture by taking a convex combination of the old mixture and the new expert:

\begin{equation}
    \Omega'(x) = \alpha(x) \, \Omega_{N+1}(x) + \bigl(1-\alpha(x)\bigr) \, \Omega(x).
\end{equation}

We choose the gating function \(\alpha(x)\) to focus on regions where \(\chi(x) > \delta\). In particular, we set:

\begin{equation}
    \alpha(x) = \mathbf{1}\{\chi(x)>\delta\}\, a(x),
\end{equation}

and for simplicity we take \(a(x)=a_0\) for those \(x\) with \(\chi(x)>\delta\) where $0 < a_0 < 1$. Next, we consider the KL divergence between \(q(x)\) and the original mixture $\Omega(x)$,

\begin{equation}
    \mathrm{KL}(q\|\Omega) = \int q(x) \log \frac{q(x)}{\Omega(x)}\, dx,
\end{equation}

and the KL divergence between \(q(x)\) and the updated mixture $\Omega'(x)$,

\begin{equation}
    \begin{aligned}
    \mathrm{KL}(q\|\Omega') &= \int q(x) \log \frac{q(x)}{\Omega'(x)}\, dx \\
    &= \int q(x) \log \frac{q(x)}{\alpha(x)\,\Omega_{N+1}(x) + (1-\alpha(x))\,\Omega(x)}\, dx\\
    &= \int q(x) \log q(x)- q(x) \log\Bigl[\alpha(x)\,\Omega_{N+1}(x) + (1-\alpha(x))\,\Omega(x)\Bigl]\, dx\\
     &\le \int q(x) \Bigl[\log q(x) - \alpha(x)\,\log \Omega_{N+1}(x) - (1-\alpha(x))\,\log \Omega(x) \Bigr] dx \\
    &= \int q(x) \log q(x)\,dx - \int q(x)\,\alpha(x) \log \Omega_{N+1}(x)\,dx - \int q(x)(1-\alpha(x)) \log \Omega(x)\,dx.
\end{aligned}
\end{equation}

We can achieve the above because the logarithm is concave (and thus \(-\log\) is convex), Jensen’s inequality implies that:

\begin{equation}
    \begin{aligned}
    \log \Omega'(x) &= \log\Bigl(\alpha(x)\,\Omega_{N+1}(x) + \bigl(1-\alpha(x)\bigr)\,\Omega(x)\Bigr)\\ &\ge \alpha(x) \log \Omega_{N+1}(x) + \bigl(1-\alpha(x)\bigr) \log \Omega(x)\\
    \end{aligned}
\end{equation}

\begin{equation}
    \begin{aligned}
        &\Longrightarrow -\log \Omega'(x) \le -\alpha(x) \log \Omega_{N+1}(x) - (1-\alpha(x)) \log \Omega(x).\\
    &\Longrightarrow -\int q(x) \log \Omega'(x)\,dx \le -\int q(x)\,\alpha(x) \log \Omega_{N+1}(x)\,dx - \int q(x)(1-\alpha(x)) \log \Omega(x)\,dx.
    \end{aligned}
\end{equation}

Recall that the KL divergence between \(q\) and \(\Omega\) is defined as:

\begin{equation}
    \begin{aligned}
        \mathrm{KL}(q\|\Omega) &= \int q(x) \log \frac{q(x)}{\Omega(x)}\,dx\\
    &= \int q(x) \log q(x)\,dx - \int q(x) \log \Omega(x)\,dx.
    \end{aligned}  
\end{equation}

Similarly, for the updated mixture \(\Omega'(x)\), we have the following.

\begin{equation}
    \begin{aligned}
   \mathrm{KL}(q\|\Omega') &= \int q(x) \log \frac{q(x)}{\Omega'(x)}\,dx \\
   &= \int q(x) \log q(x)\,dx - \int q(x) \log \Omega'(x)\,dx.
\end{aligned}
\end{equation}

If we subtract the original divergence from the new one, the terms involving \(\int q(x) \log q(x)\,dx\) cancel, leaving:

\begin{equation}
   \begin{aligned}
\mathrm{KL}(q\|\Omega') - \mathrm{KL}(q\|\Omega) &= \left[\int q(x) \log q(x)\,dx - \int q(x) \log \Omega'(x)\,dx\right] \\
&\quad - \left[\int q(x) \log q(x)\,dx - \int q(x) \log \Omega(x)\,dx\right] \\
&= - \int q(x) \log \Omega'(x)\,dx + \int q(x) \log \Omega(x)\,dx.
\end{aligned} 
\end{equation}

\begin{equation}
    \begin{aligned}
\mathrm{KL}(q\|\Omega') - \mathrm{KL}(q\|\Omega) &\le \left[-\int q(x)\,\alpha(x) \log \Omega_{N+1}(x)\,dx - \int q(x)(1-\alpha(x)) \log \Omega(x)\,dx\right] \\
&\quad + \int q(x) \log \Omega(x)\,dx \\
&= -\int q(x)\,\alpha(x) \log \Omega_{N+1}(x)\,dx - \int q(x)(1-\alpha(x)) \log \Omega(x)\,dx \\
&\quad + \int q(x) \log \Omega(x)\,dx \\
&= -\int q(x)\,\alpha(x) \Bigl[\log \Omega_{N+1}(x) - \log \Omega(x)\Bigr]dx.
\end{aligned}
\end{equation}

This completes the derivation. We have shown that the difference in the KL divergence between \(q\) and the updated mixture \(\Omega'\) and that between \(q\) and the original mixture \(\Omega\) is bounded above by:

\begin{equation}
    \mathrm{KL}(q\|\Omega') - \mathrm{KL}(q\|\Omega) \le -\int q(x)\,\alpha(x) \left[\log \Omega_{N+1}(x) - \log \Omega(x)\right]dx.
\end{equation}

Notice that on the region where \(\chi(x)\le \delta\), the indicator in \(\alpha(x)\) is zero, so we only integrate over the region where \(\chi(x)>\delta\). Setting \(\alpha(x)=a_0\) on that region, we obtain:

\begin{equation}
    \mathrm{KL}(q\|\Omega') - \mathrm{KL}(q\|\Omega) \le -a_0 \int_{\chi(x)>\delta} q(x) \log \frac{\Omega_{N+1}(x)}{\Omega(x)} \, dx.
\end{equation}

Now, for any \(x\) with \(\chi(x) > \delta\), we have $\log \frac{q(x)}{\Omega(x)} > \delta$ so that $\frac{q(x)}{\Omega(x)} > \exp(\delta)$. Suppose further that the new expert is designed such that, on this region, $\Omega_{N+1}(x) \ge c\, q(x)$, where \(c \in (0,1)\) is a constant. Then it holds that:

\begin{equation}
  \frac{\Omega_{N+1}(x)}{\Omega(x)} \ge c\, \frac{q(x)}{\Omega(x)} > c\, \exp(\delta).  
\end{equation}

Taking logarithms gives:
\begin{equation}
    \log \frac{\Omega_{N+1}(x)}{\Omega(x)} \ge \delta + \log c.
\end{equation}

Thus, if we let:
\begin{equation}
    \eta = \int_{\chi(x)>\delta} q(x) \, dx,
\end{equation}

and assume (through technical equivalence between \(q(x)\) and \(p(x)\)) that \(\eta \ge \epsilon_0 > 0\), then we obtain:
\[
\int_{\chi(x)>\delta} q(x) \log \frac{\Omega_{N+1}(x)}{\Omega(x)} \, dx \ge (\delta+\log c)\,\epsilon_0.
\]

Therefore, the reduction in KL divergence satisfies:
\[
\mathrm{KL}(q\|\Omega') - \mathrm{KL}(q\|\Omega) \le -a_0\, (\delta+\log c)\,\epsilon_0.
\]

Defining:
\[
\gamma(\delta,\epsilon) = a_0\, (\delta+\log c)\,\epsilon_0,
\]

We conclude that:
\[
\mathrm{KL}(q\|\Omega') \le \mathrm{KL}(q\|\Omega) - \gamma(\delta,\epsilon),
\]

which shows that the addition of the new expert decreases the KL divergence by at least \(\gamma(\delta,\epsilon) > 0\). Thus, this completes our proof.



\textbf{Corollary 3} (Effectiveness of Expert Addition and Gating Retraining).
Suppose that at each iteration, a new expert is added in regions where the discrepancy $\chi(x)>\delta$, and the gating network is retrained using a softmax function that produces a convex combination of expert outputs. The updated mixture then takes the form $\Omega^{(t+1)}(x)=\sum_{i=0}^{N+1} w_i^{\prime}(x) \Omega_i(x)$, where the new gating weights $w_i^{\prime}(x)$ are derived from a softmax layer.

We have that the loss function for the gating network, defined as the KL divergence

$$
\mathcal{L}_g\left(\varsigma_g\right)=\int q(x) \log \frac{q(x)}{\sum_{i=0}^{N+1} w_i^{\prime}\left(x ; \varsigma_g\right) \Omega_i(x)} d x
$$

is $L_g$-smooth and $\mu_g$-strongly convex in a neighborhood of the optimal parameter $\varsigma_g^*$. Then, if we update the gating parameters via gradient descent with a learning rate small enough to ensure convergence, the KL divergence between the target distribution $q(x)$ and the current mixture $\Omega^{(t)}(x)$ decreases by at least a fixed amount $\gamma>0$ at each iteration. That is,

$$
\mathrm{KL}\left(q \| \Omega^{(t+1)}\right) \leq \mathrm{KL}\left(q \| \Omega^{(t)}\right)-\gamma
$$

As a result, the sequence of mixtures $\left\{\Omega^{(t)}(x)\right\}$ converges to $q(x)$, in the sense that

$$
\lim _{t \rightarrow \infty} \mathrm{KL}\left(q \| \Omega^{(t)}\right)=0, \quad \text { so } \quad \Omega^{(t)}(x) \rightarrow q(x) \text { almost everywhere. }
$$

Finally, if $q(x)$ is a good approximation of the true target distribution $p(x)$-for example, because it comes from a properly trained energy-based model-then the mixture also converges to $p(x)$. In the limit, we obtain the desired equilibrium where

$$
\Omega(x)=q(x)=p(x),
$$

showing that the mixture model has successfully learned the target distribution through iterative refinement.

{\em Proof:} After the new expert is added, the gating network is retrained using a softmax layer that produces new weights

\[
w'_i(x) = \frac{\exp(f'_i(x))}{\sum_{j=0}^{N+1} \exp(f'_j(x))}, \quad i=0,\dots,N+1,
\]

so that the updated mixture becomes

\[
\Omega'(x) = \sum_{i=0}^{N+1} w'_i(x)\, \Omega_i(x).
\]

We define the gating network’s loss as the KL divergence

\[
\mathcal{L}_g(\varsigma_g) = \int q(x) \log \frac{q(x)}{\sum_{i=0}^{N+1} w'_i(x;\varsigma_g)\, \Omega_i(x)} \,dx.
\]

Because softmax produces a convex combination, the mapping \(x \mapsto \sum_{i=0}^{N+1} w'_i(x;\varsigma_g)\, \Omega_i(x)\) is convex in the gating weights. Under the assumptions that \(\mathcal{L}_g\) is \(L_g\)-smooth and \(\mu_g\)-strongly convex near the optimum \(\varsigma_g^*\), we update the parameters by gradient descent:

\[
\varsigma_g^{(t+1)} = \varsigma_g^{(t)} - \eta_g \, \nabla_{\varsigma_g} \mathcal{L}_g(\varsigma_g^{(t)}).
\]

By standard results, the error in the gating network decreases as

\[
\|\varsigma_g^{(t+1)} - \varsigma_g^*\|^2 \le \left(1 - \eta_g (2\mu_g - \eta_g L_g^2)\right) \|\varsigma_g^{(t)} - \varsigma_g^*\|^2,
\]

ensuring convergence provided the learning rate \(\eta_g\) is chosen sufficiently small. As the gating network converges, the new mixture \(\Omega'(x) = \sum_{i=0}^{N+1} w'_i(x;\varsigma_g)\, \Omega_i(x)\) becomes an even better approximation of \(q(x)\), reinforcing the reduction in KL divergence we established earlier. If this process of detecting regions where \(\chi(x)>\delta\), adding a new expert, and retraining the gating network is repeated iteratively, we obtain a sequence of mixtures \(\{\Omega^{(t)}(x)\}\) for which

\[
\mathrm{KL}(q\|\Omega^{(t+1)}) \le \mathrm{KL}(q\|\Omega^{(t)}) - \gamma(\delta,\epsilon).
\]

Thus, by repeated application, we ensure

\[
\lim_{t\to\infty} \mathrm{KL}(q\|\Omega^{(t)}) = 0.
\]

Assuming that \(q(x)\) is a good proxy for \(p(x)\) (due to proper training of the EBM), it follows that

\[
\Omega^{(t)}(x) \to q(x) \quad \text{and ultimately } \Omega^{(t)}(x) \to p(x),
\]

establishing the desired Nash equilibrium \(p(x)=q(x)=\Omega(x)\).

\subsection{Normalization Free Function}
\label{app:proof_thm3}

\textbf{Theorem 3 (Normalization Free Function): } 
The objective function $\min_{q}\,\max_{\Omega}\{\mathrm{KL}(p\|q)-\mathrm{KL}(\Omega\|q)\},$ is normalizing free which is independent with \(Z\).

{\em Proof.} By definition, the KL divergence between \(p(x)\) and \(q(x)\) is $   \mathrm{KL}(p\|q)=\int p(x)\log\frac{p(x)}{q(x)}\,dx$. Since $q(x)=\frac{\exp\left(-\frac{E(\mathbf{x})}{\tau}\right)}{Z},$ we can rewrite it as: 
$$
\begin{aligned}
\mathrm{KL}(p\|q) &= \int p(x)\log\frac{p(x)}{q(x)}\,dx \\
&= \int p(x)\Bigl(\log p(x) - \log q(x)\Bigr)\,dx \\
&= \int p(x)\log p(x)\,dx - \int p(x)\log q(x)\,dx \\
&= \int p(x)\log p(x)\,dx - \int p(x)\left[\log \left(\frac{\exp\left(-\frac{E(\mathbf{x})}{\tau}\right)}{Z}\right)\right]dx \\
&= \int p(x)\log p(x)\,dx - \int p(x)\left[-\frac{E(x)}{\tau} - \log Z\right]dx \\
&= \int p(x)\log p(x)\,dx + \frac{1}{\tau}\int p(x)E(x)\,dx + \log Z\int p(x)\,dx \\
&= \int p(x)\log p(x)\,dx + \frac{1}{\tau}\int p(x)E(x)\,dx + \log Z,
\end{aligned}
$$

Similarly, the KL divergence between \(\Omega(x)\) and \(q(x)\) is:
   \[\begin{aligned}
          \mathrm{KL}(\Omega\|q) &= \int \Omega(x)\log\frac{\Omega(x)}{q(x)}\,dx \\
&= \int \Omega(x)\log \Omega(x)\,dx - \int \Omega(x)\log q(x)\,dx \\
&= \int \Omega(x)\log \Omega(x)\,dx - \int \Omega(x)\left[-\frac{E(x)}{\tau}-\log Z\right]dx \\
&= \int \Omega(x)\log \Omega(x)\,dx + \frac{1}{\tau}\int \Omega(x)E(x)\,dx + \log Z\int \Omega(x)\,dx \\,
&= \int \Omega(x)\log \Omega(x)\,dx + \frac{1}{\tau}\int \Omega(x)E(x)\,dx + \log Z
   \end{aligned}
   \]

The difference between the two KL divergences is then:
   \[
   \begin{aligned}
   \mathrm{KL}(p\|q)-\mathrm{KL}(\Omega\|q)
   &= \Biggl[\int p(x)\log p(x)\,dx + \frac{1}{\tau}\int p(x)E(x)\,dx + \log Z\Biggr] \\
   &\quad - \Biggl[\int \Omega(x)\log \Omega(x)\,dx + \frac{1}{\tau}\int \Omega(x)E(x)\,dx + \log Z\Biggr] \\
   &= \int p(x)\log p(x)\,dx - \int \Omega(x)\log \Omega(x)\,dx \\
   &\quad + \frac{1}{\tau}\int p(x)E(x)-\frac{1}{\tau}\int\Omega(x)E(x)\,dx.\\
   &= \underbrace{\int p(x)\log p(x)\,dx - \int \Omega(x)\log \Omega(x)\,dx}_{\textit{Entropy Different}} \\
   &\quad + \frac{1}{\tau}\Bigl\{\mathbb{E}_{p}[E_\theta(x)]-\mathbb{E}_{\Omega}[E_\theta(x)]\Bigr\}\\
   \end{aligned}
   \]
   
Notice that the \(\log Z\) terms cancel, so the difference is independent of \(Z\).

\textbf{Corollary 4} (Reduction to Energy Expectation Difference)
Under the setting of Theorem 3, the minimax objective\[
\min_{\theta}\,\max_{q}\Bigl\{\mathrm{KL}\bigl(p\|q\bigr)\;-\;\mathrm{KL}\bigl(\Omega\|q\bigr)\Bigr\}
\]
is equivalent (up to additive and multiplicative constants independent of \(\theta\)) to
\[
\min_{\theta}\;\Bigl[
\mathbb{E}_{x\sim p}\bigl[E_\theta(x)\bigr]
\;-\;
\mathbb{E}_{x\sim \Omega}\bigl[E_\theta(x)\bigr]
\Bigr].
\]

\textbf{Proof.} By Theorem 3 we have:

\begin{align}
\mathrm{KL}(p\|q)-\mathrm{KL}(\Omega\|q)
&= \int p(x)\log p(x)\,dx - \int p(x)\log q(x)\,dx 
   - \Bigl(\int \Omega(x)\log \Omega(x)\,dx\\ &- \int \Omega(x)\log q(x)\,dx\Bigr)\\
&= \int p(x)\log p(x)\,dx - \int \Omega(x)\log \Omega(x)\,dx
   + \int\bigl[\Omega(x)-p(x)\bigr]\log q(x)\,dx\\
&= \underbrace{\int p(x)\log p(x)\,dx - \int \Omega(x)\log \Omega(x)\,dx}_{C} \\&
   + \int\bigl[p(x)-\Omega(x)\bigr]\biggl(-\frac{1}{\tau}E_\theta(x)-\log Z\biggr)\,dx\\
&= C 
   + \frac{1}{\tau}\int\bigl[p(x)-\Omega(x)\bigr]\,E_\theta(x)\,dx
   \quad\biggl(\text{since }\int\bigl[p(x)-\Omega(x)\bigr]\,dx=0\biggr)\\
&= C 
   + \frac{1}{\tau}\Bigl(\mathbb{E}_{x\sim p}\bigl[E_\theta(x)\bigr]
   - \mathbb{E}_{x\sim \Omega}\bigl[E_\theta(x)\bigr]\Bigr).
\end{align}

Since adding the constant \(C\) and scaling by \(1/\tau\) do not affect the location of the minimizer in \(\theta\), it follows that
\[
\arg\min_{\theta}
\Bigl\{\mathrm{KL}(p\|q)-\mathrm{KL}(\Omega\|q)\Bigr\}
\;=\;
\arg\min_{\theta}
\Bigl\{\mathbb{E}_{p}[E_\theta(x)]-\mathbb{E}_{\Omega}[E_\theta(x)]\Bigr\},
\]
which establishes the claimed reduction.

\subsection{Differentiable Reward Function} 
\label{app:proof_lemma1}

\textbf{Lemma 1} Let $\mathfrak{F}_\theta: \mathcal{X} \rightarrow \mathbb{R}$ be differentiable on an open set $\mathcal{X} \subset \mathbb{R}^m$. The reward function
$
\mathcal{R}(\mathbf{x})$ is differentiable on $\mathcal{X}$.

{\em Proof.} Let $\mathfrak{F}_{\boldsymbol{\theta}}(G, s, t ; \mathbf{x})$ be a neural estimator of shortest-path cost from source node $s$ to target node $t$, parameterized by $\boldsymbol{\theta}$. Define the reward function as:

$$
\mathcal{R}(\mathbf{x})=\underbrace{\sum_{(s, t) \in \mathcal{K}} \pounds\left(\mathfrak{F}_{\boldsymbol{\theta}}(G, s, t ; \mathbf{x})-T\right)}_{\text {Smooth feasibility Term}}-\varkappa \cdot \underbrace{\log \left(1+\|\Upsilon(\mathbf{x})\|_1\right)}_{\text {Soft Cost Penalty Term}}
$$

where $\pounds(z)=\frac{1}{1+e^{-\zeta z}}$ is a sigmoid function with slope parameter $\zeta>0$, $\Upsilon(\mathbf{x})=\log \left(1+e^{\mathbf{x}}\right)$, applied coordinate-wise, and $T>0$ is the feasibility threshold, and $\kappa>0$ is a regularization hyperparameter. We aim to show that $\mathcal{R}(\mathbf{x})$ is continuously differentiable on any open domain $\mathcal{X} \subset \mathbb{R}^m$.  First, the estimator $\mathfrak{F}_{\boldsymbol{\theta}}$ is a deep graph neural network constructed as a composition of $L$ differentiable layers:

$$
\mathfrak{F}_{\boldsymbol{\theta}}=f_L \circ f_{L-1} \circ \cdots \circ f_1$$

where each $f_i$ may involve affine maps, ELU activations, attention mechanisms, and message-passing steps, all of which are differentiable. Hence, by the chain rule, $\mathfrak{F}_{\boldsymbol{\theta}}(G, s, t ; \cdot)$ is differentiable on $\mathcal{X}$. For each $(s, t) \in \mathcal{K}$, define:

$$
\pounds_{s, t}(\mathbf{x}):=\pounds\left(\mathfrak{F}_\theta(G, s, t ; \mathbf{x})-T\right) .
$$

Since $\mathfrak{F}_{\boldsymbol{\theta}}$ is differentiable and $\pounds(z)$ is smooth ( $C^{\infty}$ ) on $\mathbb{R}$ with $\pounds^{\prime}(z)=\zeta \cdot \pounds(z) \cdot(1-\pounds(z))$, it follows from the chain rule that $\pounds_{s, t}(\mathbf{x})$ is differentiable:

$$
\nabla \pounds_{s, t}(\mathbf{x})=\pounds^{\prime}\left(\mathfrak{F}_{\boldsymbol{\theta}}(G, s, t ; \mathbf{x})-T\right) \cdot \nabla \mathfrak{F}_{\boldsymbol{\theta}}(G, s, t ; \mathbf{x}) .
$$

Summing over all $(s, t)$ in $\mathcal{K}$, we obtain:

$$
\digamma(\mathbf{x}):=\sum_{(s, t) \in \mathcal{K}} \pounds_{s, t}(\mathbf{x})
$$

which is a finite sum of differentiable functions and therefore differentiable on $X$. Let $\Upsilon: \mathbb{R}^m \rightarrow \mathbb{R}^m$ be the vector-valued softplus function such as:

$$
\Upsilon(\mathbf{x}):=\left[\log \left(1+e^{x_1}\right), \ldots, \log \left(1+e^{x_m}\right)\right]
$$

where each component $\Upsilon_i\left(x_i\right)$ is differentiable, with:

$$
\Upsilon_i^{\prime}\left(x_i\right)=\frac{e^{x_i}}{1+e^{x_i}}=\pounds\left(x_i\right)
$$

Hence $\Upsilon \in C^{\infty}\left(\mathbb{R}^m\right)$, and so the map $\mathbf{x} \mapsto\|\Upsilon(\mathbf{x})\|_1=\sum_{i=1}^m \Upsilon_i\left(x_i\right)$ is also differentiable as a finite sum of differentiable functions.
Now we consider the scalar penalty $\Lambda(\mathbf{x}):=\log \left(1+\|\Upsilon(\mathbf{x})\|_1\right)$. Because $\Upsilon_i\left(x_i\right)>0$ for all $x_i$, we have $\|\Upsilon(\mathbf{x})\|_1>0$, and $\log (1+u)$ is smooth on $(0, \infty)$. Hence, by the chain rule:

$$
\nabla \Lambda(\mathbf{x})=\frac{1}{1+\|\Upsilon(\mathbf{x})\|_1} \cdot \nabla\left(\sum_{i=1}^m \Upsilon_i\left(x_i\right)\right)
$$

which is well-defined and continuous on $\mathbb{R}^m$. Putting both together that $\digamma(\mathbf{x})=\sum_{(s, t) \in \mathcal{K}} \pounds_{s, t}(\mathbf{x})$ and  $\Lambda(\mathbf{x})=\log \left(1+\|\Upsilon(\mathbf{x})\|_1\right)$ is both differentiable. Therefore, the reward function $\mathcal{R}(\mathbf{x})=\digamma(\mathbf{x})-\varkappa \cdot \Lambda(\mathbf{x})$ is differentiable on any open set $\mathcal{X} \subset \mathbb{R}^m$, i.e., $\mathcal{R} \in C^1(X)$.

This concludes the proof.










\subsection{Reward Estimation Consistency}
\label{app:proof_thm4}

\label{appendixB1}
\textbf{Theorem 4} (Reward Estimation Consistency)  Assume that $\Omega$ has converged properly, for any perturbed latent vector $\hat{z}_i := z_i + \hat{\epsilon} \cdot \nabla_{z_i} \mathcal{R}(\mathcal{M}_\phi(z_i, \textbf{c}))$ with small $\hat{\epsilon} > 0$, we have  $\mathcal{R}\left(\hat{\mathbf{x}}_{i}\right) > \mathcal{R}\left(\mathbf{x}_{i}\right)$, where $\hat{\mathbf{x}}_{i} = \mathcal{M}_\phi(\hat{z}_i, \textbf{c})$.

{\em Proof.} To prove this theorem, we first prove that for a well-converged $\Omega_\Theta$,  \( \mathcal{M}_\phi \) is Lipschitz-continuous.

Consider the decoder \(\mathcal{M}_\phi: \mathbb{R}^d \rightarrow \mathbb{R}^P\) composed of \( N \) layers. For \( j = 1 \) to \( N-1 \), each layer computes:
\[
h_j = \mathfrak{q}_j(h_{j-1}) = \text{ReLU}(W_j h_{j-1} + b_j),
\]
where \( h_0 = z_i \in \mathbb{R}^d \), \( W_j \in \mathbb{R}^{d_j \times d_{j-1}} \), and \( b_j \in \mathbb{R}^{d_j} \). The output layer computes:
\[
\mathbf{x}_{i} = \mathcal{M}_\phi(z_i) = \mathfrak{q}_N(h_{N-1}) = W_N h_{N-1} + b_N,
\]
with \( W_N \in \mathbb{R}^{P \times d_{N-1}} \) and \( b_N \in \mathbb{R}^P \).

To prove that \(\mathcal{M}_\phi\) is Lipschitz continuous, consider two inputs \( z_i, \hat{z}_i \in \mathbb{R}^d \). We aim to show:
\[
\| \mathcal{M}_\phi(z_i) - \mathcal{M}_\phi(\hat{z}_i) \| \leq K \| z_i - \hat{z}_i \|,
\]
where \( K \) is a finite constant.

Starting from the output layer:
\[
\begin{aligned}
\| \mathcal{M}_\phi(z_i) - \mathcal{M}_\phi(\hat{z}_i) \| &= \| \mathfrak{q}_N(h_{N-1}^{(z_i)}) - \mathfrak{q}_N(h_{N-1}^{(\hat{z}_i)}) \| \\
&= \| W_N h_{N-1}^{(z_i)} + b_N - W_N h_{N-1}^{(\hat{z}_i)} - b_N \| \\
&= \| W_N (h_{N-1}^{(z_i)} - h_{N-1}^{(\hat{z}_i)}) \| \\
&\leq \| W_N \|_2 \| h_{N-1}^{(z_i)} - h_{N-1}^{(\hat{z}_i)} \|,
\end{aligned}
\]
where \( \| W_N \|_2 \) denotes the spectral norm of \( W_N \).

For each hidden layer \( j = N-1 \) down to \( 1 \):
\[
\begin{aligned}
\| h_j^{(z_i)} - h_j^{(\hat{z}_i)} \| &= \| \mathfrak{q}_j(h_{j-1}^{(z_i)}) - \mathfrak{q}_j(h_{j-1}^{(\hat{z}_i)}) \| \\
&= \| \text{ReLU}(W_j h_{j-1}^{(z_i)} + b_j) - \text{ReLU}(W_j h_{j-1}^{(\hat{z}_i)} + b_j) \| \\
&\leq \| W_j h_{j-1}^{(z_i)} - W_j h_{j-1}^{(\hat{z}_i)} \| \quad (\text{since ReLU is 1-Lipschitz}) \\
&\leq \| W_j \|_2 \| h_{j-1}^{(z_i)} - h_{j-1}^{(\hat{z}_i)} \|.
\end{aligned}
\]

By recursively applying these inequalities, we obtain:
\[
\| h_j^{(z_i)} - h_j^{(\hat{z}_i)} \| \leq \left( \prod_{k=1}^{j} \| W_{N-k+1} \|_2 \right) \| h_0^{(z_i)} - h_0^{(\hat{z}_i)} \| = \left( \prod_{k=1}^{j} \| W_{N-k+1} \|_2 \right) \| z_i - \hat{z}_i \|.
\]

At the output layer:
\[
\| \mathcal{M}_\phi(z_i) - \mathcal{M}_\phi(\hat{z}_i) \| \leq \| W_N \|_2 \| h_{N-1}^{(z_i)} - h_{N-1}^{(\hat{z}_i)} \|.
\]

Substituting the recursive bound:
\[
\| \mathcal{M}_\phi(z_i) - \mathcal{M}_\phi(\hat{z}_i) \| \leq \left( \prod_{j=1}^{N} \| W_j \|_2 \right) \| z_i - \hat{z}_i \|.
\]

Define \( K = \prod_{j=1}^{N} \| W_j \|_2 \). To ensure \( K \) is finite, we enforce bounds (Layer Normalization) on the spectral norms: \( \| W_j \|_2 \leq s_j \), where \( s_j \) are finite constants. Then:
\[
K \leq \prod_{j=1}^{N} s_j.
\]


If we choose \( s_j = s \leq 1 \) for all \( j \), then \( K \leq s^N \leq 1\), which is finite. Therefore, \(\mathcal{M}_\phi\) is Lipschitz continuous with Lipschitz constant \( K \), satisfying:
\[
\| \mathcal{M}_\phi(z_i) - \mathcal{M}_\phi(\hat{z}_i) \| \leq K \| z_i - \hat{z}_i \|.
\]

Thus, we finished proving that \( \mathcal{M}_\phi \) is Lipschitz-continuous. Given $\Omega_\Theta$ is well converged, with $\mathcal{M}_\phi$ being Lipschitz continuous and differentiable, a small learning rate $\hat{\epsilon}$ induces a small change in latent vector $z_i$ which results in a small change in the data point $\mathbf{x}_{i}$ reconstructed by C-VAE. We can use the first-order Taylor expansion for small $\Delta z_i=\hat{z}_i-z_i$ :

$$
\hat{\mathbf{x}}_{i}=\mathcal{M}_\phi\left(\hat{z}_i\right) \approx \mathcal{M}_\phi\left(z_i\right)+J_{\mathcal{M}_\phi}\left(z_i\right) \cdot \Delta z_i
$$

where $J_{\mathcal{M}_\phi}\left(z_i\right)$ is the Jacobian matrix of $\mathcal{M}_\phi$ at $z_i$.

From the update rule:

$$
\Delta z_i=\hat{z}_i-z_i=\hat{\epsilon} \cdot \nabla_{z_i} \mathcal{R}(\mathbf{x}_{i})
$$

Thus, the change in $\mathbf{x}_{i}$ is:

$$
\hat{\mathbf{x}}_{i}-\mathbf{x}_{i} \approx J_{\mathcal{M}_\phi}\left(z_i\right) \cdot \Delta z_i=\hat{\epsilon} \cdot J_{\mathcal{M}_\phi}\left(z_i\right) \cdot \nabla_{z_i} \mathcal{R}(\mathbf{x}_{i})
$$

Since $\mathbf{x}_{i}=\mathcal{M}_\phi\left(z_i\right)$, by the chain rule, we have:

$$
\nabla_{z_i} \mathcal{R}(\mathbf{x}_{i})=J_{\mathcal{M}_\phi}^{\top}\left(z_i\right) \cdot \nabla_{\mathbf{x}_{i}} \mathcal{R}(\mathbf{x}_{i})
$$

Therefore:

$$
\hat{\mathbf{x}}_{i}-\mathbf{x}_{i} \approx \hat{\epsilon} \cdot J_{\mathcal{M}_\phi}\left(z_i\right) \cdot J_{\mathcal{M}_\phi}^{\top}\left(z_i\right) \cdot \nabla_{\mathbf{x}_{i}} \mathcal{R}(\mathbf{x}_{i})
$$

Let $\mathfrak{Z}=J_{\mathcal{M}_\phi}\left(z_i\right) \cdot J_{\mathcal{M}_\phi}^{\top}\left(z_i\right)$, which is a positive semi-definite matrix.
Thus:

$$
\hat{\mathbf{x}}_{i}-\mathbf{x}_{i} \approx \hat{\epsilon} \cdot \mathfrak{Z} \cdot \nabla_{\mathbf{x}_{i}} \mathcal{R}(\mathbf{x}_{i})
$$

Using a first-order Taylor expansion of $r$ around $\mathbf{x}_{i}$ :

$$
\Delta \mathcal{R}=\mathcal{R}(\hat{\mathbf{x}}_{i})-\mathcal{R}(\mathbf{x}_{i}) \approx \nabla_{\mathbf{x}_{i}} \mathcal{R}\left(\mathbf{x}_{i}\right)^{\top}\left(\hat{\mathbf{x}}_{i}-\mathbf{x}_{i}\right)
$$

Substituting $\hat{\mathbf{x}}_{i}-\mathbf{x}_{i}$, we obtain:

$$
\Delta \mathcal{R} \approx \hat{\epsilon} \cdot \nabla_{\mathbf{x}_{i}} \mathcal{R}\left(\mathbf{x}_{i}\right)^{\top} \mathfrak{Z} \cdot \nabla_{\mathbf{x}_{i}} \mathcal{R}\left(\mathbf{x}_{i}\right)
$$

Since $\mathfrak{Z}$ is positive semi-definite and $\hat{\epsilon}>0$ :

$$
\Delta \mathcal{R} \geq 0
$$

More specifically, $\Delta \mathcal{R}=0$ if and only if $\nabla_{\mathbf{x}_{i}} \mathcal{R}\left(\mathbf{x}_{i}\right)=0$. Otherwise, $\mathcal{R}>0$.
Therefore, under the given conditions and for a sufficiently small $\hat{\epsilon}$ :

$$
\mathcal{R}\left(\hat{\mathbf{x}}_{i}\right)>\mathcal{R}\left(\mathbf{x}_{i}\right)
$$

This completes the proof.

\begin{remark}[Trade-off between Lipschitz control and expressiveness]
Imposing strict Lipschitz constraints on the decoder $\mathcal{M}_\phi$, for instance by enforcing 
$\|W_j\|_2 \leq s_j$ for all layers so that $\prod_{j=1}^N s_j \leq K$, guarantees global smoothness 
but inevitably limits the network’s representational capacity. 
Such strong spectral normalization can suppress high-frequency components essential for accurate reconstruction, 
leading to degradation in performance. 
Hence, while a bounded $\mathcal{M}_\phi$ ensures stability, it often sacrifices fine-grained generative fidelity.
\end{remark}

\begin{remark}[Local Lipschitz behavior on latent support]
In practice, it is not necessary to enforce a global Lipschitz constraint across all decoder layers. 
The decoder $\mathcal{M}_\phi$ only needs to be \emph{locally Lipschitz} over the high-density region of the latent prior $p(z)=\mathcal{N}(0,I)$. 
Formally, there exists a constant $L_{\text{local}}>0$ such that for any $z_1,z_2$ within 
$\Omega = \{ z \mid p(z) > \epsilon \}$,
\[
\| \mathcal{M}_\phi(z_1) - \mathcal{M}_\phi(z_2) \| \le L_{\text{local}} \| z_1 - z_2 \|.
\]
This localized smoothness naturally emerges from the VAE objective, which regularizes 
$q_\psi(z|x)$ toward $p(z)$ and thereby aligns nearby $z$’s with semantically close reconstructions. 
\end{remark}

\subsection{Safe Ball In Latent Space}
\label{app:Safe ball in latent space}


\textbf{Assumption 1 (Edge–cost Lipschitz).}  
Each edge function $f_e$ is $L_e$–Lipschitz on $[0,b_e]$ and  
\[
  L_f:=\max_{e\in E}L_e .
  \tag{A1}
\]

\textbf{Assumption 2 (Estimator Uniform Error).}  
The SPAGAN surrogate satisfies  
\[
0\le  \bigl|\mathfrak{J}_{\boldsymbol\theta}(G,s,t;\mathbf{x})-\mathrm{SP}_G(s,t;\mathbf{x})\bigr|
  \le\varepsilon_{\text{spa}},\qquad\forall\,(s,t),\mathbf{x} .
  \tag{A2}\label{assump:spagan}
\]


\textbf{Proposition 2 (Safe Ball In Latent Space).} Let $\mathbf{x}_\star=\mathcal{M}_{\phi}(z_\star,c)$ be feasible and denote the true safety margin by $\bar{\Delta}=\min_{(s,t)\in\mathcal K}\bigl[\mathrm{SP}_G(s,t;\mathbf{x}_\star)-T\bigr]>0$.
Assume a well trained SPAGAN’s error is uniformly bounded by $\varepsilon_{\text{spa}}$($0\leq\varepsilon_{\text{spa}}<\bar{\Delta}$).
With $h=\lceil T/w_{\min}\rceil,\;L_f=\max_{e}L_e,\;m=|E|,$ and decoder-Lipschitz constant $L_{\mathcal{M}}$, every latent vector $z$  satisfying

$$
\bigl\|z-z_\star\bigr\|_2\le
\frac{\bar{\Delta}-\varepsilon_{\text{spa}}}{h\,L_f\,\sqrt{m}\,L_{\mathcal{M}}}
$$

is decoded to $\mathbf{x}=\mathcal{M}_{\phi}(z,c)$ for which
$\mathfrak{J}_\theta(G, s, t ; \mathbf{x}) \geq T-2 \epsilon_{s p a}$ for all $(s,t)\in\mathcal K$; Thus, exploration within this latent ball guarantees provable safety and restores near full feasibility.

\textbf{Proof.} We begin by recalling Decoder $\mathcal{M}_{\phi}:\mathbb R^{d}\!\to\!\mathbb R^{m}$ (with $m=|E|$) is globally $L_{\mathcal{M}}$-Lipschitz

\begin{equation}
    \|\mathcal{M}_{\phi}(z_1,c)-\mathcal{M}_{\phi}(z_2,c)\|_2\le L_{\mathcal{M}}\|z_1-z_2\|_2,\qquad\forall z_1,z_2\in\mathbb R^{d}.
    \label{eqn:lip_decoder}
\end{equation}

With $w_{\min}:=\min_{e}w_e$, let $h:=\bigl\lceil T/w_{\min}\bigr\rceil$ 
so every shortest path $\ddot\rho$ under threshold $T$ satisfies $|\ddot\rho|\le h$. For any two perturbations $\mathbf{x},\mathbf{x}'$ and any $(s,t)\in\mathcal K$, we have: 

\begin{equation}
    \begin{aligned}
\bigl|\mathrm{SP}_G(s,t;\mathbf{x})-\mathrm{SP}_G(s,t;\mathbf{x}')\bigr|
&=\Bigl|\,\sum_{e\in\ddot{\rho}}f_e(\mathbf{x}_e)-\sum_{e\in\ddot{\rho}}f_e(\mathbf{x}_e')\Bigr| \\[2pt]
&\le \sum_{e\in\ddot{\rho}}L_e|\mathbf{x}_e-\mathbf{x}_e'| \\
&\le h\,L_f\,\|\mathbf{x}-\mathbf{x}'\|_1                              \\[4pt]
&\le h\,L_f\,\sqrt{m}\,\|\mathbf{x}-\mathbf{x}'\|_2. 
\end{aligned}
\label{eqn:lips_SP_G_1}
\end{equation}

Switch to the estimator using assumption \ref{assump:spagan} and by applying the triangle inequality, which states $|a+b+c| \leq|a|+|b|+|c|$, we have:

\begin{equation}
\label{eq:estimator_diff_expand}
\begin{aligned}
&\bigl|\mathfrak J_{\boldsymbol\theta}(s,t;\mathbf{x})-\mathfrak J_{\boldsymbol\theta}(s,t;\mathbf{x}_\star)\bigr| =
   \Bigl|
     \underbrace{ \mathfrak J_{\boldsymbol\theta}(s,t;\mathbf{x})-\mathrm{SP}_G(s,t;\mathbf{x})
      }_{a} + \underbrace{\mathrm{SP}_G(s,t;\mathbf{x})-\mathrm{SP}_G(s,t;\mathbf{x}_\star)}_{b} \\& \qquad \qquad \qquad \qquad \qquad \qquad+ \underbrace{\mathrm{SP}_G(s,t;\mathbf{x}_\star)-\mathfrak J_{\boldsymbol\theta}(s,t;\mathbf{x}_\star)}_{c}
   \Bigr| \\[10pt]
&\quad\le 
   \underbrace{\bigl|\mathfrak J_{\boldsymbol\theta}(s,t;\mathbf{x})-\mathrm{SP}_G(s,t;\mathbf{x})\bigr|}_{\le \,\varepsilon_{\text{spa}}}
   \;+\;
   \underbrace{\bigl|\mathrm{SP}_G(s,t;\mathbf{x})-\mathrm{SP}_G(s,t;\mathbf{x}_\star)\bigr|}_{\text{true–cost gap}}
   \;+\;
   \underbrace{\bigl|\mathrm{SP}_G(s,t;\mathbf{x}_\star)-\mathfrak J_{\boldsymbol\theta}(s,t;\mathbf{x}_\star)\bigr|}_{\le \,\varepsilon_{\text{spa}}} \\[10pt]
&\quad {\le}
      \varepsilon_{\text{spa}}
      \;+\;
      \bigl|\mathrm{SP}_G(s,t;\mathbf{x})-\mathrm{SP}_G(s,t;\mathbf{x}_\star)\bigr|
      \;+\;
      \varepsilon_{\text{spa}} \\[10pt]
&\quad=
      2\varepsilon_{\text{spa}}
      \;+\;
      \bigl|\mathrm{SP}_G(s,t;\mathbf{x})-\mathrm{SP}_G(s,t;\mathbf{x}_\star)\bigr| \\[10pt]
&\quad{\le}
      2\varepsilon_{\text{spa}}
      \;+\;
      h\,L_f\,\sqrt{m}\,
      \bigl\|\mathcal{M}_{\phi}(z,c)-\mathcal{M}_{\phi}(z_\star,c)\bigr\|_2 \\[10pt]
&\quad{\le}
      2\varepsilon_{\text{spa}}
      \;+\;
      h\,L_f\,\sqrt{m}\,L_{\mathcal{M}}\,
      \|z-z_\star\|_2 \\
&\Longrightarrow - \bigl|\mathfrak J_{\boldsymbol\theta}(s,t;\mathbf{x})-\mathfrak J_{\boldsymbol\theta}(s,t;\mathbf{x}_\star)\bigr| \geq - (2\varepsilon_{\text{spa}}
      \;+\;
      h\,L_f\,\sqrt{m}\,L_{\mathcal{M}}\,
      \|z-z_\star\|_2 )
\end{aligned}
\end{equation}

The initial solution $\mathbf{x}_*$ is feasible with a true safety margin $\bar{\Delta}$ that $ S P_G\left(s, t ; \mathbf{x}_*\right) \geq T+\bar{\Delta}$. Moreoever, from Assumption 2, we have

\begin{equation}
    \begin{aligned}
&\left|\mathfrak{F}_\theta\left(G, s, t ; \mathbf{x}_*\right)-S P_G\left(s, t ; \mathbf{x}_*\right)\right| \leq \epsilon_{s p a}\\
&\Longleftrightarrow -\epsilon_{s p a} \leq \mathfrak{J}_\theta\left(G, s, t ; \mathbf{x}_*\right)-S P_G\left(s, t ; \mathbf{x}_*\right) \leq \epsilon_{s p a}
\end{aligned}
\end{equation}

From this compound inequality, we are interested in the left-hand side to find a lower bound for $\mathfrak{F}_\theta\left(G, s, t ; \mathbf{x}_*\right):$

\begin{equation}
    \begin{aligned}
        &-\epsilon_{s p a} \leq \mathfrak{J}_\theta\left(G, s, t ; \mathbf{x}_*\right)-S P_G\left(s, t ; \mathbf{x}_*\right)\\
        &\Longleftrightarrow \mathfrak{F}_\theta\left(G, s, t ; \mathbf{x}_*\right) \geq S P_G\left(s, t ; \mathbf{x}_*\right)-\epsilon_{s p a}\\
        &\Longleftrightarrow \mathfrak{F}_\theta\left(G, s, t ; \mathbf{x}_*\right) \geq T+\bar{\Delta}-\epsilon_{s p a}
    \end{aligned}
\end{equation}

Since the proposition states the condition $\left\|z-z_{\star}\right\|_2 \leq \frac{\bar{\Delta}-\epsilon_{s p a}}{h L_f \sqrt{m} L_{\mathcal{M}}}$ and we know that $h, L_{-} f, \sqrt{m}, L_{\mathcal{M}}$ are all positive, this implies $h L_f \sqrt{m} L_{\mathcal{M}}\left\|z-z_{\star}\right\|_2 \leq \bar{\Delta}-\epsilon_{s p a}$, hence:

\begin{equation}
\begin{aligned}
\mathfrak{J}_\theta(G, s, t ; \mathbf{x}) &\geq \mathfrak{J}_\theta\left(G, s, t ; \mathbf{x}_*\right)-\left|\mathfrak{J}_\theta(G, s, t ; \mathbf{x})-\mathfrak{J}_\theta\left(G, s, t ; \mathbf{x}_*\right)\right| \\
&\text{(Note: since $A \geq B - |A-B|$)}\\
&\geq (T+\bar{\Delta}-\epsilon_{s p a} ) - \left|\mathfrak{J}_\theta(G, s, t ; \mathbf{x})-\mathfrak{J}_\theta\left(G, s, t ; \mathbf{x}_*\right)\right|\\ &\text{(Note: Substituting the lower bound for } \mathfrak{J}_\theta(G,s,t;\mathbf{x}_*))\\
&\geq (T+\bar{\Delta}-\epsilon_{s p a} ) - \left(2\varepsilon_{\text{spa}} + h\,L_f\,\sqrt{m}\,L_{\mathcal{M}}\, \|z-z_\star\|_2 \right)\\ &\text{(Note: Substituting the upper bound for } |\mathfrak{J}_\theta(\mathbf{x})-\mathfrak{J}_\theta(\mathbf{x}_*)| \text{)}\\
&= T+\bar{\Delta}-\epsilon_{s p a} - 2\varepsilon_{\text{spa}} - h\,L_f\,\sqrt{m}\,L_{\mathcal{M}}\, \|z-z_\star\|_2 \\
&= T+\bar{\Delta}-3 \epsilon_{s p a}-h L_f \sqrt{m} L_{\mathcal{M}}\left\|z-z_*\right\|_2\\
&\geq T+\bar{\Delta}-3 \epsilon_{s p a} - (\overline{\Delta} - \epsilon_{spa})\\ &\text{(Note: Since } -h L_f \sqrt{m} L_{\mathcal{M}}\left\|z-z_*\right\|_2 \ge -(\overline{\Delta} - \epsilon_{spa}) \text{)}\\
&= T+\bar{\Delta}-3 \epsilon_{s p a} - \overline{\Delta} + \epsilon_{spa} \\
&= T - 2\epsilon_{s p a}
\end{aligned}
\end{equation}

This completes the proof.

\section{DETAILS EXPERIMENTS AND ABLATION STUDIES}
\label{app:experiments}

This section provides comprehensive details of our experimental setup, evaluation metrics, and ablation studies. We aim to assess both the effectiveness and generalizability of the proposed Hephaestus framework across a range of real-world network topologies and baselines. The experiments are designed to evaluate performance from multiple perspectives: solution feasibility, budget efficiency, scalability, and training dynamics. In addition, ablation studies are conducted to isolate the contributions of key architectural components—including the SPAGAN-based path estimator, mixture of generative experts, and reinforcement-based refinement—and to examine how each contributes to the overall performance. We also detail the hyperparameter choices, compute infrastructure, and exact solver setups used for benchmarking.

\subsection{Dataset Details} \label{app:dataset_details}

Table\ref{tab:network-stats} provides detailed statistics of the real-world network datasets used in our experiments, each chosen to represent different structural and functional properties. The Email network is a directed communication graph among 1,005 individuals with 25,571 edges and a diameter of 7, capturing organizational email interactions characterized by strong community structures. The Gnutella dataset represents a decentralized peer-to-peer file-sharing system from August 2002. It contains 6,301 nodes and 20,777 directed connections, with a small diameter of 9 and low clustering, reflecting its unstructured topology. The RoadCA network is a large-scale undirected graph of California's road infrastructure with approximately 1.96 million nodes and 2.77 million edges, and an unusually large diameter of 849—typical of sparse, planar transportation networks. Finally, the Skitter dataset models the Internet at the autonomous system (AS) level using traceroute data, comprising 1.7 million nodes and 11.1 million directed edges with a diameter of 25. This dataset captures the hierarchical structure of inter-AS connectivity. Collectively, these datasets span a wide spectrum of scales, densities, and network types, forming a 
comprehensive benchmark for evaluating the scalability and generalizability of Hephaestus and its baselines.

\begin{table}[htp]
\centering
\caption{Statistics of real network datasets used in experiments.}
\label{tab:network-stats}
\begin{tabular}{l|l|r|r|r}
\toprule 
\textbf{Data} & \textbf{Type} & \textbf{Nodes} & \textbf{Edges} & \textbf{Diameter} \\
\midrule 
Email   & Directed & 1,005  & 25,571  & 11 \\ 
Gnutella & Directed   & 6,301 & 20,777 & 9   \\
RoadCA   & Undirected & 1.96 M  & 2.77 M  & 849 \\ 
Skitter  & Directed   & 1.7 M  & 11.1 M & 25  \\
\bottomrule 
\end{tabular}
\end{table}

\begin{table}[htp]
\centering
\caption{Hyperparameter Settings.}
\label{tab:hyperparameters} 
\setlength{\tabcolsep}{5pt}
\renewcommand{\arraystretch}{1.1}
\begin{tabular}{@{}lll@{}}
\toprule
\textbf{Component/Phase} & \textbf{Parameter} & \textbf{Value} \\
\midrule

\multicolumn{3}{l}{\textit{\textbf{Phase 1: Forge}}} \\
SPAGAN ($\mathfrak{F}_{\boldsymbol{\theta}}$) & Network Architecture & 5 Shortest Path GAT layers, 512 units, 8 heads \\
& Activation Function & ELU \\
& Learning Rate ($\alpha$) & \num{5e-4} \\
& Optimizer & Adam \\
& Adam $\beta_1, \beta_2$ & (0.9, 0.999) \\
& Batch Size & 256 \\
& Training Epochs & 3000 \\
& Loss Function & Huber Loss \\
& Max per-edge budget ($b_e$) & T \\
\midrule

\multicolumn{3}{l}{\textit{\textbf{Phase 2: Morph}}} \\ 
EBM ($q(\mathbf{x})$) & Network Architecture & 6 MLP layers, 512 units \\
& Activation Function & Swish \\
& Learning Rate ($\alpha_{EBM}$) & \num{2e-4} \\
& Optimizer & Adam \\
& Regularization ($\hat{\gamma}$) & 1.0 \\
& Batch Size (EBM Update) & 256 \\
& Training Iterations (Minimax) & 50000 \\
\midrule 
Mix-CVAE ($\Omega$) & Latent Dimension of each CVAE ($d$) & 128 \\
& Encoder Architecture & GAT Encoder + MLP \\
& Decoder Architecture & MLP Decoder \\
& Activation Function & LeakyReLU \\
& Learning Rate ($\alpha_{\Omega}$) & \num{8e-4} \\
& Optimizer & Adam \\
& Batch Size (CVAE Update) & 256 \\
& Initial Experts ($N_{init}$) & 1 \\
& Max Experts ($N_{max}$) & 9 \\
& Expert Add Threshold ($\delta$) & 0.425 \\
& KL Weight ($\beta_{KL}$) & 0.1 \\
& Prior & $\mathcal{N}(0, I)$ \\
\midrule

\multicolumn{3}{l}{\textit{\textbf{Phase 3: Refine}}} \\
RL Agent ($\pi$) & Policy/Value Net Arch. & 4 MLP layers, 256 units \\
& Activation Function & LeakyReLU \\
& Learning Rate ($\alpha_{RL}$) & \num{1e-4} \\
& Optimizer & Adam \\
& Discount Factor ($\gamma_{RL}$) & 0.99 \\
& Reward Smoothness ($\zeta$) & 5.0 \\
& Reward Cost Weight ($\varkappa$) & 0.05 \\
& Gradient Ascent Step ($\hat{\epsilon}_{GA}$) & \num{2e-3} \\
& RL Training Episodes & 50000 \\
& Top-K Solutions & 10 \\

\bottomrule
\end{tabular}
\end{table}

\subsection{Hyperparameter Settings}  \label{app:hyperparameters}

In Table~\ref{tab:hyperparameters}, we provide a detailed summary of the hyperparameters used across the three core phases of the \PIMMA framework: Forge, Morph, and Refine. These settings were chosen through extensive trial runs and empirical tuning to identify the best-performing configurations. The selected values aim to ensure training stability, strong generalization across diverse graph instances, and consistent performance throughout the pipeline.

In Forge, we train the SPAGAN model to estimate shortest-path costs efficiently. After trying several network configurations, we found that using 5 layers of Graph Attention Networks (GAT), each with 512 hidden units and 8 attention heads, gave reliable results on a wide range of graph structures. The ELU activation function was chosen because it helps avoid the dead neuron problem that can happen with ReLU, especially during early stages of training. For the learning rate, we tested multiple values and observed that $5 \times 10^{-4}$ consistently led to stable convergence within 3000 epochs. Larger learning rates often made training unstable, while smaller ones slowed down progress too much. When choosing a loss function, we initially experimented with mean squared error (MSE), but observed that it was highly sensitive to a few long-path outliers, which harms the overall learning process. To address this, we adopted the Huber loss, which combines the benefits of MSE and MAE (Mean Absolute Error): it behaves quadratically for small errors to ensure smooth optimization and transitions to a only linear penalty for large errors. This property allowed us to reduce the influence of extreme outliers while still penalizing them, resulting in more stable convergence and improved predictive accuracy across diverse graph instances. The maximum per-edge budget $b_e = T$ defines the allowable range of perturbations and ensures that a feasible solution always exists for any input graph. Specifically, by assigning $x_e = T$ to every edge $e \in E$, the cost of any path becomes at least $T$, thus trivially satisfying the QoSD constraint. This choice guarantees feasibility without loss of generality, while still leaving many rooms for the model to explore more efficient and sparse perturbations to reduce total perturbation cost. Finally, we used a batch size of 256, which provided reliable gradients while fitting comfortably within GPU memory during training. 

In Morph, the EBM is a 6-layer MLP with 512 units and Swish activation, which improves smoothness of the learned energy surface. The EBM is trained with a learning rate of $2 \times 10^{-4}$, using a minimax schedule for 50,000 iterations, along with regularization term $\hat{\gamma} = 1.0$ to prevent gradient explosion. For the Mix-CVAE, a latent dimension of 128 was sufficient to model the diverse structure of perturbations, while the encoder incorporates graph context via GAT layers. The KL weight $\beta_{KL} = 0.15$ balances reconstruction and latent regularization. We initialize with a single expert and allow expansion up to 9 experts, adding new CVAEs when the density ratio $\chi > 0.425$, indicating insufficient coverage by existing experts. 

In Refine, we implement policy $\pi$ as a 4-layer MLP with 256 units per layer and LeakyReLU activations which helps mitigate vanishing gradients. It commonly occur during late-stage training when the model begins to converge and reward differences become minimal. A conservative learning rate of $1 \times 10^{-4}$ was selected to ensure policy stability, while a discount factor of 0.99 promotes long-term planning over greedy improvements. The reward shaping parameters $\zeta = 5.0$ and $\varkappa = 0.05$ control the smoothness of feasibility and cost feedback, respectively. We perturb the latent vector with a small gradient ascent step $\hat{\epsilon}_{GA} = 2 \times 10^{-3}$ to explore reward-improving directions, and retain the top-10 feasible solutions per episode to enrich the training buffer. We follow the model architecture of PPO based on works \cite{self-rl-controlling, do2024mimreasonerlearningtheoreticalguarantees, 9423319}

\textbf{Hardware Specification.} All experiments were conducted on a workstation equipped with an Intel Core i9-14900K CPU, 192 GB RAM, and 2× NVIDIA RTX 4090 GPUs (total 48 GB VRAM). While the GPUs played a critical role in training the SPAGAN, Mix-CVAE, and RL components efficiently, the large RAM and CPU core count were especially important for evaluating exact solvers like Gurobi and approximation algorithms. In particular, we relied on Gurobi to refine and benchmark outputs from baseline methods such as DIFFILO, Predict-and-Search, and L-MILPOPT, etc. These refinement steps often required solving large-scale ILPs with huge number of constraints and integer variables, where runtime and memory bottlenecks were significant. Thus, the high-performance CPU and 192 GB RAM were essential for verifying solution quality and feasibility in our exact evaluation pipeline.

\textbf{Gurobi Heuristic Setup (Important).} To assess the quality of learned solutions and evaluate optimality under exact conditions, we use Gurobi as a ground-truth solver. However, solving the full QoSD problem exactly is highly challenging at scale due to the exponential number of potential constraints. Recall $h = \left\lceil \frac{T}{w_{\min}} \right\rceil$ denotes the maximum number of edges in any feasible path, where $T$ is the budget threshold and $w_{\min}$ is the minimum edge weight.  Since each path can consist of at most $h$ edges, and the graph contains $n$ nodes, the total number of feasible paths is upper-bounded by $n^h$. This results in an exponentially large constraint space in the corresponding MILP formulation. This issue is especially severe when the threshold $T$ is high, since a larger number of short paths remain feasible. In our experiments, we observed that Gurobi consistently fails to solve problem instances with more than 10,000 nodes at maximum density when all feasible paths are explicitly enumerated under medium $T$. 

To address this scalability bottleneck, we adopt a heuristic strategy based on sampling for Gurobi refinement, which we refer to as Gurobi-Heuristic. Instead of enumerating all paths, we iteratively sample the current shortest paths between critical source-target pairs—using Dijkstra—and enforce path constraints only on these sampled paths. Gurobi then solves the reduced problem and updates the edge weights accordingly. This process is repeated: new shortest paths are sampled under the updated perturbation $\mathbf{x}$, and the solver is rerun. The iteration stops once no newly sampled path violates the threshold constraint $T$, thereby ensuring feasibility without needing to enumerate the full exponential set of feasible paths shorter than $T$. This strategy is crucial for making it feasible to refine the solutions produced by any ML-based method—including Hephaestus or baselines such as DIFFILO and Predict-and-Search—on large-scale graphs. Without it, exact refinement via Gurobi would be computationally infeasible due to the overwhelming number of constraints. To further improve performance under such settings, Gurobi is configured with heuristic-oriented parameters including `Heuristics=0.5`, `MIPFocus=1`, and full multi-threading support.

\subsection{SPAGAN Generalization} \label{app:spagan_validation}

\begin{figure*}[htp]
    \centering
    \captionsetup[subfigure]{skip=0.1pt}  
    \begin{subfigure}[t]{0.32\textwidth}
        \centering
        \includegraphics[width=\linewidth]{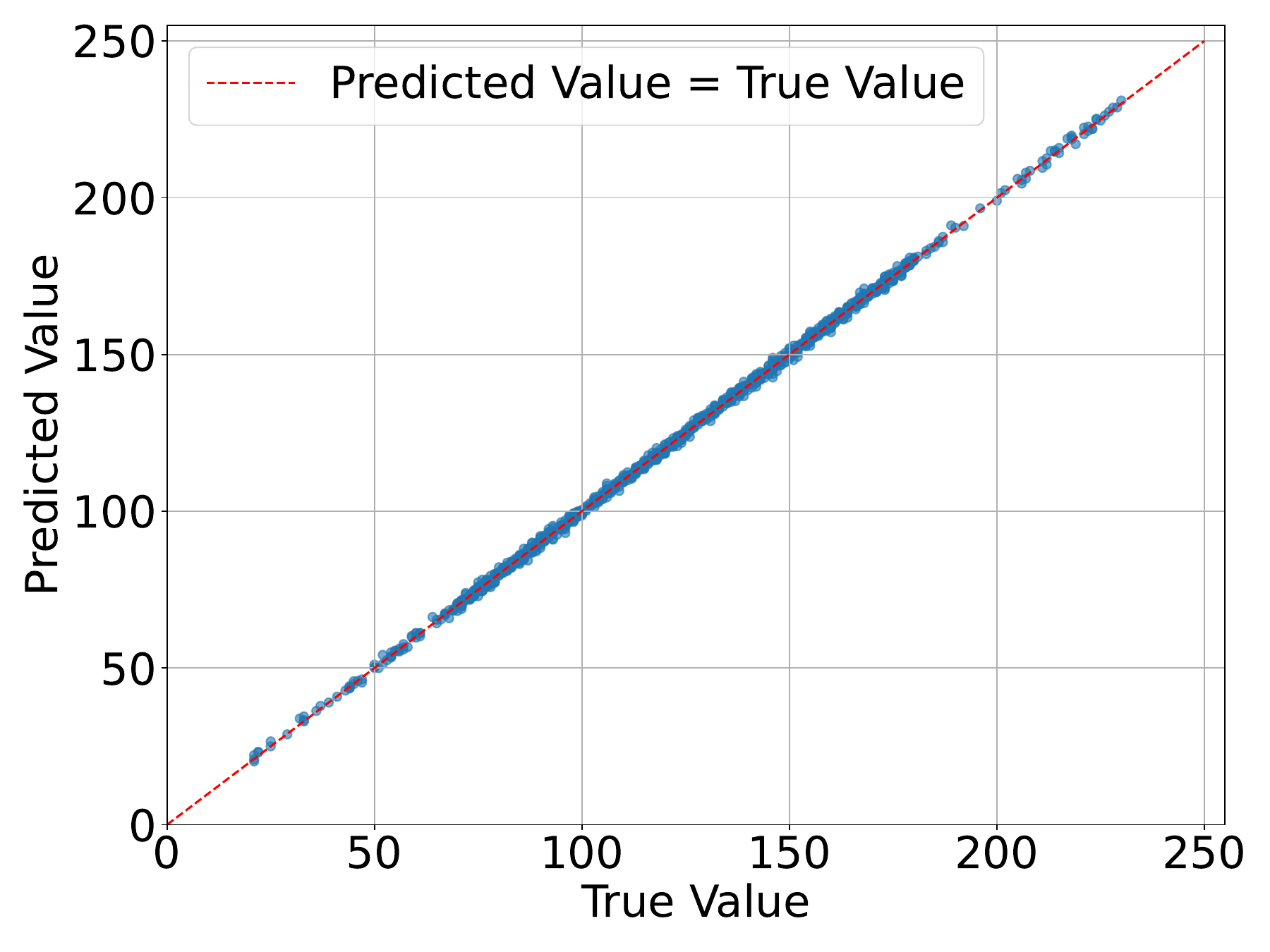}
        \caption{Graph 1000}
    \end{subfigure}
    \hfill 
    \begin{subfigure}[t]{0.32\textwidth}
        \centering
        \includegraphics[width=\linewidth]{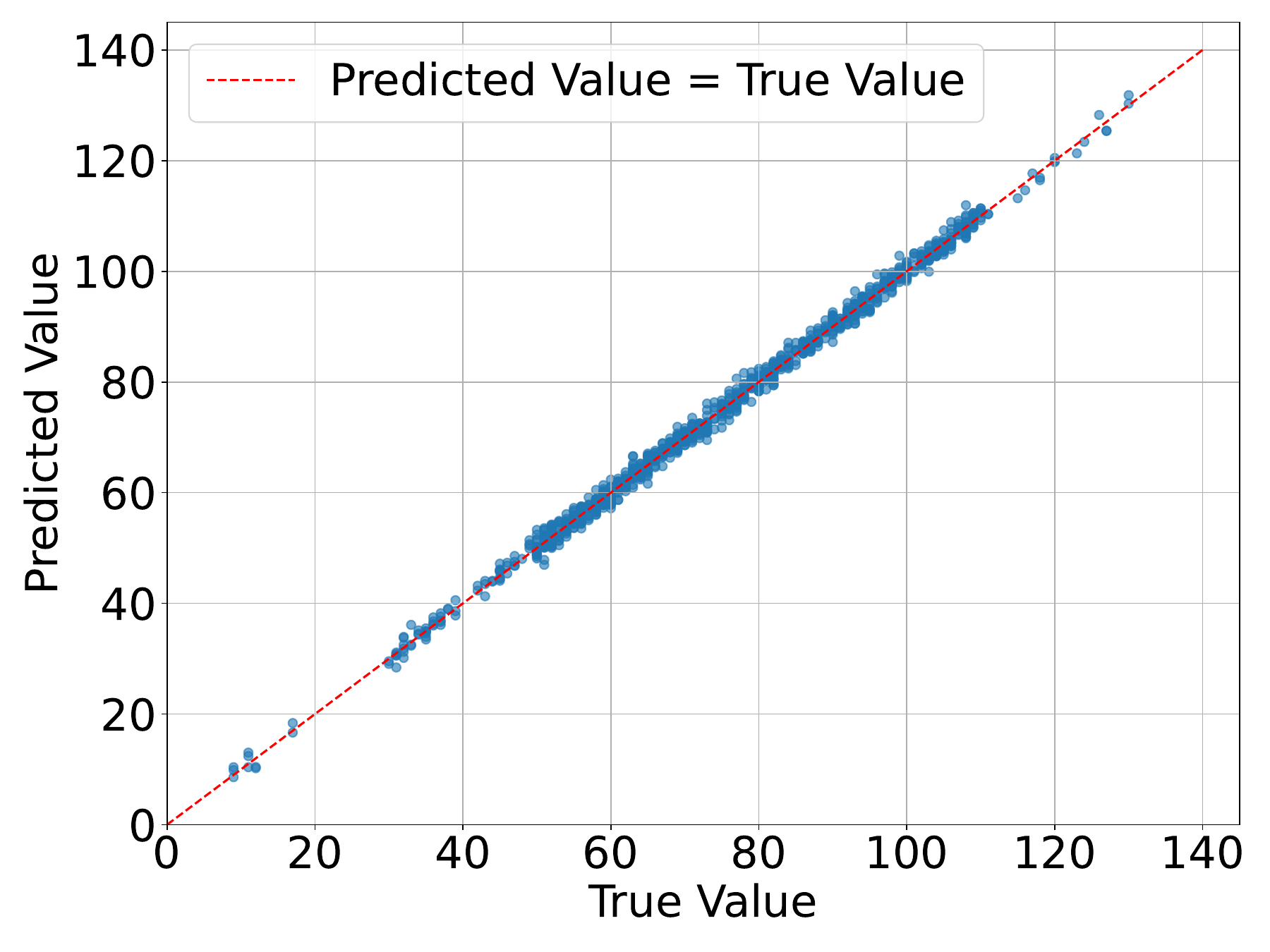}
        \caption{Graph 2000}
    \end{subfigure}
    \hfill 
    \begin{subfigure}[t]{0.32\textwidth}
        \centering
        \includegraphics[width=\linewidth]{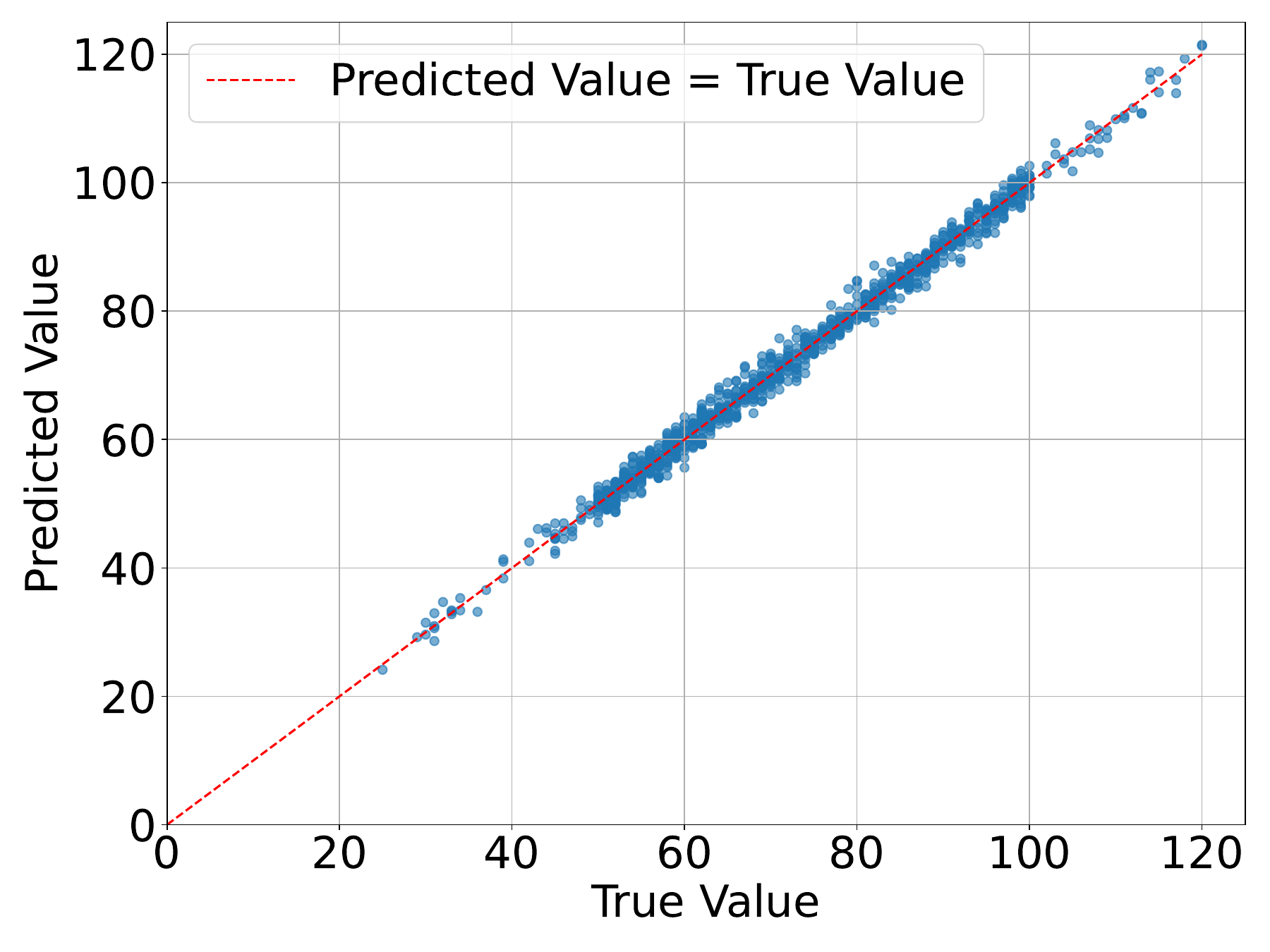}
        \caption{Graph 3000}
    \end{subfigure}
    
    \vspace{10pt} 
    
    \begin{subfigure}[t]{0.32\textwidth}
        \centering
        \includegraphics[width=\linewidth]{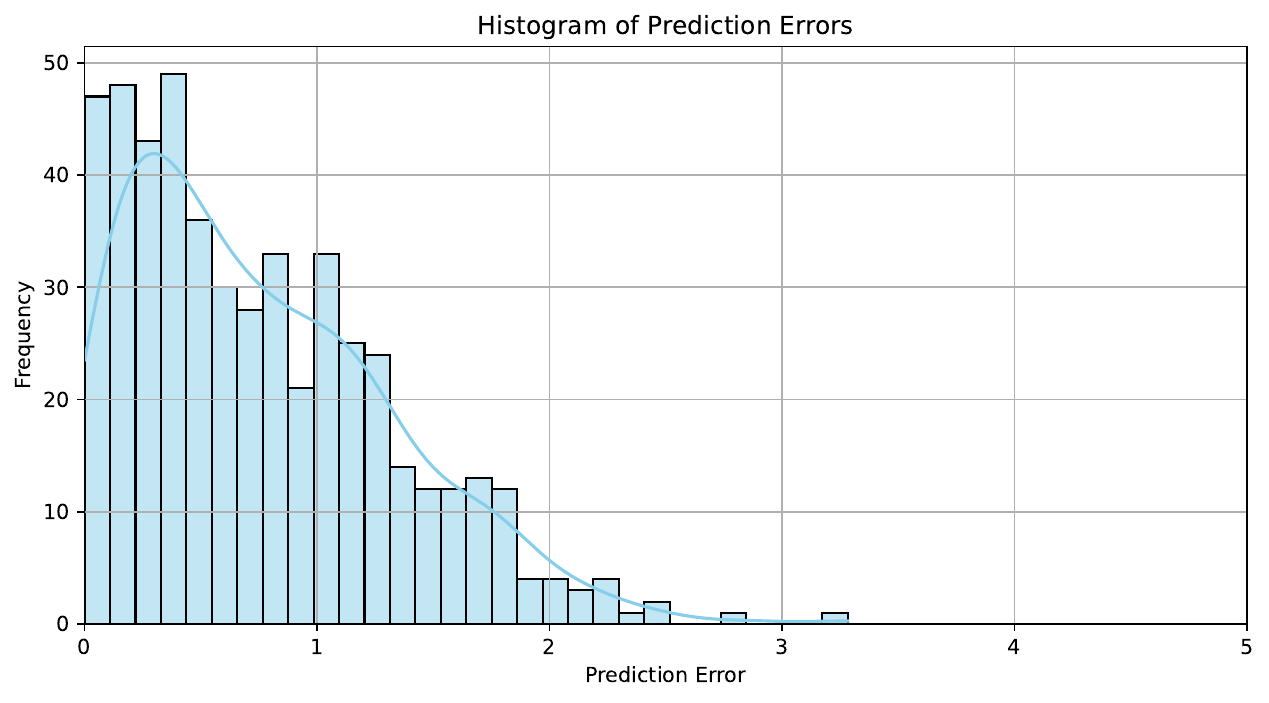}
        \caption{Graph 1000}
    \end{subfigure}
    \hfill 
    \begin{subfigure}[t]{0.32\textwidth}
        \centering
        \includegraphics[width=\linewidth]{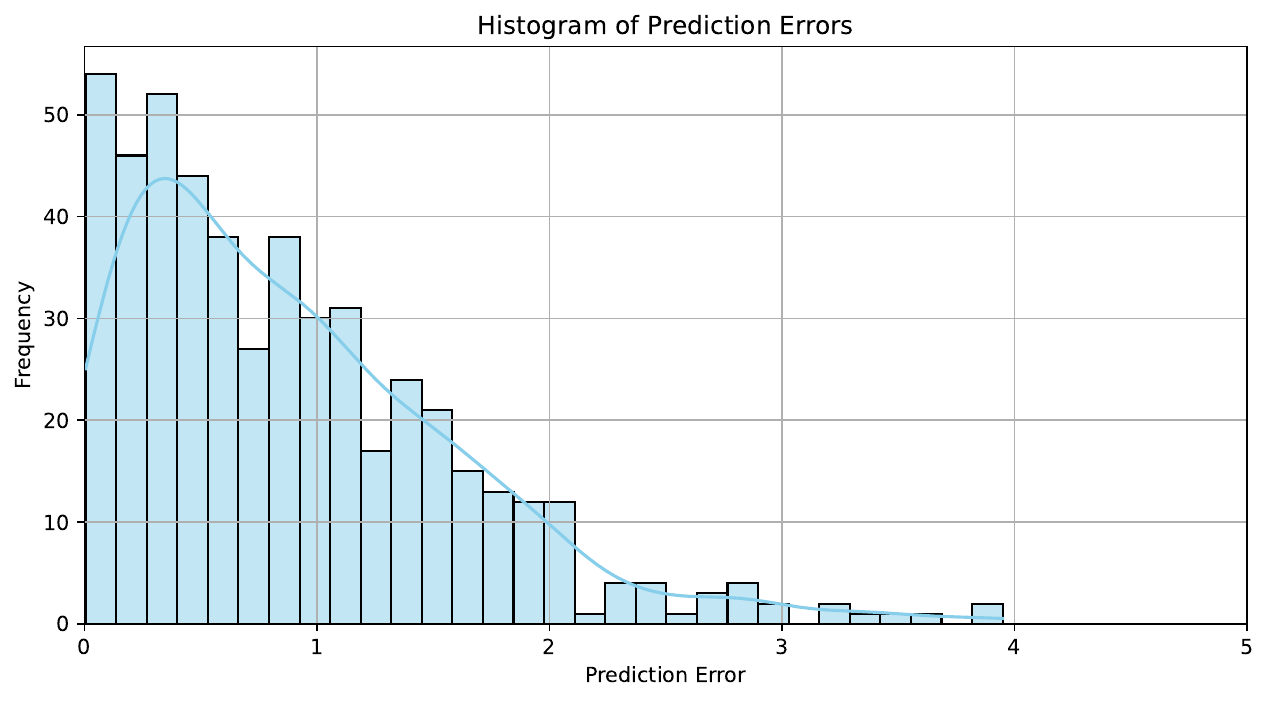}
        \caption{Graph 2000}
    \end{subfigure}
    \hfill 
    \begin{subfigure}[t]{0.32\textwidth}
        \centering
        \includegraphics[width=\linewidth]{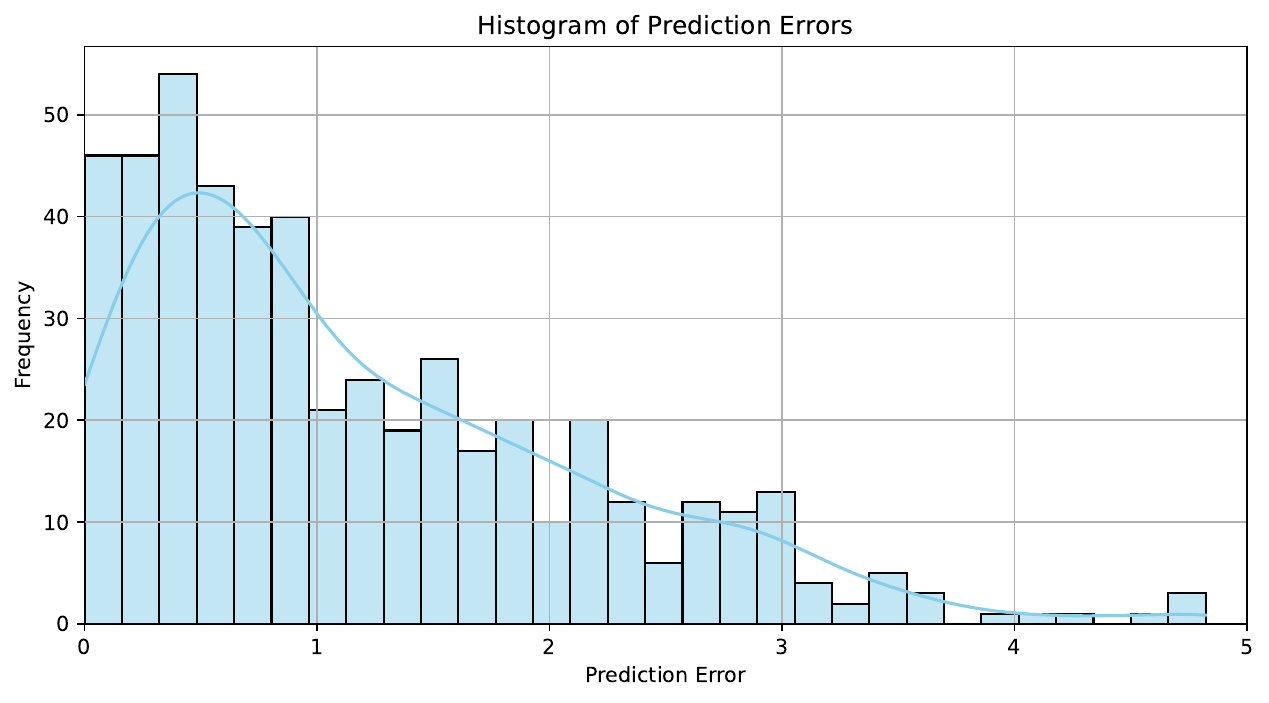}
        \caption{Graph 3000}
    \end{subfigure}
    
    \caption{Predictive Accuracy Across Different Graph Sizes.}
    \label{fig:spagan_1}
\end{figure*}

\begin{figure*}[htp]
    \centering
    \captionsetup[subfigure]{skip=0.1pt}  
    \begin{subfigure}[t]{0.32\textwidth}
        \centering
        \includegraphics[width=\linewidth]{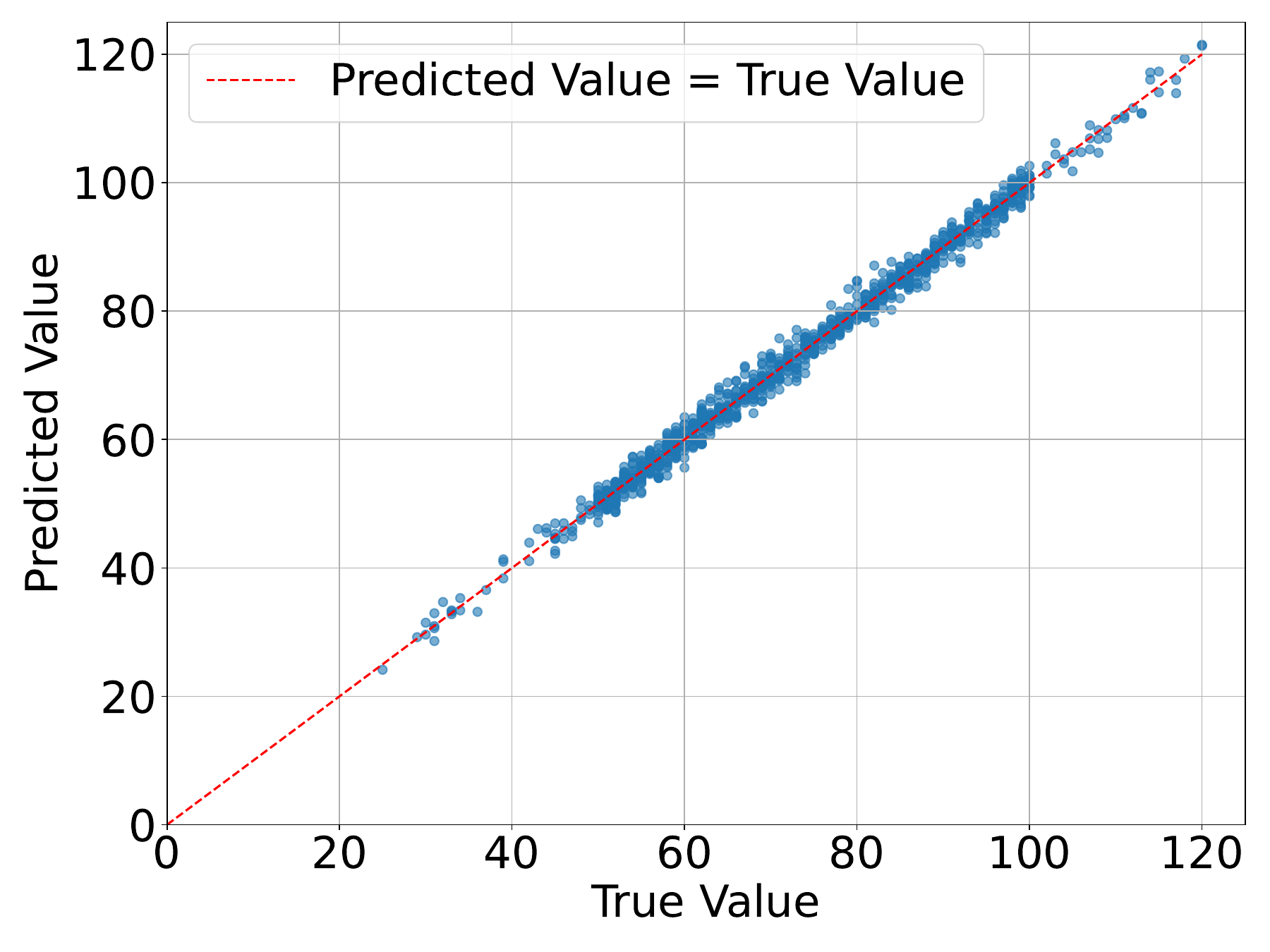}
        \caption{Erdős-Rényi Model}
    \end{subfigure}
    \hfill 
    \begin{subfigure}[t]{0.32\textwidth}
        \centering
        \includegraphics[width=\linewidth]{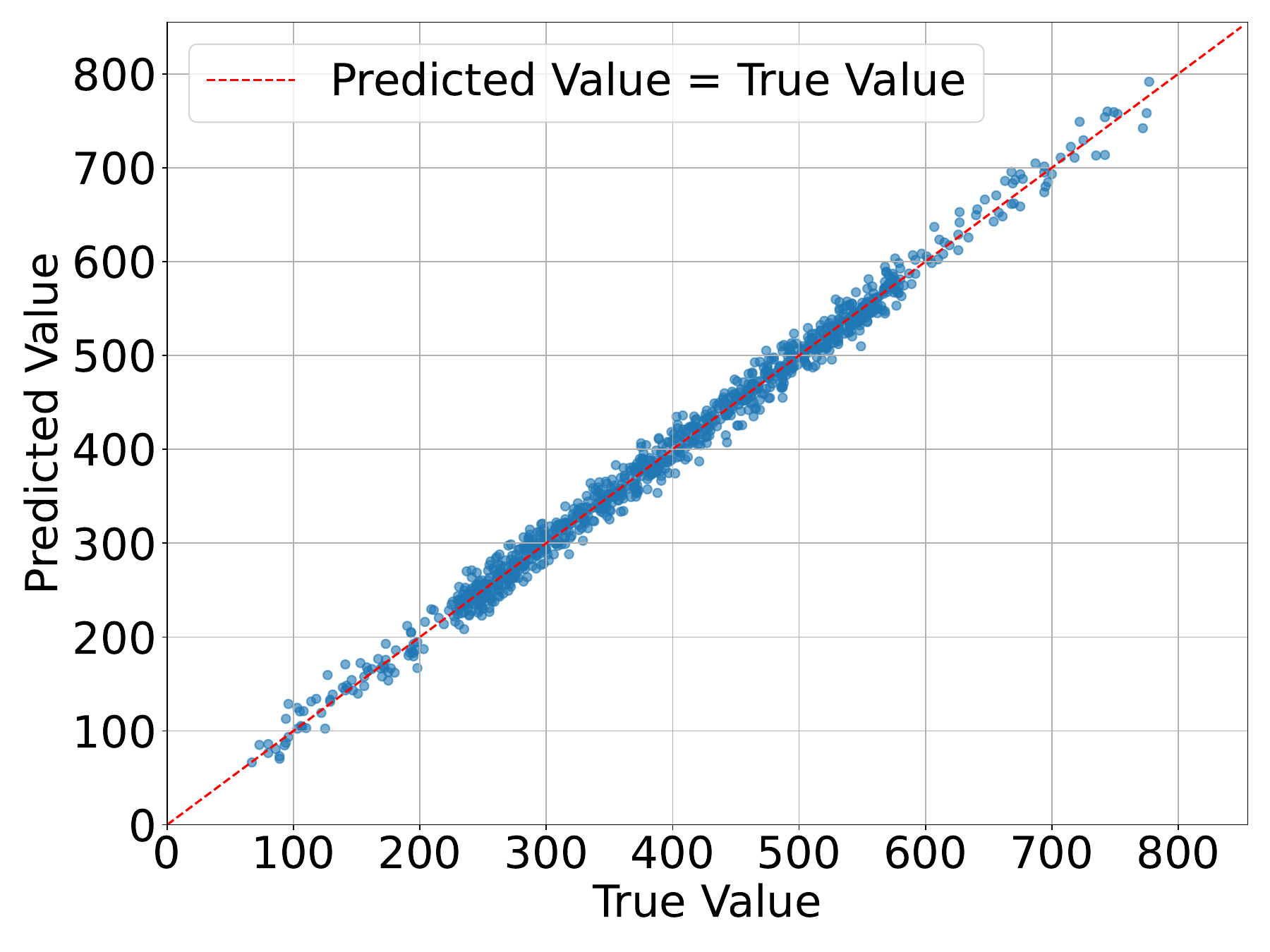}
        \caption{Barabási-Albert Model}
    \end{subfigure}
    \hfill 
    \begin{subfigure}[t]{0.32\textwidth}
        \centering
        \includegraphics[width=\linewidth]{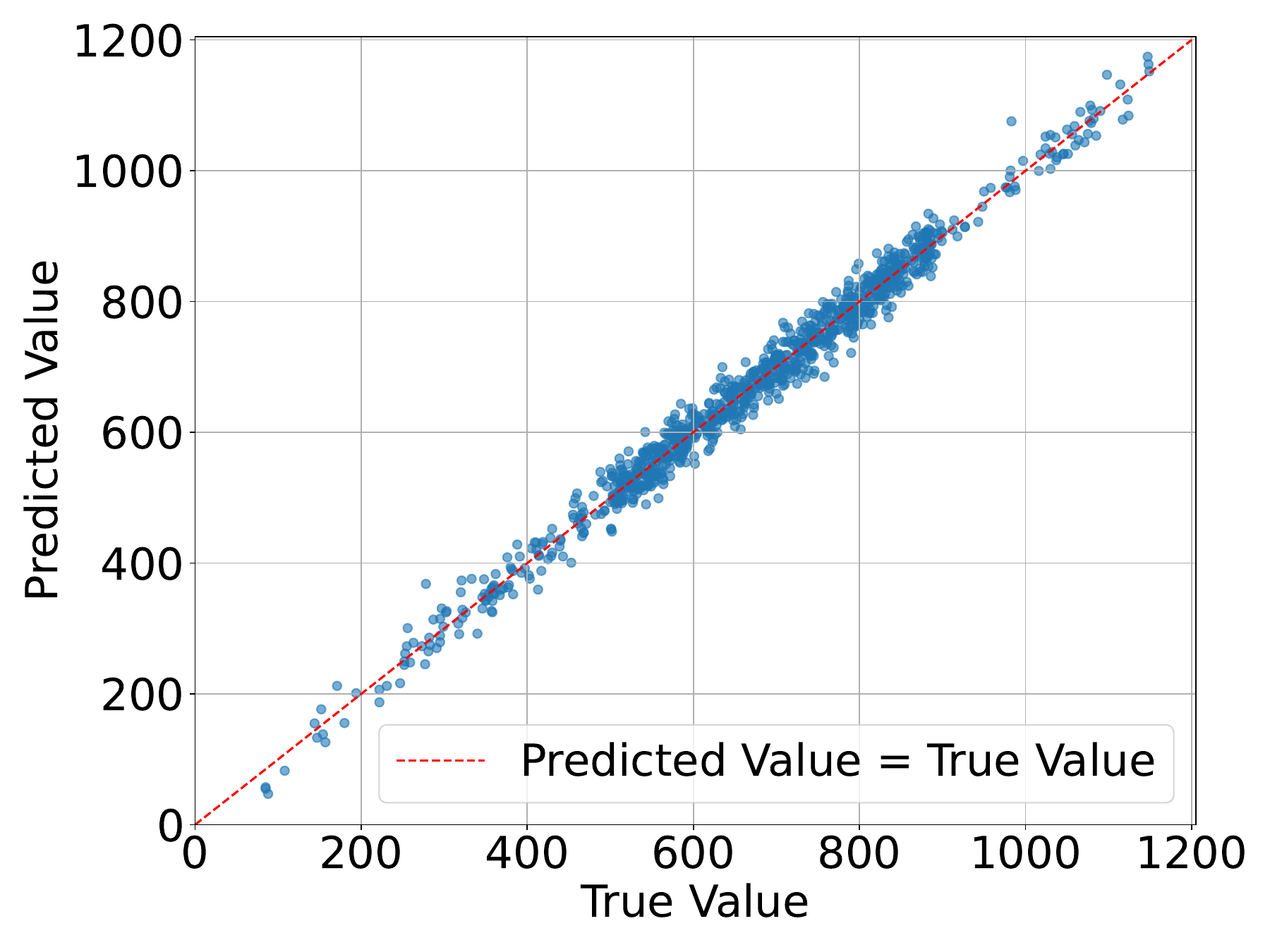}
        \caption{Watts-Strogatz Model}
    \end{subfigure}
    
    \vspace{10pt} 
    
    \begin{subfigure}[t]{0.32\textwidth}
        \centering
        \includegraphics[width=\linewidth]{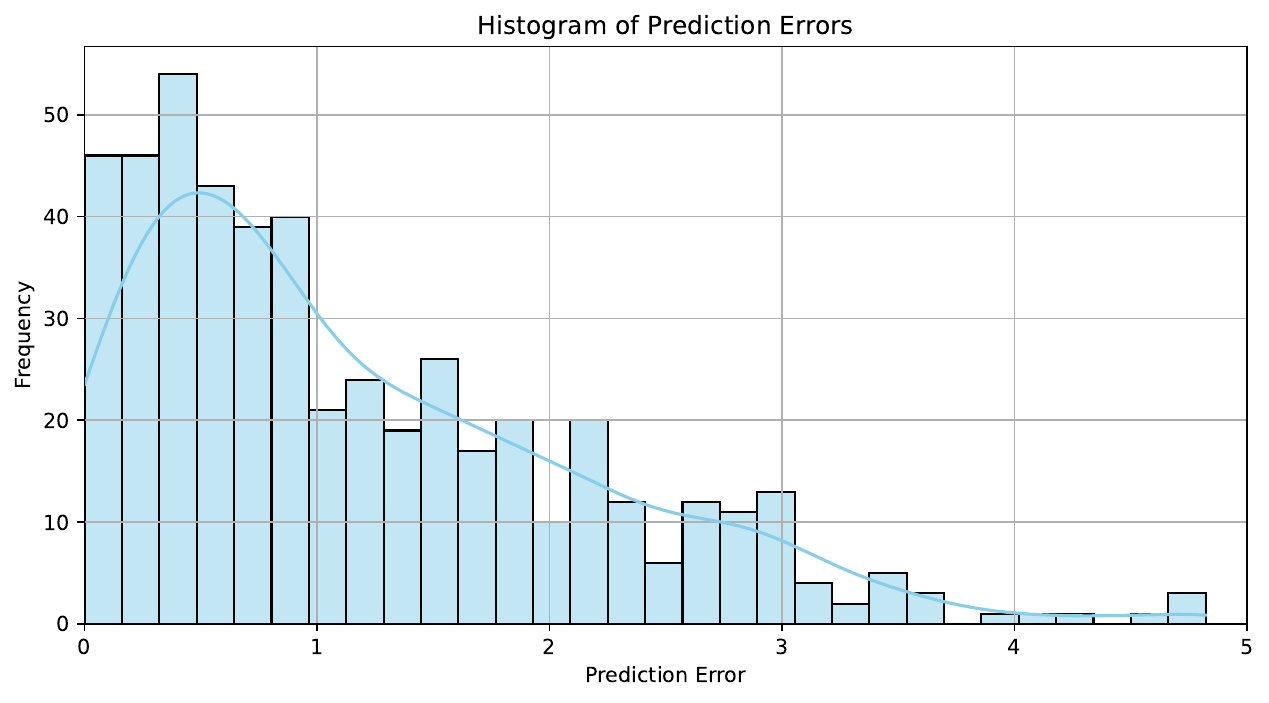}
        \caption{Erdős-Rényi Model}
    \end{subfigure}
    \hfill 
    \begin{subfigure}[t]{0.32\textwidth}
        \centering
        \includegraphics[width=\linewidth]{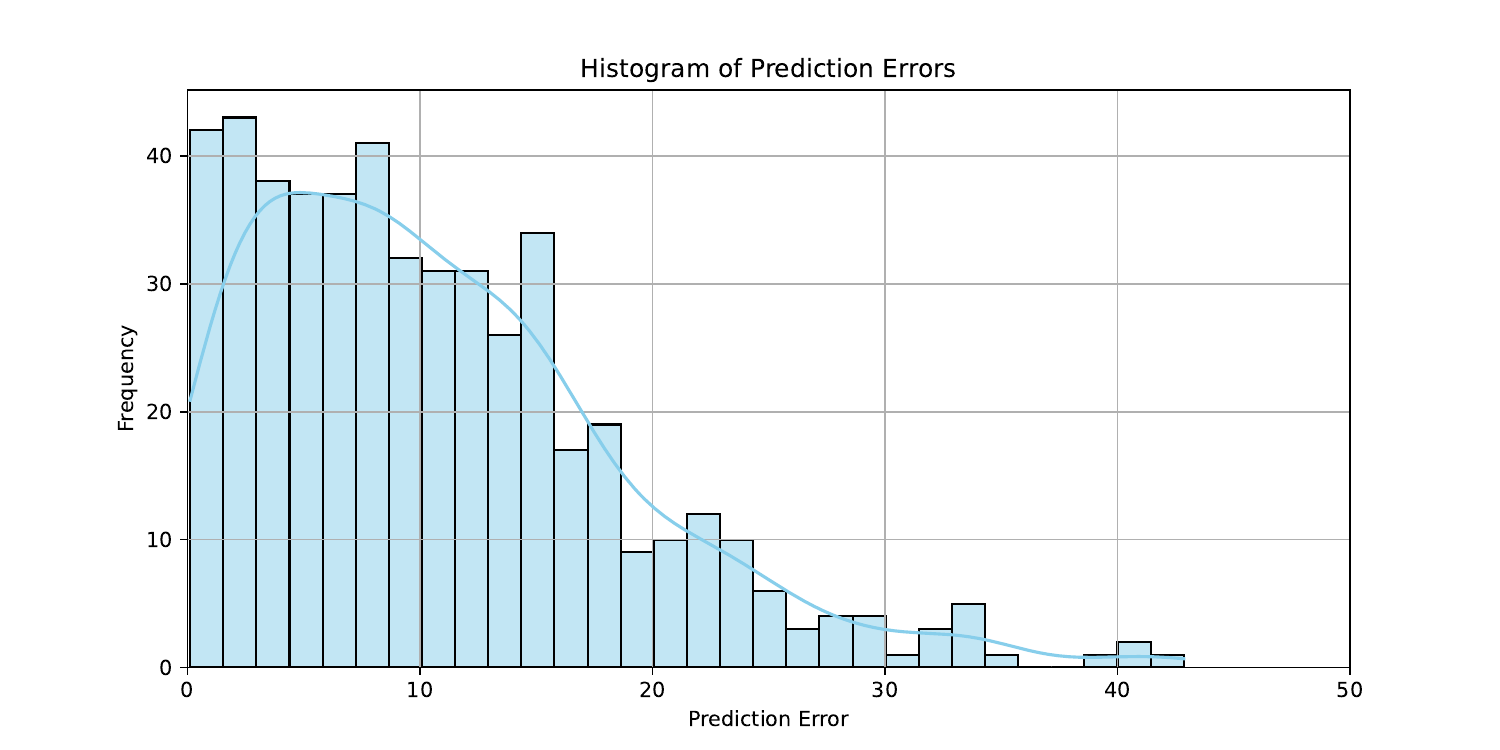}
        \caption{Barabási-Albert Model}
    \end{subfigure}
    \hfill 
    \begin{subfigure}[t]{0.32\textwidth}
        \centering
        \includegraphics[width=\linewidth]{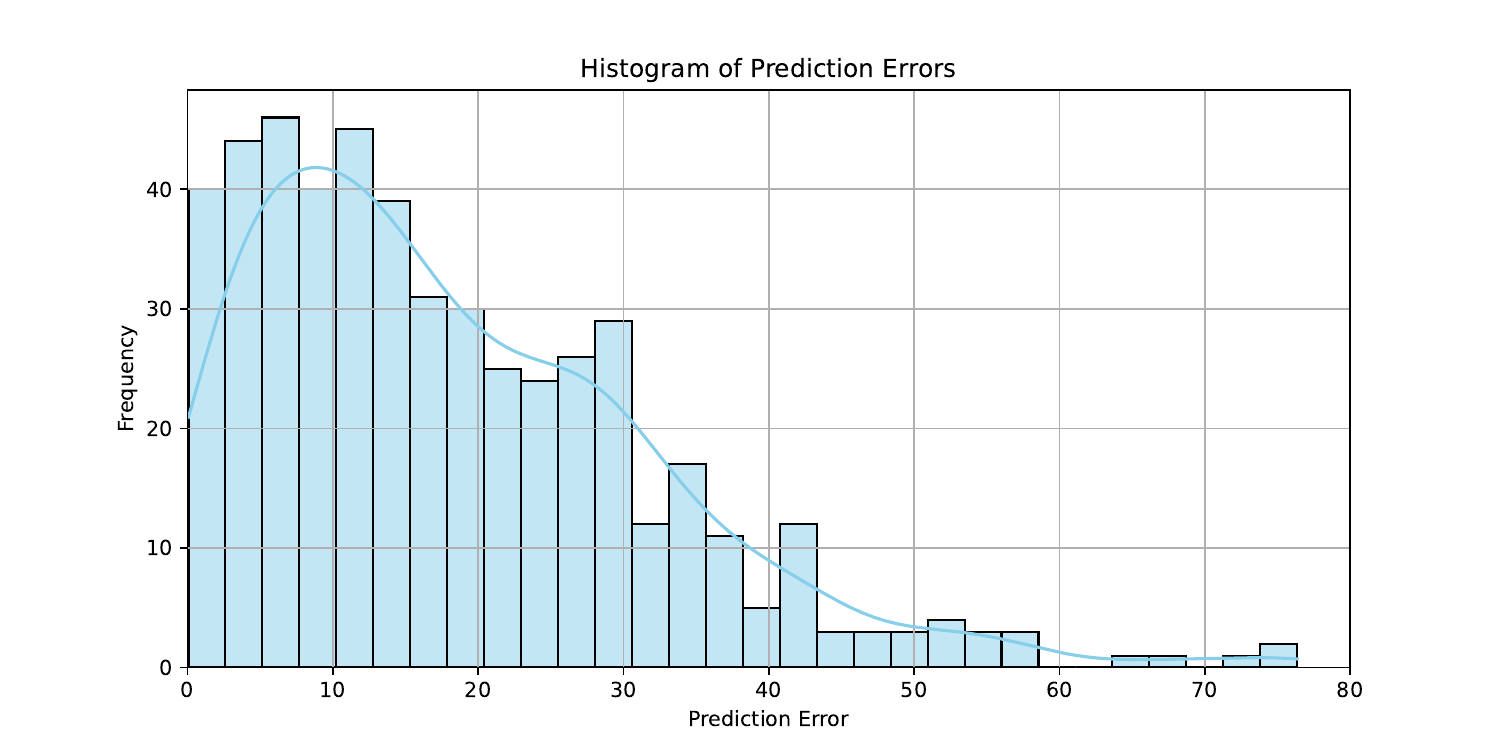}
        \caption{Watts-Strogatz Model}
    \end{subfigure}
    
    \caption{Predictive Accuracy Across Different Graph Topologies.}
    \label{fig:spagan_2}
\end{figure*}

In our framework, SPAGAN model ($\mathfrak{F}_{\boldsymbol{\theta}}$) plays a vital role in Forge, where it serves as a fast and accurate estimator of shortest-path costs. These estimates directly support the Predictive Path Stressing (PPS) algorithm in generating initial feasible perturbation vectors without requiring full shortest-path solver calls. Therefore, we want to evaluate the generalization capability of SPAGAN, by designing two sets of experiments using synthetic graphs. The first examines how well the model scales across different graph sizes, while the second investigates its performance across distinct graph structures.

To test scalability, we trained SPAGAN on Erdős–Rényi (ER) graphs with 1,000 nodes at maximum density, and then evaluated its predictions on ER graphs with larger sizes—2,000 and 3,000 nodes—also at maximum density. Figures~\ref{fig:spagan_1} and \ref{fig:spagan_2} show the results. The scatter plots in Figure~\ref{fig:spagan_1} (a–c) compare SPAGAN’s predictions with ground-truth shortest-path distances obtained via Dijkstra’s algorithm. In all cases, the predicted values closely follow the diagonal (Predicted = True), indicating high prediction accuracy. The corresponding error histograms in Figure~\ref{fig:spagan_2} (a–c) further support this, with prediction errors concentrated around zero and minimal high-magnitude deviations, suggesting both low bias and low variance.

In the second evaluation, to evaluate SPAGAN’s ability to generalize across different graph topologies, we fixed the training on ER graphs with 1,000 nodes and tested the model on two structurally distinct types: Barabási–Albert (scale-free) and Watts–Strogatz (small-world), each with the same node count. As shown in Figures~\ref{fig:spagan_1} and \ref{fig:spagan_2}, SPAGAN’s predictions remain highly accurate. The scatter plots in Figure~\ref{fig:spagan_2} (a–c) continue to show strong alignment between predicted and true path costs, and the histograms in Figure~\ref{fig:spagan_2} (a–c) maintain the same error concentration pattern seen in the size generalization study. Taken two evaluations together, these results demonstrate that SPAGAN is capable of generalizing effectively across both varying graph sizes and structural types. Its predictive accuracy and stability under diverse conditions make it a reliable and efficient surrogate for shortest-path estimation within the broader \PIMMA framework.

\subsection{Robustness to SPAGAN Errors and Exact-Path Safeguard}
\label{app:spagan-robust}

\subsubsection{Exact-Path Feasibility Check and Safeguard (PPS-I)}

Let $G=(V,E)$ be a weighted directed graph, where each edge $e\in E$ has a perturbed weight $f_e(x_e)$ determined by the allocated budget $x_e$ from the perturbation vector $\mathbf{x}\in\mathbb{R}_{\ge0}^{|E|}$.
Given a set of source–target pairs $\mathcal{K}=\{(s_i,t_i)\}_{i=1}^m$ and a threshold $T>0$, 
we define the overall \emph{feasibility rate} of a perturbation $\mathbf{x}$ as
\[
\mathrm{Feas}(\mathbf{x})
\;:=\;
\frac{1}{m}
\sum_{i=1}^m
\mathbf{1}\!\left\{
\operatorname{SP}_G(s_i,t_i;\mathbf{x})
\;\ge\;
T
\right\},
\]
where the shortest-path length under perturbation $\mathbf{x}$ is given by $\operatorname{SP}_G(s,t;\mathbf{x})
\;=\;
\min_{\rho\in\mathcal{P}(s,t)}
\sum_{e\in\rho} f_e(x_e).$ \textbf{PPS-I} takes any candidate $x$ (from PPS or RL) and performs:
(i) compute $d_x(s_i,t_i)$ for all pairs via Dijkstra; 
(ii) while $d_x(s_i,t_i)<T$ for any $i$, identify a shortest path $\rho^\star$ and increment $\{x_e\}_{e\in\rho^\star}$ by the minimum amount to push $d_x(s_i,t_i)$ to $T$.
This yields a final $\hat x$ with $\mathrm{Feas}(\hat x)=1$.

\paragraph{Proposition (Guarantee of PPS-I).}
Let $\mathcal{K}$ be the set of QoSD pairs with threshold $T>0$,
and let the per-pair slack under a perturbation $\mathbf{x}$ be
\[
\operatorname{slack}_{\mathbf{x}}(s,t)=\max\{0,\,T-\operatorname{SP}_G(s,t;\mathbf{x})\},
\]
with total slack $S(\mathbf{x})=\sum_{(s,t)\in\mathcal{K}}\operatorname{slack}_{\mathbf{x}}(s,t)$.
Assume every edge-weight function $f_e(\cdot)$ is non-decreasing and locally Lipschitz, with a positive lower bound on its incremental gain:
\[
f_e(x_e+1)-f_e(x_e)\ge c_{\min}>0.
\]

Then, for any initial perturbation $\mathbf{x}$, PPS-I returns an updated $\hat{\mathbf{x}}$
such that $\operatorname{SP}_G(s,t;\hat{\mathbf{x}})\ge T$ for all $(s,t)\in\mathcal{K}$,
thereby ensuring \textbf{100\% feasibility}.
Moreover, the total additional cost is bounded by
\[
\|\hat{\mathbf{x}}\|_1-\|\mathbf{x}\|_1\le \tfrac{1}{c_{\min}}\,S(\mathbf{x}).
\]

\textbf{Proof.} At each iteration, PPS-I selects a violating pair $(s,t)$ with
$\operatorname{SP}_G(s,t;\mathbf{x})<T$,
extracts its exact shortest path $\rho^\star$ under current $\mathbf{x}$,
and increases budgets on edges $e\in\rho^\star$ by the smallest nonnegative increments
$\Delta x_e$ such that
\[
\sum_{e\in\rho^\star}\!\bigl[f_e(x_e+\Delta x_e)-f_e(x_e)\bigr]
\;\ge\;
T-\operatorname{SP}_G(s,t;\mathbf{x}).
\]
Since each $f_e$ is non-decreasing, all shortest-path lengths are non-decreasing under the update.
Hence the total slack
$S(\mathbf{x})=\sum_{(u,v)\in\mathcal{K}}\max\{0,\,T-\operatorname{SP}_G(u,v;\mathbf{x})\}$
strictly decreases whenever there is a violation.
Because $\mathcal{K}$ is finite and slack is bounded below by $0$, PPS-I must terminate in finitely many steps at some $\hat{\mathbf{x}}$ with $S(\hat{\mathbf{x}})=0$, i.e.,
$\operatorname{SP}_G(s,t;\hat{\mathbf{x}})\ge T$ for all $(s,t)\in\mathcal{K}$.

We now relate the total increase in edge weights to the increase in budgets.
From the discrete lower bound assumption
\[
f_e(x_e+1)-f_e(x_e)\ge c_{\min},
\]
each unit increase of $x_e$ increases the corresponding edge cost by at least $c_{\min}$.  
For multiple steps, we can sum this inequality $\Delta x_e$ times:
\[
\begin{aligned}
f_e(\hat{x}_e)-f_e(x_e)
&=\sum_{j=0}^{\Delta x_e-1}\big[f_e(x_e+j+1)-f_e(x_e+j)\big] \\
&\ge \sum_{j=0}^{\Delta x_e-1} c_{\min}
= c_{\min}(\hat{x}_e-x_e).
\end{aligned}
\]
This simply means that $f_e$ grows at least linearly with slope $c_{\min}$. In other words, the function lies above a line of slope $c_{\min}$ passing through $(x_e,f_e(x_e))$. Summing this inequality over all edges yields
\[
\sum_{e\in E}\big[f_e(\hat{x}_e)-f_e(x_e)\big]\ge c_{\min}\sum_{e\in E}(\hat{x}_e-x_e)
= c_{\min}\|\hat{\mathbf{x}}-\mathbf{x}\|_1.
\]
The left-hand side represents the total increase in edge weights caused by all PPS-I updates.
By construction, this cumulative increase cannot exceed the initial total slack $S(\mathbf{x})$, since $S(\mathbf{x})$ quantifies exactly the total amount of shortest-path length that must be compensated to reach feasibility.
Hence
\[
c_{\min}\|\hat{\mathbf{x}}-\mathbf{x}\|_1 \le S(\mathbf{x})
\quad\Rightarrow\quad
\|\hat{\mathbf{x}}\|_1-\|\mathbf{x}\|_1 \le \tfrac{S(\mathbf{x})}{c_{\min}}.
\]

Dually, the incremental upper bound (by Lipschitz) implies
\[
f_e(\hat{x}_e)-f_e(x_e)\;\le\; C_{\max}\,(\hat{x}_e-x_e)
\quad\Rightarrow\quad
\sum_{e\in E}\bigl[f_e(\hat{x}_e)-f_e(x_e)\bigr]\;\le\; C_{\max}\,\|\Delta\mathbf{x}\|_1.
\]
To eliminate the initial violations, the cumulative increase of shortest-path lengths must be at least the total deficit one needs to fill, which is lower-bounded by the initial total slack:
\[
\sum_{e\in E}\bigl[f_e(\hat{x}_e)-f_e(x_e)\bigr]\;\ge\; S(\mathbf{x}).
\]
Therefore,
\[
S(\mathbf{x})
\;\le\; C_{\max}\,\|\Delta\mathbf{x}\|_1
\quad\Rightarrow\quad
\|\hat{\mathbf{x}}\|_1-\|\mathbf{x}\|_1
\;\ge\; \frac{S(\mathbf{x})}{C_{\max}}.
\]

It gives us the desired sandwich bound
\(
\frac{S(\mathbf{x})}{C_{\max}}
\le
\|\hat{\mathbf{x}}\|_1-\|\mathbf{x}\|_1
\le
\frac{S(\mathbf{x})}{c_{\min}}
\),
together with guaranteed feasibility at termination.

\subsubsection{Robustness to SPAGAN Prediction Noise}
\label{app:noise-sensitivity}

The purpose of this experiment is to evaluate the robustness of \textsc{Hephaestus} under imperfect shortest-path predictions produced by SPAGAN. 
In real-world deployment, SPAGAN may encounter topologies or edge-weight distributions not seen during training, leading to degraded path estimates that could cascade through subsequent stages of the pipeline. 
Hence, it is crucial to quantify how such prediction noise affects overall feasibility, cost efficiency, and the effectiveness of our safeguard module \textsc{PPS-I}. 

To model degraded SPAGAN predictions, we inject independent zero-mean noise with rate $\eta\in\{30\%,10\%,5\%\}$ into SPAGAN-estimated shortest-path scores during \textsc{PPS} and \textsc{RL} decision-making. 
Let $\tilde{d}(s,t)$ denote the SPAGAN-predicted distance, we perturb it as $\tilde{d}_\eta(s,t)=\tilde{d}(s,t)+\epsilon_{s,t}$, 
where $\epsilon_{s,t}$ is sampled to achieve the target noise level (calibrated on the validation split).
We then evaluate:
(i) \textsc{PPS} (greedy optimization on noisy estimates); 
(ii) \textsc{RL} refinement in latent space (also using noisy estimates for reward shaping); and 
(iii) \textsc{PPS-I}, which applies exact post-check correction via Dijkstra to guarantee feasibility. 
We report feasibility rate (\%) over $m=50$ source–target pairs and the total perturbation budget cost.

\begin{table}[htp]
\centering
\caption{Sensitivity to SPAGAN noise and exact-path safeguard (\textsc{PPS-I}). 
We report feasibility (\%) and total cost after each stage; \textsc{PPS-I} also shows wall-clock fix time (s) for the exact Dijkstra-based correction.}
\label{tab:noise-sensitivity}
\resizebox{\linewidth}{!}{
\begin{tabular}{lccccccccc}
\toprule
Dataset & Noise & PPS Feas. & PPS Cost & RL Feas. & RL Cost & PPS-I Feas. & PPS-I Cost & PPS-I Time \\
\midrule
Email  & 30\% & 90.0 & 3284 & 94.0 & 2911 & \textbf{100.0} & 3005 & 20.2 \\
       & 10\% & 94.0 & 3185 & 96.0 & 2759 & \textbf{100.0} & 2802 & 15.7 \\
       &  5\% & 96.0 & 3157 & 98.0 & 2640 & \textbf{100.0} & 2712 &  6.1 \\
       &  0\% & \textbf{100.0} & 3091 & \textbf{100.0} & 2691 & \textbf{100.0} & 2691 &  0.0 \\
\midrule
RoadCA & 30\% & 78.0 & 13730 & 86.0 & 10985 & \textbf{100.0} & 11061 & 214.8 \\
       & 10\% & 88.0 & 12556 & 92.0 & 10148 & \textbf{100.0} & 10795 & 137.2 \\
       &  5\% & 92.0 & 12123 & 94.0 &  9892 & \textbf{100.0} &  9907 &  48.6 \\
       &  0\% & \textbf{100.0} & 11283 & \textbf{100.0} &  9184 & \textbf{100.0} &  9277 &   0.0 \\
\bottomrule
\end{tabular}
}
\end{table}

From Table~\ref{tab:noise-sensitivity} (with $m{=}50$ source–target pairs), we observe four key findings:

\begin{enumerate}[leftmargin=*]
    \item \textbf{PPS degrades under noisy predictions:} Feasibility drops by up to 22\% (e.g., \textit{RoadCA} with 30\% noise), and cost increases by about 20\% as perturbation budgets are misallocated to irrelevant edges.
    \item \textbf{RL refinement recovers performance:} RL improves feasibility and reduces cost over PPS by leveraging gradient signals in latent space. For instance, on the \textit{RoadCA} dataset with 30\% noise, RL improves feasibility from 78\% to 86\% and reduces cost from 13{,}730 to 10{,}985.
    \item \textbf{PPS-I ensures 100\% feasibility in all cases:} Regardless of SPAGAN noise level, \textsc{PPS-I} consistently restores feasibility via exact Dijkstra-based correction. This confirms the pipeline’s correctness guarantee.
    \item \textbf{Cost penalty of PPS-I is modest:} Even in the worst case (\textit{RoadCA}, 30\% noise), the \textsc{PPS-I} cost increases by only $\sim$19\% relative to the ideal (0\% noise) case i.e., 11{,}061 vs.\ 9{,}277, while feasibility remains perfect.
\end{enumerate}

\subsection{Resilience to Weak Initial Data}
\label{app:weak-data}
 
This experiment aims to assess the system’s ability to recover from weak or suboptimal initial data and to validate whether the latent-space refinement mechanism truly corrects, rather than amplifies, early-stage imperfections. 
In other words, we examine the self-reinforcing capability of the Morph–Refine loop under noisy or incomplete initialization. Our framework is explicitly designed to mitigate the influence of suboptimal or noisy solutions generated in the early phase. 
Unlike heuristic post-processing, our refinement operates entirely in the latent space, where a reinforcement-learning (RL) agent optimizes a differentiable reward that reflects both feasibility and cost efficiency. 
Consequently, even if the initial perturbation samples from the \textsc{PPS} stage are imperfect, the refinement progressively corrects them rather than propagating the errors.

To validate this behavior, we conduct a cycle-level robustness experiment. 
After each refinement cycle, the top-$k$ highest-reward latent samples are added back to the pre-trained solution set $\mathcal{D}_{\text{sol}}$, forming a self-reinforcing feedback loop between the Morph (Mix-CVAE) and Refine (RL) phases. 
This feedback mechanism provides two benefits:
\begin{enumerate}[label=(\roman*)]
    \item Mix-CVAE retraining incorporates improved latent samples, enriching the coverage of feasible solution modes.
    \item The RL policy learns on an increasingly structured latent manifold, enabling it to discover even better perturbations in subsequent cycles.
\end{enumerate}

On \textit{Email} with $T{=}260\%$, the improvement saturates after roughly five refinement cycles, yielding a total cost reduction of $\sim18.5\%$.  
On the larger-scale \textit{RoadCA} dataset, cost reductions continue to increase even after eight cycles, peaking at over $\sim26\%$ under high thresholds.  
These results confirm that \textbf{\textsc{Hephaestus} remains robust to weak or noisy initial data}, and that the latent-space feedback loop effectively improves data quality over successive cycles.

Table~\ref{tab:weak-data} reports the relative cost reduction (\%) across refinement cycles under varying feasibility thresholds $T$. 
Both \textit{Email} and \textit{RoadCA} datasets exhibit clear monotonic improvements, showing that later cycles consistently achieve higher-quality perturbations.

\begin{table}[htp]
\centering
\caption{Cycle-level improvement in cost reduction (\%) across thresholds $T$. 
The values represent budget savings relative to the initial solution (Cycle 0).}
\label{tab:weak-data}
\resizebox{0.9\linewidth}{!}{
\begin{tabular}{lccccc}
\toprule
Dataset & Cycle & $T{=}140\%$ & $T{=}180\%$ & $T{=}220\%$ & $T{=}260\%$ \\
\midrule
Email  & 0 & 0.00 & 0.00 & 0.00 & 0.00 \\
       & 1 & 4.01 & 5.25 & 6.82 & 7.95 \\
       & 2 & 8.67 & 10.15 & 12.44 & 14.21 \\
       & 3 & 10.86 & 12.50 & 15.03 & 17.88 \\
       & 4 & 11.07 & 12.97 & 15.58 & 18.04 \\
       & 5 & 11.23 & 13.11 & 15.98 & 18.52 \\
\midrule
RoadCA & 0 & 0.00 & 0.00 & 0.00 & 0.00 \\
       & 1 & 7.01 & 8.55 & 10.11 & 11.58 \\
       & 2 & 10.91 & 12.98 & 15.03 & 17.34 \\
       & 3 & 12.83 & 15.21 & 17.85 & 20.19 \\
       & 4 & 15.67 & 17.89 & 20.45 & 22.87 \\
       & 5 & 17.42 & 19.53 & 22.10 & 24.53 \\
       & 6 & 18.27 & 20.33 & 22.98 & 25.66 \\
       & 7 & 18.51 & 20.76 & 23.42 & 26.09 \\
       & 8 & 18.61 & 20.89 & 23.54 & 26.21 \\
\bottomrule
\end{tabular}}
\end{table}

\begin{figure*}[htp]
    \centering
        \includegraphics[width=\linewidth]{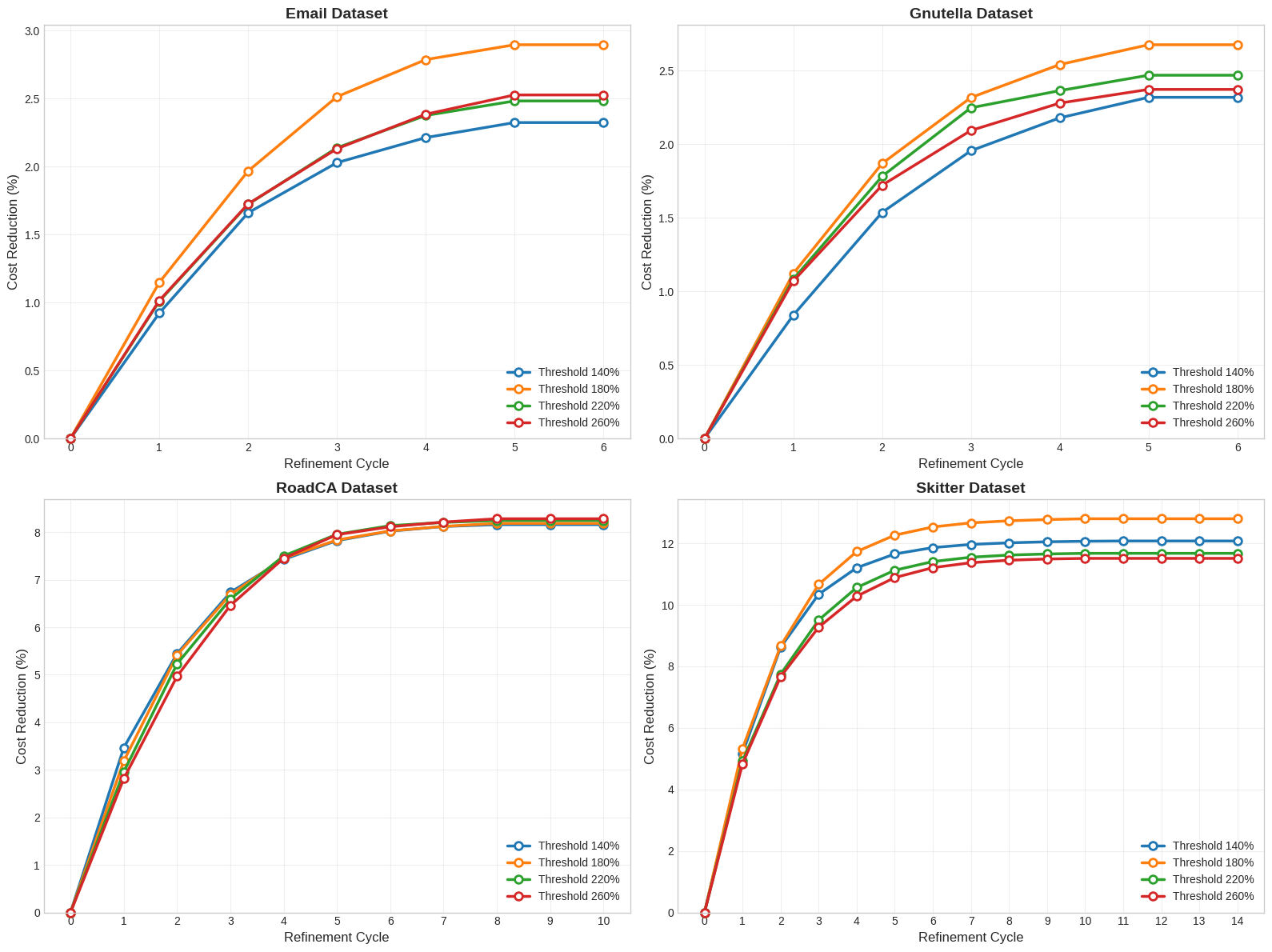}
    \caption{Evolution of Total Budget during the Refine phase (RL agent training)}
\label{fig:rl_improve}
\end{figure*}

\subsubsection{RL Convergence Behavior under Different Problem Scales}
\label{app:rl-convergence}

To analyze how the RL policy behaves across different problem scales, we provide detailed training statistics on two real-world datasets: \textit{Email} (1005 nodes, 25{,}571 edges) and \textit{RoadCA} (1.96M nodes, 2.77M edges). Both experiments are conducted using clean SPAGAN predictions to isolate convergence dynamics.

\begin{table}[h!]
\centering
\caption{Training-phase convergence statistics on datasets of different scales. We report feasibility, expected total budget, and reward over three distinct training stages.}
\label{tab:rl-convergence}
\resizebox{\textwidth}{!}{%
\begin{tabular}{@{}l lll lll@{}}
\toprule
& \multicolumn{3}{c}{\textbf{Email}} & \multicolumn{3}{c}{\textbf{RoadCA}} \\
\cmidrule(lr){2-4} \cmidrule(lr){5-7}
\textbf{Training Phase} & \textbf{Feas. (\%)} & \textbf{Expected Total Budget} & \textbf{Reward} & \textbf{Feas. (\%)} & \textbf{Expected Total Budget} & \textbf{Reward} \\
\midrule

\addlinespace
\multicolumn{7}{l}{\textbf{Stage 1: Feasibility Search Behavior}} \\
Episodes 0 - 5K & $80.0 \to 100$ & $1921.65 \to 15147.21$ & $201.44 \to 551.95$ & & & \\
Episodes 0 - 10K & & & & $40.0 \to 100$ & $2641.58 \to 64453.49$ & $545.37 \to 2348.62$ \\
\addlinespace
\multicolumn{7}{l}{\textbf{Stage 2: Cost Optimization Behavior}} \\
Episodes 5K - 15K & $98.0 \pm 2$ & $15147.21 \to 2697.84$ & $551.95 \to 874.72$ & & & \\
Episodes 10K - 35K & & & & $98.0 \pm 2$ & $64453.49 \to 9265.72$ & $2348.62 \to 3975.03$ \\
\addlinespace
\multicolumn{7}{l}{\textbf{Stage 3: Convergence}} \\
Episodes 15K - 50K & $98.0 \pm 2$ & $2697.84 \pm 17$ & $874.72 \pm 14$ & & & \\
Episodes 35K - 50K & & & & $98.0 \pm 2$ & $9265.72 \pm 49$ & $3975.03 \pm 37$ \\
\bottomrule
\end{tabular}%
}
\end{table}
As shown in Table~\ref{tab:rl-convergence}, training proceeds in three stages based on episode count. In Stage 1 (Episodes 0–5K / 0–10K), the policy rapidly improves feasibility (maximizing term 1 of reward function) from 80\% to 100\% (Email) and from 40\% to 100\% (RoadCA), demonstrating its ability to learn constraint satisfaction from scratch. In Stage 2, the focus shifts to cost optimization (term 2 of reward function): total budget drops significantly from 15147.21 to 2697.84 on Email, and from 64453.49 to 9265.72 on RoadCA, while feasibility remains approximately 98\%. Finally, in Stage 3, the reward and cost metrics converge with low variance, e.g., on RoadCA, reward stabilizes at 3975.03 ± 37 and budget at 9265.72 ± 49, confirming stable convergence even in large-scale settings.

\subsection{Comparison with Alternative Latent Optimization Strategies}
\label{app:latent-optimization}

A central motivation for the \textsc{Hephaestus} framework is that reinforcement learning (RL) can effectively optimize latent representations beyond what static or population-based methods can achieve. 
To verify this, we conduct controlled comparisons against two alternative latent optimization strategies: Bayesian Optimization (BO) and Evolutionary Strategies (ES). 
This experiment aims to evaluate whether the RL-based refinement phase indeed provides superior sample efficiency, scalability to higher latent dimensions, and robustness to noisy SPAGAN predictions. 

Each method is evaluated on the \textit{Gnutella} dataset under varying latent dimensionalities $d$, reporting the final perturbation cost (lower is better) and feasibility rate (\%). 
All evaluations are conducted using clean SPAGAN predictions (i.e., without additional noise) for a fair comparison.

\begin{table}[htp]
\centering
\caption{Comparison between RL-based refinement and alternative latent-space optimization strategies. 
Lower cost indicates better efficiency.}
\label{tab:latent-optimization}
\resizebox{\linewidth}{!}{
\begin{tabular}{lccc}
\toprule
\textbf{Method} & \textbf{Latent Dim. $d$} & \textbf{Final Cost $\downarrow$} & \textbf{Feas. Rate $\uparrow$} \\
\midrule
PPS (no refinement) & -- & 4118 & 100\% \\
\midrule
RL (Ours) & 16 & 3435 & 100\% \\
           & 64 & \textbf{3419} & \textbf{100\%} \\
\midrule
Bayesian Opt. (BO) & 16 & 3590 & 98\% \\
                   & 28 & 4055 & 52\% \\
\midrule
Evolutionary Strat. (ES) & 16 & 3612 & 94\% \\
                         & 64 & 3598 & 98\% \\
\bottomrule
\end{tabular}}
\end{table}

\begin{table}[htp]
\centering
\caption{Comparison of RL and ES under different levels of SPAGAN noise. 
All experiments are conducted on the \textit{Gnutella} dataset with latent dimensions $d\in\{16,64\}$. 
Lower cost indicates better performance, and higher feasibility denotes stronger robustness.}
\label{tab:rl-vs-es-noisy}
\resizebox{\linewidth}{!}{
\begin{tabular}{lcccc}
\toprule
\textbf{Method} & \textbf{Latent Dim. $d$} & \textbf{SPAGAN Noise (\%)} & \textbf{Final Cost $\downarrow$} & \textbf{Feas. Rate $\uparrow$} \\
\midrule
PPS & -- & 0 & 4118 & 100\% \\
    &    & 5 & 4221 & 96\% \\
    &    & 10 & 4318 & 92\% \\
    &    & 30 & 4392 & 86\% \\
\midrule
RL (Ours) & 16 & 0 & 3435 & 100\% \\
           &    & 5 & 3596 & 98\% \\
           &    & 10 & 3714 & 92\% \\
           &    & 30 & 3832 & 86\% \\
\cmidrule(lr){2-5}
           & 64 & 0 & \textbf{3419} & \textbf{100\%} \\
           &    & 5 & 3550 & 98\% \\
           &    & 10 & 3668 & 96\% \\
           &    & 30 & 3775 & 92\% \\
\midrule
Evo.\ Strat.\ (ES) & 16 & 0 & 3612 & 94\% \\
                   &    & 5 & 3798 & 91\% \\
                   &    & 10 & 3945 & 87\% \\
                   &    & 30 & 4201 & 81\% \\
\cmidrule(lr){2-5}
                   & 64 & 0 & 3598 & 98\% \\
                   &    & 5 & 3756 & 95\% \\
                   &    & 10 & 3912 & 91\% \\
                   &    & 30 & 4187 & 85\% \\
\bottomrule
\end{tabular}}
\end{table}

We observe that BO performs well only in low-dimensional latent spaces ($d\le30$) but suffers from severe sample inefficiency and surrogate modeling errors as dimensionality increases. 
ES scales better but converges slowly, often resulting in suboptimal costs. 
In contrast, our RL-based refinement achieves both the lowest final cost and full constraint satisfaction (100\% feasibility) across all latent dimensions. 
This confirms that the policy-gradient–based refinement is more robust and sample-efficient than model-based (BO) or population-based (ES) strategies, especially in high-dimensional latent manifolds. 

\noindent\textbf{Robustness under Noisy SPAGAN Predictions.} 
To further evaluate the resilience of these methods, we extend the comparison to noisy settings by injecting controlled levels of perturbation into SPAGAN’s path predictions (5\%, 10\%, and 30\%). 
This additional test closes the loop with the robustness discussion in Section~\ref{app:noise-sensitivity} and examines how optimization strategies behave when their latent objectives are corrupted by upstream model uncertainty.

From these results, we observe that Evolutionary Strategies are more sensitive to SPAGAN prediction noise, 
likely due to their strong reliance on the quality of the pre-trained solution set $\mathcal{D}^{\mathrm{sol}}$ generated during the Forge phase. 
In contrast, our RL-based refinement maintains both higher feasibility and lower cost under noise, 
demonstrating stronger robustness across latent dimensions and noise levels.

\subsection{Energy Distribution Convergence during Minimax Training} \label{appendix:ebm_mixcvae_training}

\begin{figure*}[htp]
    \centering
    \begin{subfigure}[t]{0.19\linewidth}
        \centering
        \includegraphics[width=\linewidth]{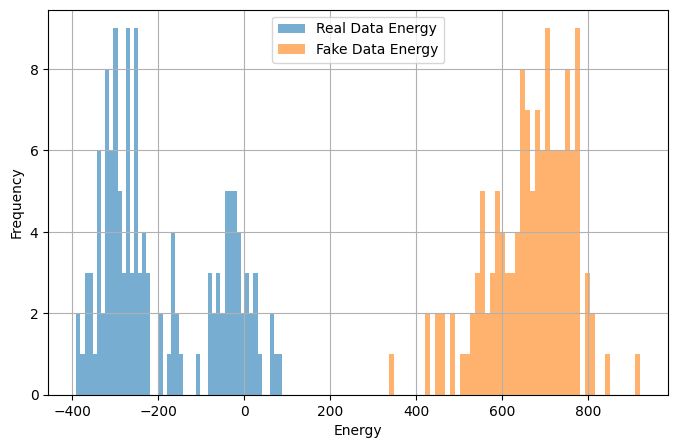}
        \caption{Step 8}
    \end{subfigure}
    \hfill
        \begin{subfigure}[t]{0.19\linewidth}
        \centering
        \includegraphics[width=\linewidth]{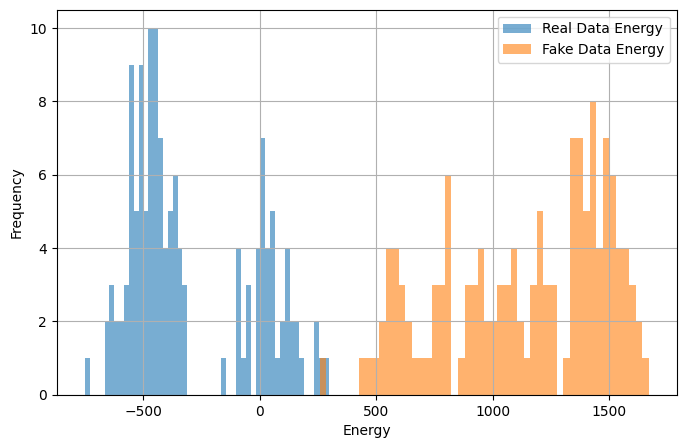}
        \caption{Step 32}
    \end{subfigure}
    \hfill
    \begin{subfigure}[t]{0.19\linewidth}
        \centering
        \includegraphics[width=\linewidth]{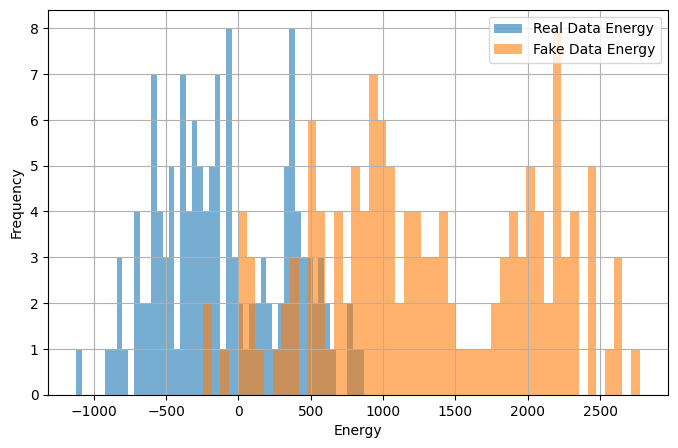}
        \caption{Step 64}
    \end{subfigure}
    \hfill
    \begin{subfigure}[t]{0.19\linewidth}
        \centering
        \includegraphics[width=\linewidth]{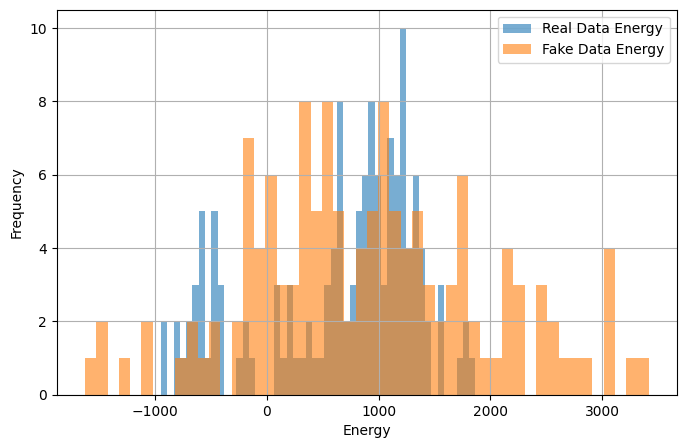}
        \caption{Step 128}
    \end{subfigure}
    \hfill
    \begin{subfigure}[t]{0.19\linewidth}
        \centering
        \includegraphics[width=\linewidth]{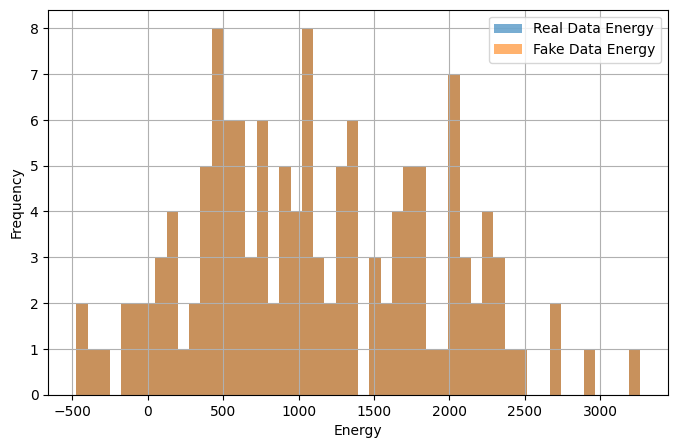}
        \caption{Step 256}
    \end{subfigure}
    \caption{Energy histograms comparing real versus fake data across minimax training. The x-axis denotes the energy score assigned by the Energy-Based Model (EBM), while the y-axis indicates the frequency of samples observed at each energy level. Real data samples (blue), drawn from the ground-truth dataset $\mathfrak{D}^{\text{sol}}$, typically have low energy values and cluster near zero on the left of the x-axis, reflecting the EBM's preference for them. Fake data samples (orange), generated by the Mix-CVAE $\Omega$, initially have higher energies but gradually shift leftward as training progresses. This convergence of the two distributions indicates that $\Omega$ is learning to generate samples that align more closely with the EBM’s learned energy landscape.}
    \label{fig:Energy_histograms_comparing_real}
\end{figure*}

The objective of this experiment is to evaluate how well the generative model Mix-CVAE ($\Omega$) can progressively align its generated samples with the energy distribution implicitly defined by the Energy-Based Model (EBM) during adversarial co-training. Specifically, we aim to assess whether, as training proceeds, $\Omega$ learns to generate samples that lie in low-energy regions—those that the EBM assigns to real data. Since the true data distribution is unknown and cannot be visualized directly, we rely on the EBM's energy outputs as a surrogate for this alignment. The hypothesis is that if $\Omega$ successfully learns the real data distribution, the energy distributions of fake (generated) and real samples will gradually converge.

Figure \ref{fig:Energy_histograms_comparing_real} illustrates this convergence process by plotting energy histograms over five key training steps. Real data from the dataset $\mathfrak{D}^{\text{sol}}$ are shown in blue (low energy region on left-side in x-axis), while fake data sampled from $\Omega$ are shown in orange on the right side. At early stages such as Step 8 and Step 32, there is a significant contrast: the EBM assigns low energy to real samples, which are densely concentrated near the left of the histogram, while fake samples occupy much higher energy regions, reflecting their low realism. As training progresses (Steps 64 and 128), the two distributions begin to overlap, indicating that $\Omega$ is learning to produce more realistic outputs that better match the EBM's learned energy profile. By Step 256, the energy distributions of real and fake samples nearly coincide, suggesting that $\Omega$ has successfully learned to generate samples that the EBM considers indistinguishable from real data. This result highlights a key insight: although the EBM is explicitly trained to minimize energy for real data and maximize it for generated data, the Mix-CVAE generator gradually catches up. Initially, it produces unrealistic, high-energy samples, but through adversarial feedback, it learns to synthesize low-energy (high-quality) solutions. Thus, we demonstrates that the generator effectively "chases" the moving energy boundary set by the EBM and ultimately converges to regions of high data likelihood.

\subsection{Latent Space Visualization} \label{appendix:latent_space}

\begin{figure*}[htp]
    \centering
    \includegraphics[width=1.0\linewidth]{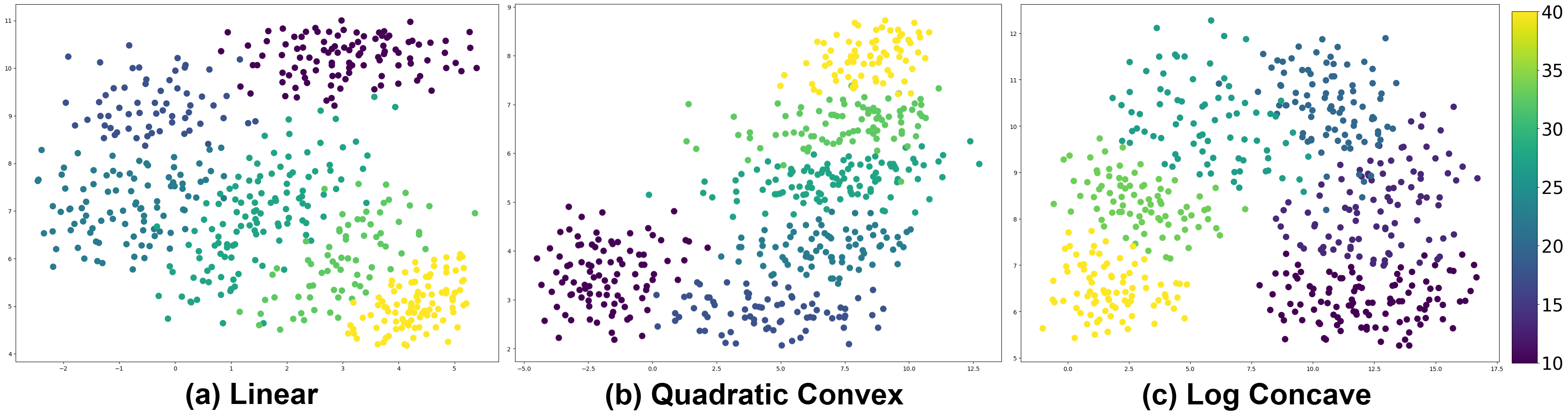} 
    \caption{Conditional Latent Space Visualization via UMAP for the trained Mix-CVAE, conducted on the same synthetic graph but under varying threshold values $T$ across different edge weight functions: Linear, Quadratic (Convex), and Log-Concave. Each point represents a latent vector $\mathbf{z}$ corresponding to a solution, colored by the associated threshold $T$. The structure of the latent space reveals how the model differentiates solution representations under changing constraints and cost dynamics, with clustering patterns indicating sensitivity to the underlying threshold conditions.}
    \label{fig:latent_space_synthetic}
\end{figure*}

\begin{figure*}[htp]
    \centering
    \includegraphics[width=0.95\linewidth]{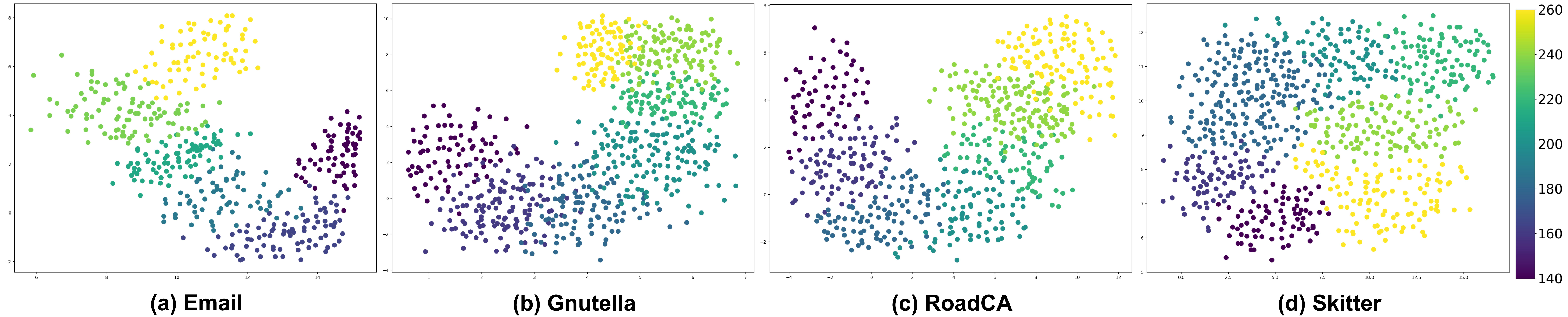} 
    \caption{Conditional Latent Space Visualization for the trained Mix-CVAE under the Linear weight function setting. Each point represents a latent vector $\mathbf{z}$ corresponding to a generated solution, colored by either the input graph type or the threshold $T$. The presence of distinct clusters or separable regions in the latent space suggests that the model captures meaningful variations related to graph structure and problem-specific constraints, indicating successful conditional encoding.}
    \label{fig:latent_space_real}
\end{figure*}

In this experiment, the goal is to assess whether the latent space learned by the conditional generative model (Mix-CVAE) meaningfully captures different constraints $T$ on the same graph. Specifically, we visualize the latent vectors $\mathbf{z}$ using UMAP projections to examine whether different threshold values $T$ result in separable or clustered embeddings—an indicator of successful conditional representation learning. Figure \ref{fig:latent_space_synthetic} shows the latent space organization when Mix-CVAE is trained and evaluated on the same synthetic graph but under three different edge weight functions: Linear, Quadratic (Convex), and Log-Concave. Across all three subfigures, we observe clear gradient-based separation and distinct clusters aligned with increasing values of $T$. This pattern suggests that the generative model is sensitive to threshold constraints and learns to organize the latent space accordingly. Interestingly, the separation structure varies with the cost function type: Linear produces smooth band-like transitions, while Log-Concave shows more localized clustering, potentially reflecting its sharper cost escalation characteristics.

Figure \ref{fig:latent_space_real} complements this analysis by testing Mix-CVAE on real-world graphs (Email, Gnutella, RoadCA, Skitter) under the same linear edge cost setting. Again, latent vectors are visualized and colored by threshold. Across different graph topologies, we still consistently observe separable clusters and gradual transitions in $\mathbf{z}$-space with respect to $T$, indicating that the generative model generalizes across input graphs and captures structural information relevant to feasibility under constraints. In particular, datasets with greater structural diversity (e.g., Skitter and RoadCA) exhibit more complex spatial patterns, highlighting the model’s ability to encode both graph topology and constraint semantics.



\subsection{Impact of Expert Addition in Mix-CVAE} \label{appendix:expert_addition}

The goal of this experiment is to evaluate the effect of adding more experts in the Mix-CVAE architecture on modeling capability and final solution quality. Figures \ref{fig:Comparison1}, \ref{fig:latent_space}, and \ref{fig:expertaddingrelativecost} together provide a comprehensive view of how the number of experts (3, 5, 7, 9) influences the generative model's ability to approximate the true solution distribution and optimize budget outcomes.

\begin{figure*}[htp]
    \centering
    \includegraphics[width=1.0\linewidth]{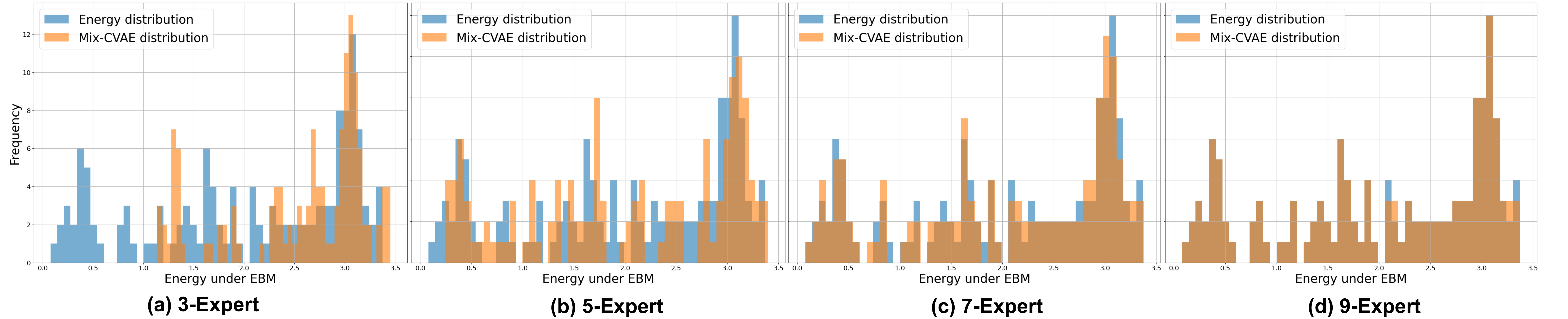}
    \caption{Comparison between the distribution from the EBM and the distribution from the Mix-CVAE with varying numbers of experts on a \textit{synthetic network} with maximum density. Increasing the number of experts improves mode coverage and alignment with the target EBM distribution.}
    \label{fig:Comparison1}
\end{figure*}

\begin{figure*}[htp]
    \centering
    \includegraphics[width=1.0\linewidth]{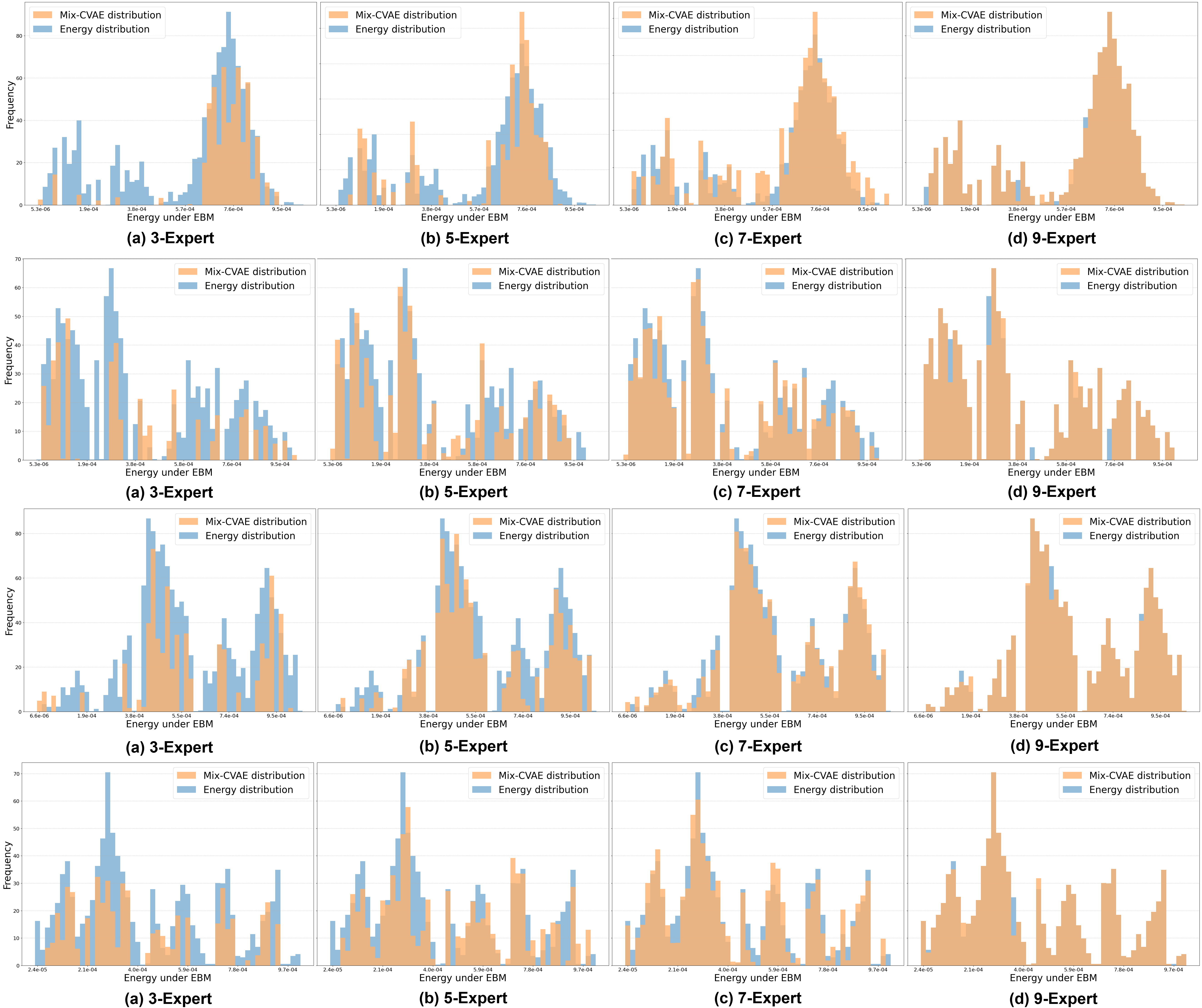} 
    \caption{Impact of adding experts on \textit{real-world networks}. From top to bottom: results on Email, Gnutella, RoadCA, and Skitter networks.}
    \label{fig:latentdistributionreal}
\end{figure*}

\textbf{Impact of Expert Addition on Generative Modeling.} Figures \ref{fig:Comparison1} and \ref{fig:latentdistributionreal} compare the energy distributions of real solutions (as evaluated under the EBM) and fake samples generated by the Mix-CVAE across both synthetic and real network settings. When only 3 experts are used, the generated samples show poor alignment with the real data distribution, evidenced by noticeable discrepancies between the orange (Mix-CVAE) and blue (EBM) bars. As the number of experts increases, this alignment improves significantly—indicating that the model is better able to capture the underlying modes and structure of the target distribution. This validates the core motivation behind expert addition: increasing the number of experts enhances the model’s capacity to represent multimodal solution spaces. Furthermore, Figure \ref{fig:latent_space} provides additional insight by visualizing the latent space of the trained Mix-CVAE via UMAP on synthetic graphs. Each point represents a latent vector $\mathbf{z}$ colored by its corresponding threshold $T$. With more experts, the latent space becomes more organized, exhibiting distinct clusters that correspond to different threshold levels. This structured separation suggests that the model is learning to conditionally represent diverse solution modes, improving both interpretability and the quality of generated samples.

\begin{figure*}[htp]
    \centering
    \includegraphics[width=1.0\linewidth]{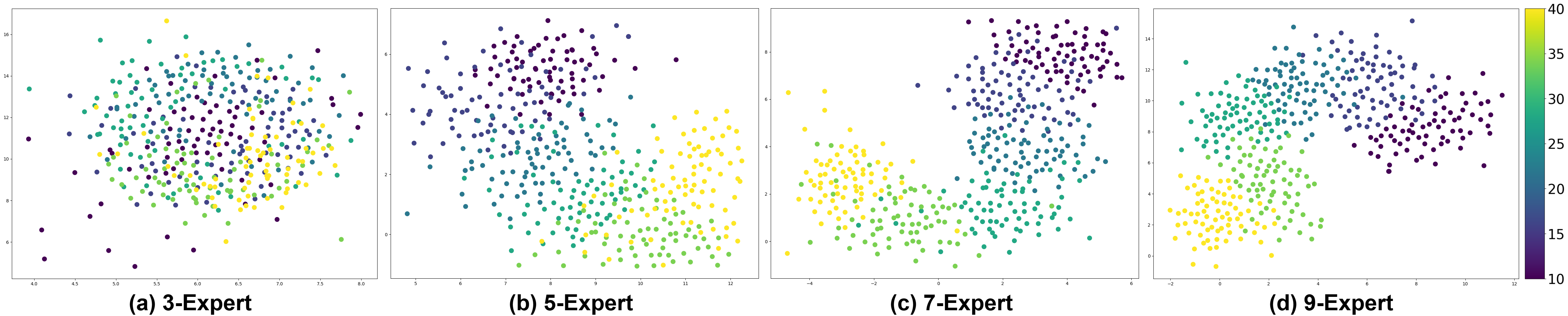} 
    \caption{An example of conditional latent space visualization for the trained Mix-CVAE on the same synthetic graph and pairs but different thresholds. Points represent latent vectors $\mathbf{z}$, colored by threshold $T$. Clear clustering shows the latent space captures meaningful patterns.}
    \label{fig:latent_space}
\end{figure*}

\begin{figure*}[htp]
    \centering
    \captionsetup[subfigure]{skip=0.1pt}  
    \begin{subfigure}[t]{0.32\linewidth}
        \centering
        \includegraphics[width=\linewidth]{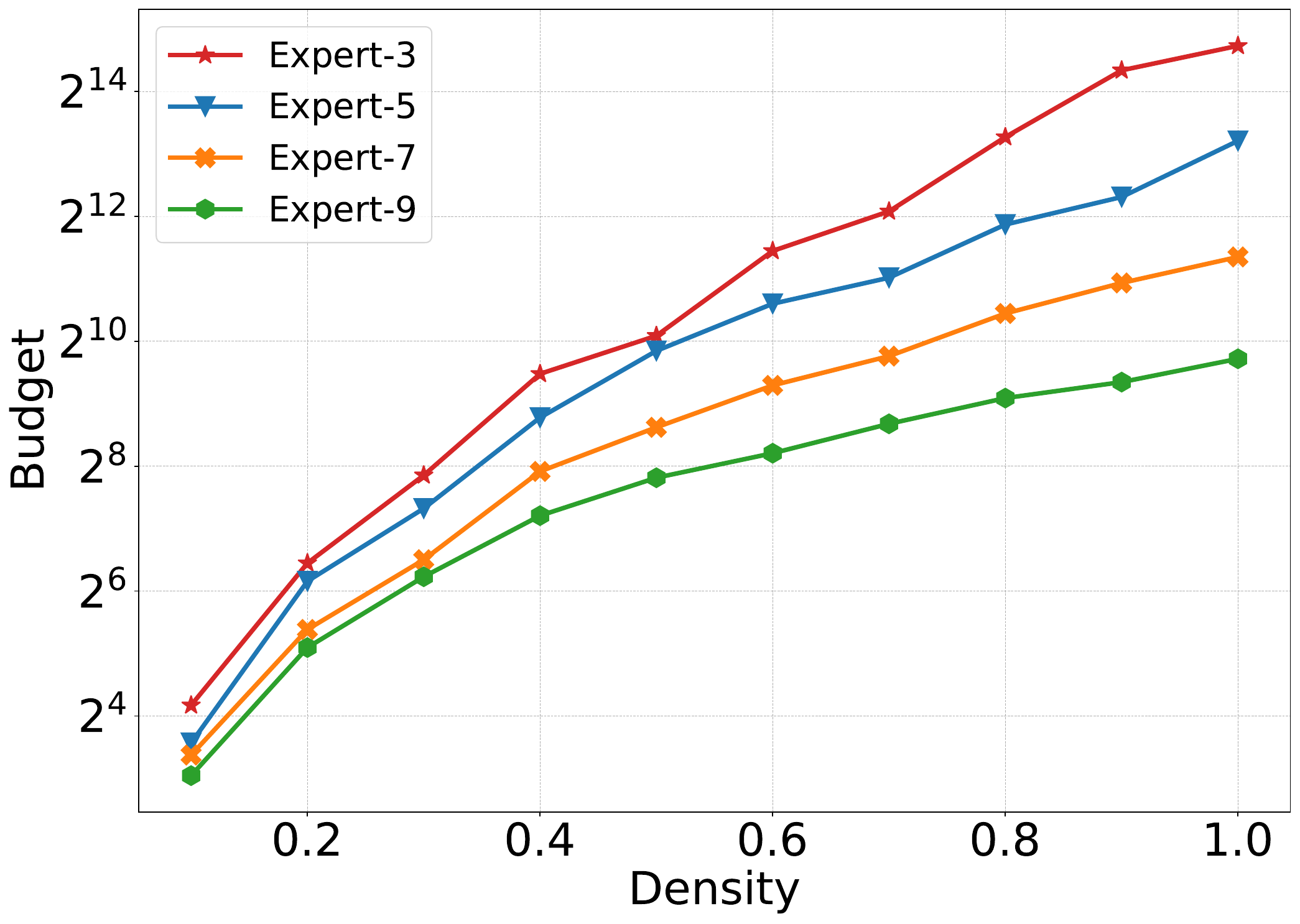}
        \caption{Linear}
    \end{subfigure}
    \hfill
    \begin{subfigure}[t]{0.32\linewidth}
        \centering
        \includegraphics[width=\linewidth]{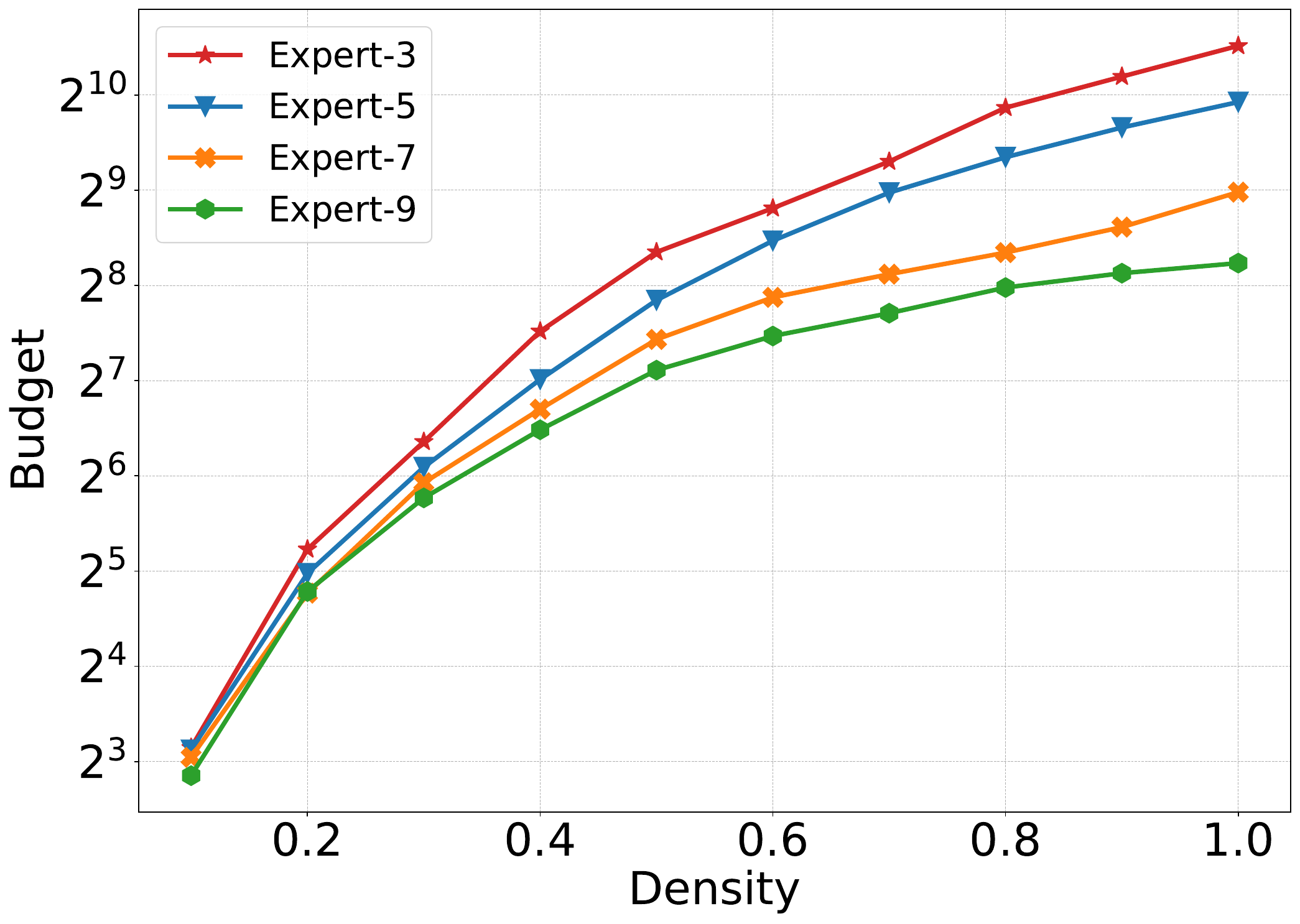}
        \caption{Quadratic Convex}
    \end{subfigure}
    \hfill
    \begin{subfigure}[t]{0.32\linewidth}
        \centering
        \includegraphics[width=\linewidth]{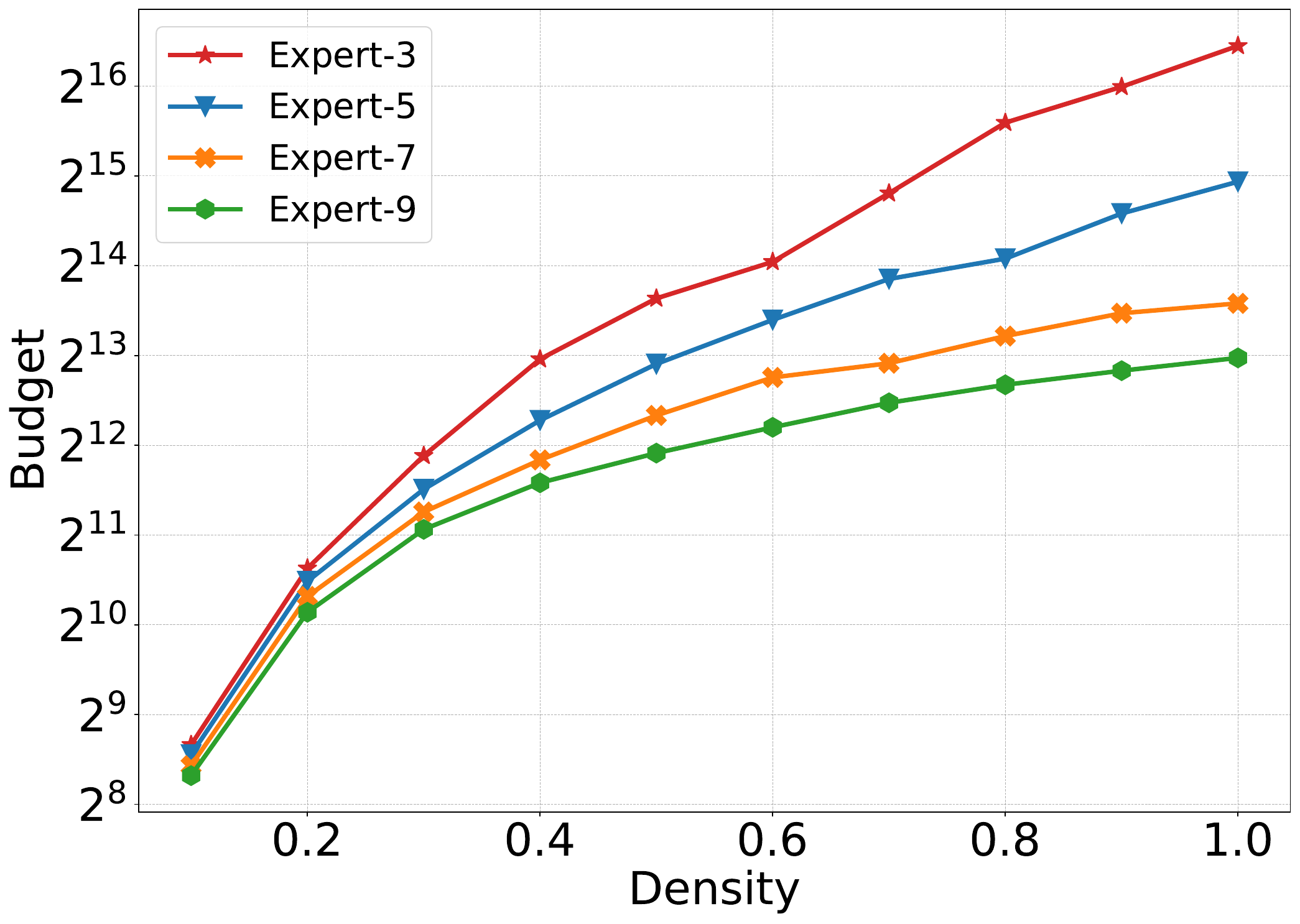}
        \caption{Log Concave}
    \end{subfigure}
    \vspace{-5pt}  
    \caption{Impact of adding of experts (3, 5, 7, and 9) on the resulting total budget across varying graph densities, testing on the synthetic graph under (a) \emph{Linear}, (b) \emph{Quadratic Convex}, and (c) \emph{Log Concave} edge weight functions.}
\label{fig:expertaddingrelativecost}
\end{figure*}

\textbf{Impact of Expert Addition on Total Cost}. To evaluate whether the improved modeling capacity from adding more experts in Mix-CVAE leads to better optimization outcomes, we analyze its effect on the total budget required to solve QoSD instances. Figure \ref{fig:expertaddingrelativecost} directly illustrates this downstream impact by plotting the average total budget (in log scale) achieved across a range of graph densities (from sparse to dense) under three distinct edge weight functions: Linear, Quadratic Convex, and Log Concave. We observe a consistent trend: increasing the number of experts from 3 to 9 leads to a progressive reduction in the final budget required to satisfy the QoSD constraint across all edge weight settings. The gap between configurations becomes more pronounced as graph density increases, where solution spaces become more complex and multimodal. Notably, the 9-expert model consistently outperforms the others, achieving the lowest budgets—especially under nonlinear edge weight functions like Log Concave and Quadratic Convex. This confirms that enhanced expressiveness in the generative model enables better exploration and exploitation of feasible regions in the solution space. 

These results demonstrate a clear benefit of the expert addition strategy: as the number of experts increases, Mix-CVAE becomes more capable of capturing diverse solution modes, which directly improves solution quality. The budget savings are especially significant under higher density graphs, where the optimization problem is inherently harder and requires stronger modeling capacity to discover high-quality solutions.





\subsection{Performance under Non-Linear Edge Weight Functions} \label{app:nonlinear_performance}

The main paper presents only performance comparisons on real-world networks under the linear weight function, where the QoSD problem can be reformulated as an ILP and solved exactly. This section will show performance of \PIMMA in real networks under non-linear edge weight settings where such reformulations are no longer tractable for existing ML-based or optimization baselines. Specifically, Table \ref{tab:performance-convex-real} evaluates the Quadratic Convex setting, where each edge weight follows the form $f_e(x_e) = \aleph(x^2)$. This convex formulation rapidly increases edge cost as budget increases, meaning that even small increases in budget can suffice to surpass the threshold $T$. Fortunately, the latest versions of Gurobi support solving quadratic objectives, hence, an exact solver is still included in this setting for benchmarking. In contrast, Table \ref{tab:performance-concave-real} examines the Log-Concave case, where the edge weight is defined as $f_e(x_e) = \aleph(\ln x)$. This class of function is much slower to grow, so achieving a total cost above $T$ often requires significantly more budget per edge. Importantly, Gurobi and other solvers do not support log-concave objectives natively in mixed-integer formulations, making exact optimization infeasible. Therefore, only approximation methods—Adaptive Trading (AT), Iterative Greedy (IG), Sampling Algorithm (SA)—and our method Hephaestus are reported in Table \ref{tab:performance-concave-real}. 

\begin{table}[htp]
\centering

\label{tab:performance}
\scriptsize 
\setlength{\tabcolsep}{1.3pt} 
\renewcommand{\arraystretch}{0.6}
\begin{tabular}{l *{4}{r} *{4}{r} *{4}{r} *{4}{r}}
\toprule
& \multicolumn{4}{c}{Email} & \multicolumn{4}{c}{Gnutella} & \multicolumn{4}{c}{RoadCA} & \multicolumn{4}{c}{Skitter} \\
\cmidrule(lr){2-5} \cmidrule(lr){6-9} \cmidrule(lr){10-13} \cmidrule(lr){14-17}
Method & 140\% & 180\% & 220\% & 260\% & 140\% & 180\% & 220\% & 260\% & 140\% & 180\% & 220\% & 260\% & 140\% & 180\% & 220\% & 260\% \\
\midrule
Adaptive Trading
        & \textbf{1933} & 4261  & 6647  & 7771  
        & 1558 & 3267  & 5842  & 7518 
        & 7747 & 12080 & 18761 & 29821 
        & 290772 & 714634 & 1637765 & 2743321 \\ 
Iterative Greedy
        & 1978 & 4301  & 6697  & 7825  
        & 1942 & 3635  & 7561  & 9774 
        & 7888 & 13893 & 19369 & 32049 
        & 347454 & 889621 & 1967255 & 3173690 \\ 
Sampling Alg.
        & 3687 & 8835  & 12532 & 16384 
        & 2803 & 4609  & 12670 & 15126 
        & 16271 & 28856 & 55014 & 60716 
        & 521783 & 1580349 & 2478424 & 5741999 \\ 
Exact Solver
        & 1907 & 4198  & 6491  & 7646  
        & 1497 & 3077  & 5499  & 7051 
        & {---}  & {---}   & {---}   & {---}   
        & {---}  & {---}   & {---}   & {---} \\ 
\textbf{\PIMMA}
        & 1944 & \textbf{4256} & \textbf{6602} & \textbf{7711} 
        & \textbf{1529} & \textbf{3126} & \textbf{5631} & \textbf{7327} 
        & \textbf{4955} & \textbf{9253} & \textbf{14107} & \textbf{20493} 
        & \textbf{171157} & \textbf{493507} & \textbf{979447} & \textbf{1716357} \\ 
    \bottomrule
\end{tabular}
\caption{Performance of HEPHAESTUS and baselines on four real datasets at different thresholds $T$. The
best is highlighted in bold excluding exact solution. (Convex)}
\label{tab:performance-convex-real}
\end{table}

\begin{table}[htp]
\centering
\scriptsize 
\setlength{\tabcolsep}{1.3pt}
\renewcommand{\arraystretch}{0.6} 
\begin{tabular}{l *{4}{r} *{4}{r} *{4}{r} *{4}{r}}
\toprule
& \multicolumn{4}{c}{Email} & \multicolumn{4}{c}{Gnutella} & \multicolumn{4}{c}{RoadCA} & \multicolumn{4}{c}{Skitter} \\
\cmidrule(lr){2-5} \cmidrule(lr){6-9} \cmidrule(lr){10-13} \cmidrule(lr){14-17}
Method & 140\% & 180\% & 220\% & 260\% & 140\% & 180\% & 220\% & 260\% & 140\% & 180\% & 220\% & 260\% & 140\% & 180\% & 220\% & 260\% \\
\midrule
Adaptive Trading
    & 3309 & 7595  & 12872  & 14759  
    & 6589 & 10897  & 15268  & 21147 
    & 20218 & 29266 & 46248 & 69457 
    & 1130533 & 4191546 & 6931595 & 9697272 \\ 
Iterative Greedy
    & 3473 & 7706  & 13095  & 16042  
    & 6639 & 11571  & 16092  & 21491 
    & 21301 & 30389 & 48197 & 71405 
    & 1339904 & 5174196 & 8753831 & 10111558 \\ 
Sampling Alg.
    & 4404 & 9589  & 15218 & 18336 
    & 7980 & 13957  & 19632 & 23286 
    & 38427 & 68553 & 95579 & 128118 
    & 5621568 & 7373689 & 10220892 & 15464201 \\ 
\textbf{\PIMMA}
    & \textbf{3188} & \textbf{7320} & \textbf{10942} & \textbf{12318} 
    & \textbf{6403} & \textbf{10754} & \textbf{13949} & \textbf{19110} 
    & \textbf{17299} & \textbf{27242} & \textbf{42311} & \textbf{64796} 
    & \textbf{424188} & \textbf{1115779} & \textbf{2608813} & \textbf{7064231} \\ 
    \bottomrule
\end{tabular}
\caption{Performance (Total Attack Budget) of HEPHAESTUS and approximation baselines on four real datasets under Log Concave edge weights. Lower is better. Best approximation/HEPHAESTUS result is bolded. Exact Solver fails to run.}
\label{tab:performance-concave-real} %
\end{table}

\textbf{Performance under Quadratic Convex $f_e$:} In Table \ref{tab:performance-convex-real} (Quadratic Convex), Hephaestus consistently achieves the lowest total cost among all approximation methods. On smaller graphs like Email and Gnutella, its performance is often close to the exact solver when available, while on larger graphs like RoadCA and Skitter, where the exact solver fails to run—Hephaestus still maintains strongest performance compared with remaining available baselines, highlighting its scalability and robustness to nonlinear cost functions.

\textbf{Performance under Log Concave $f_e$:} Table \ref{tab:performance-concave-real} (Log Concave) further highlights Hephaestus’ scalability and generality. Despite there is no longer a ground-truth solver, Hephaestus consistently outperforms AT, IG, and SA across all thresholds and network sizes. The gap in total budget becomes particularly significant for larger graphs and higher thresholds, as approximation methods struggle to effectively navigate the slow growth dynamics of the log-concave function. By contrast, Hephaestus’s generative-refinement framework allows it to adapt to the more complex shape of the cost landscape.

These findings validate that Hephaestus remains highly effective in settings where other solvers or learning-based methods are inapplicable. Moreover, the results underscore the flexibility of Hephaestus across diverse graph structures, ranging from small, sparse email networks to massive Internet-scale topologies like Skitter. Moreover, unlike many existing learning-based methods—such as DIFFILO or Predict-and-Search—which are designed specifically for linear cost functions, \PIMMA is built to work across a wide range of cost models. This is possible because it separates the learning of solution structures (e.g., which paths to attack) from the specific form of $f_e$ (e.g., linear, quadratic, log-concave). In fact, owing to its generative modeling approach in latent space, Hephaestus does not rely on any particular formula for how edge costs increase, hence, can handle different types of cost functions without needing to change its architecture. This flexibility makes Hephaestus more practical for real-world use, where both the structure of the network and how costs behave can vary significantly from one case to another.

\subsection{Soundness of the Reward Function}
\label{app:Soundness of the Reward Function}

Recall that the reward function used in the Refine phase of \PIMMA is defined as follows:

\begin{equation}
 \mathcal{R}\left(\mathbf{x}_{i+1}\right)
=
\underbrace{\digamma\left(G, \mathcal{K}, \hat{\mathbf{x}}_{i+1}\right)}_{\text{(i) Smooth feasibility term}}
-
\underbrace{\varkappa \cdot \log \left(1+\left\|\overline{\mathbf{x}}_{i+1}\right\|_1\right)}_{\text{(ii) Soft cost penalty term}}   
\end{equation}

\begin{figure*}[htp]
    \centering
    \captionsetup[subfigure]{skip=0.1pt}  
    \begin{subfigure}[t]{0.32\linewidth}
        \centering
        \includegraphics[width=\linewidth]{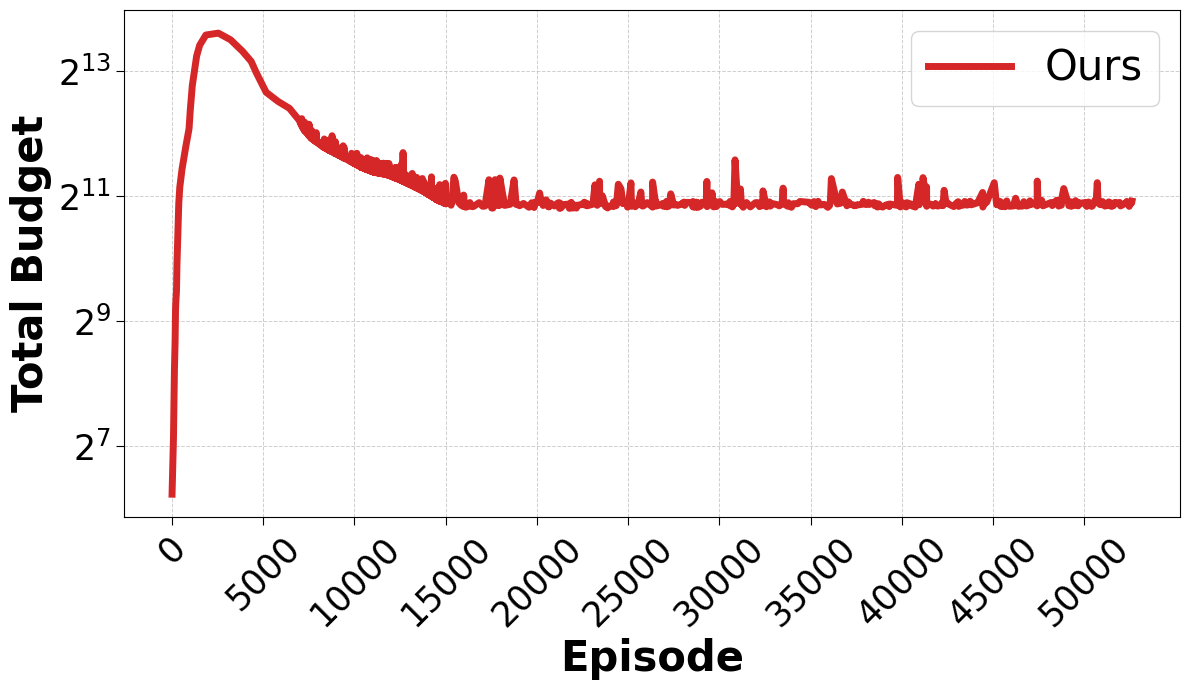}
        \caption{Graph 1000}
    \end{subfigure}
    \hfill
    \begin{subfigure}[t]{0.32\linewidth}
        \centering
        \includegraphics[width=\linewidth]{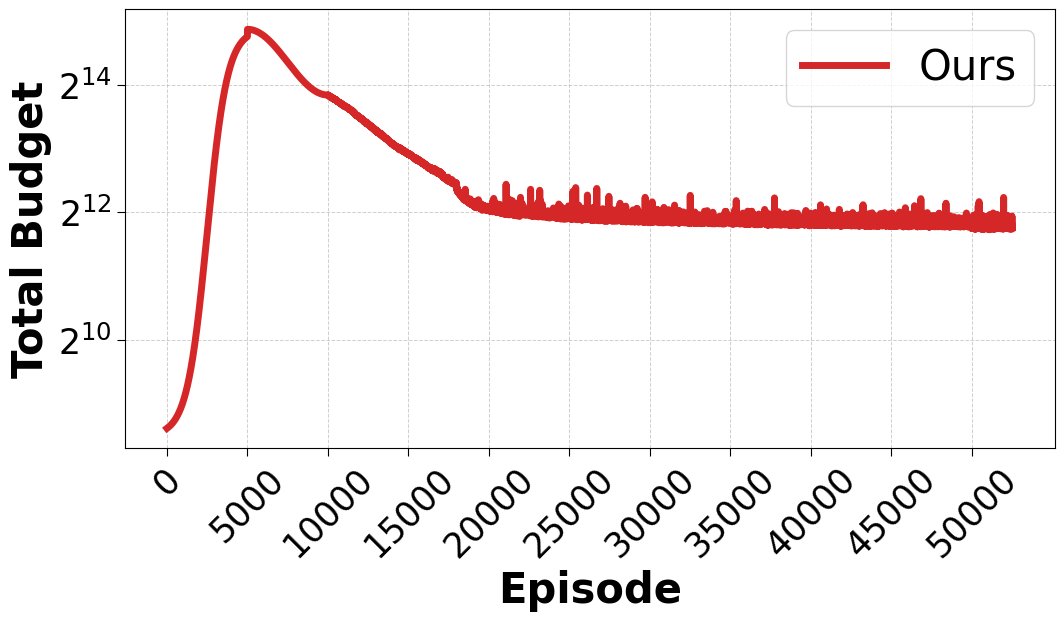}
        \caption{Graph 3000}
    \end{subfigure}
    \hfill
    \begin{subfigure}[t]{0.32\linewidth}
        \centering
        \includegraphics[width=\linewidth]{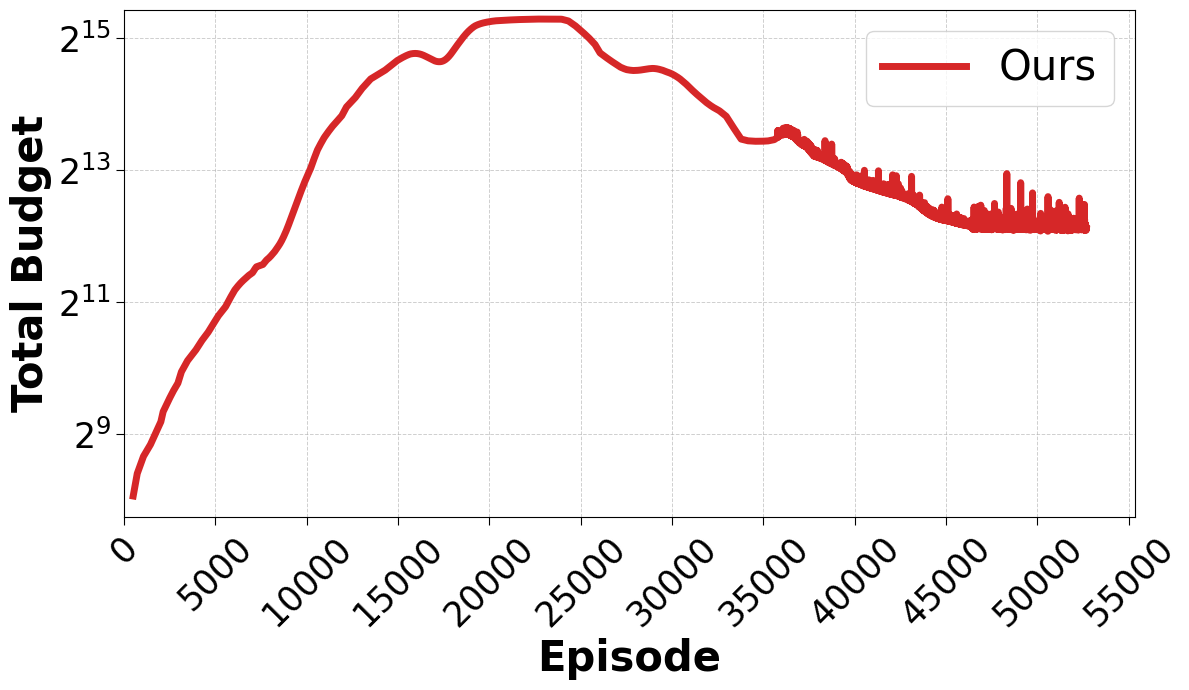}
        \caption{Graph 5000}
    \end{subfigure}
    \vspace{-5pt} 
    \caption{Evolution of Total Budget during the Refine phase (RL agent training)}
\label{fig:reward_1}
\end{figure*}

This reward formulation encourages policy $\pi$ to generate solutions that are both feasible (increasing the sigmoid-based term for all critical pairs $(s,t)$) and efficient (penalizing a large total budget using the second term). The key idea is to guide the policy to find a balance between maximizing feasibility and minimizing the total cost.

\begin{figure*}[htp]
    \centering
    \captionsetup[subfigure]{skip=0.1pt}  
    \begin{subfigure}[t]{0.32\linewidth}
        \centering
        \includegraphics[width=\linewidth]{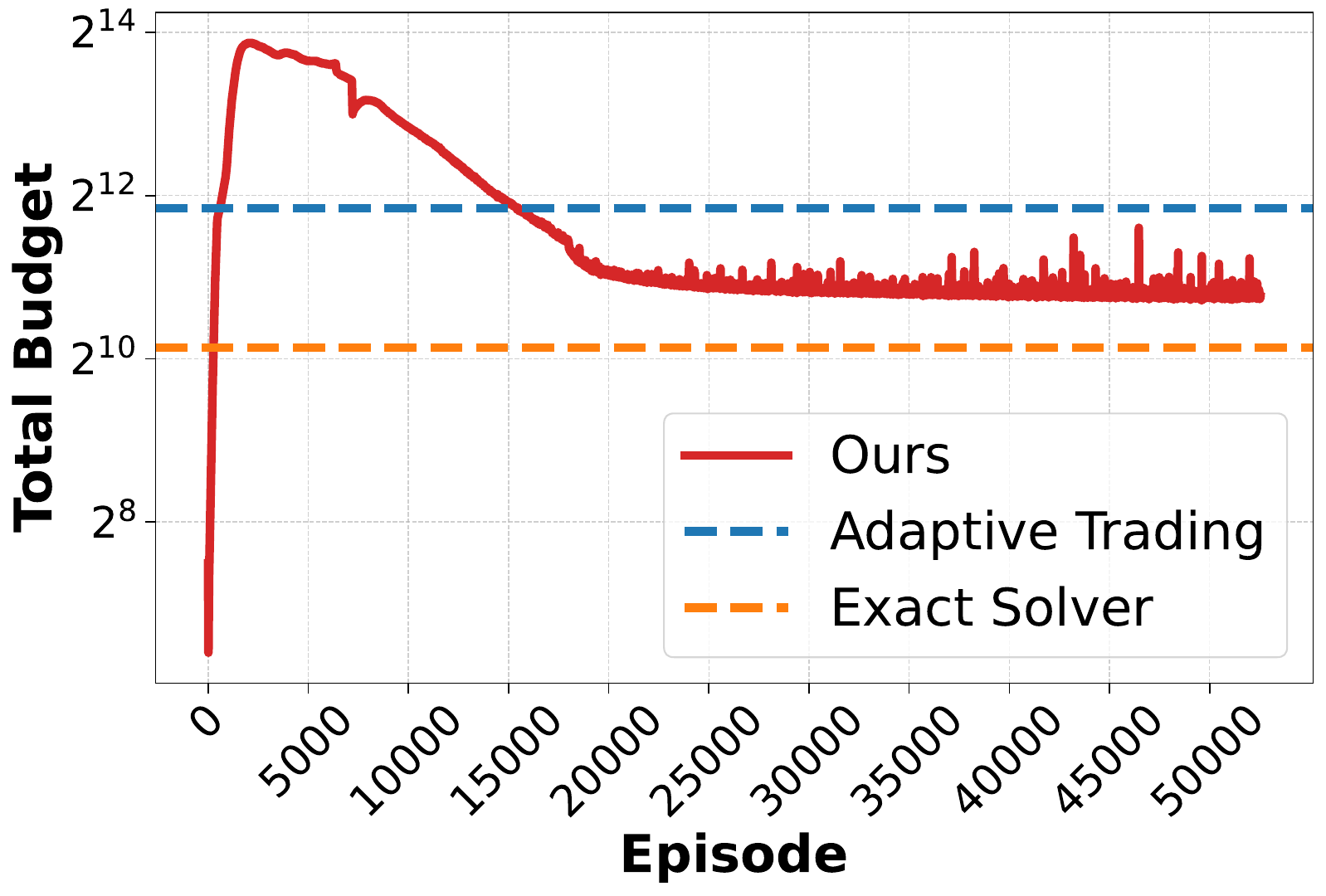}
        \caption{Linear}
    \end{subfigure}
    \hfill
    \begin{subfigure}[t]{0.32\linewidth}
        \centering
        \includegraphics[width=\linewidth]{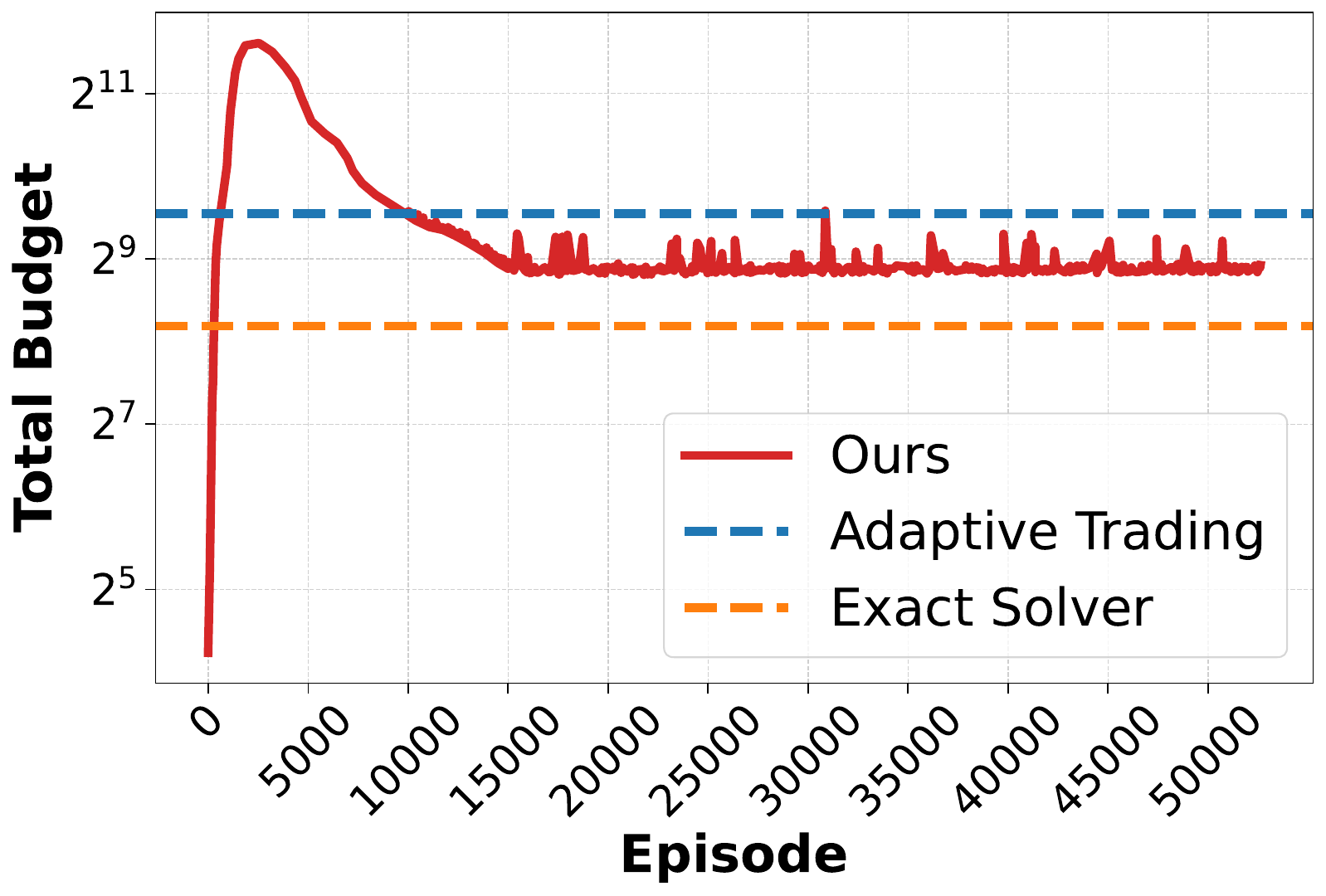}
        \caption{Quadratic Convex}
    \end{subfigure}
    \hfill
    \begin{subfigure}[t]{0.32\linewidth}
        \centering
        \includegraphics[width=\linewidth]{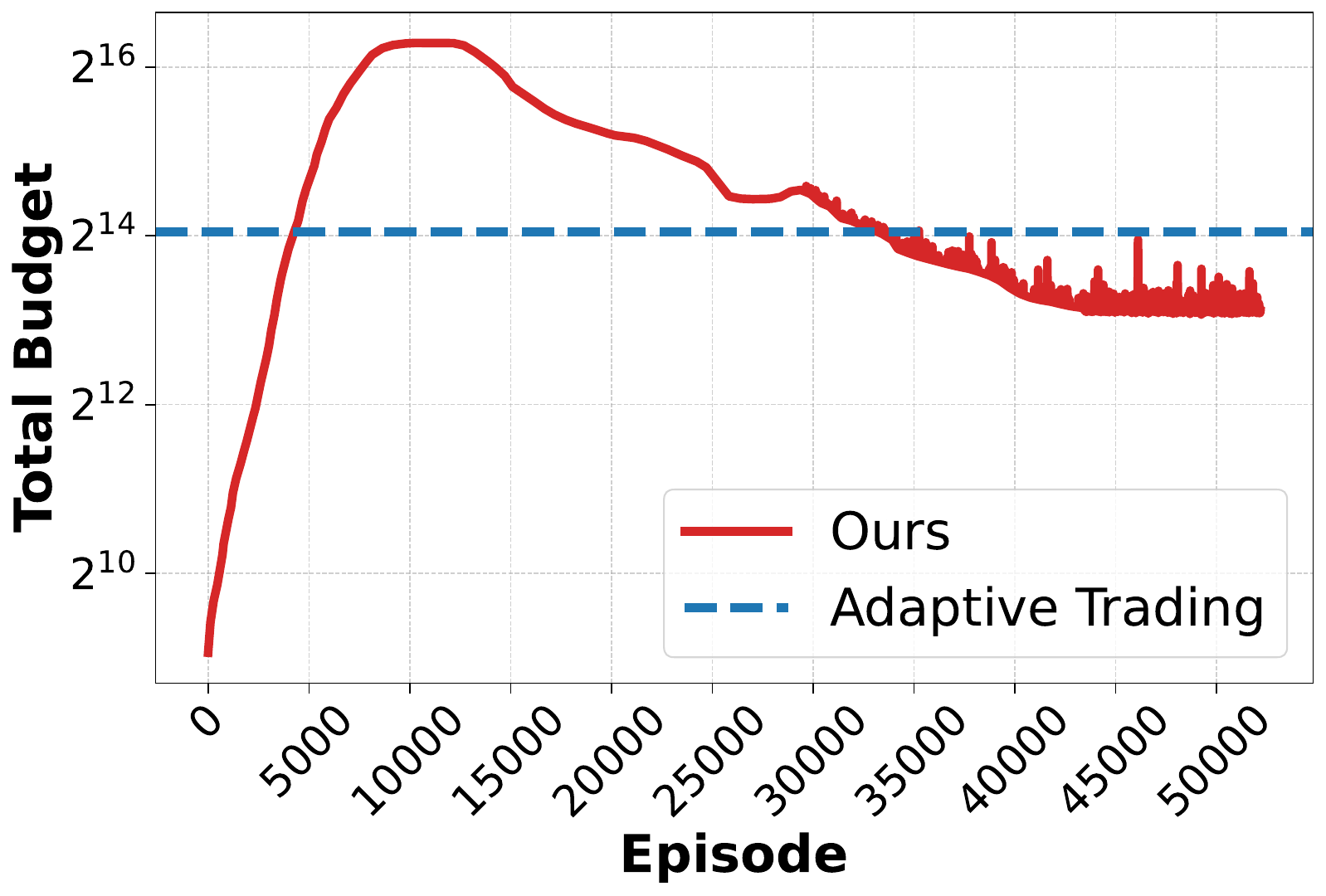}
        \caption{Log Concave}
    \end{subfigure}
    \vspace{-5pt} 
    \caption{Evolution of Total Budget during the Refine phase (RL agent training)}
\label{fig:reward_1}
\end{figure*}

\begin{figure*}[htp]
    \centering
    \captionsetup[subfigure]{skip=0.1pt}  
    \begin{subfigure}[t]{0.32\linewidth}
        \centering
        \includegraphics[width=\linewidth]{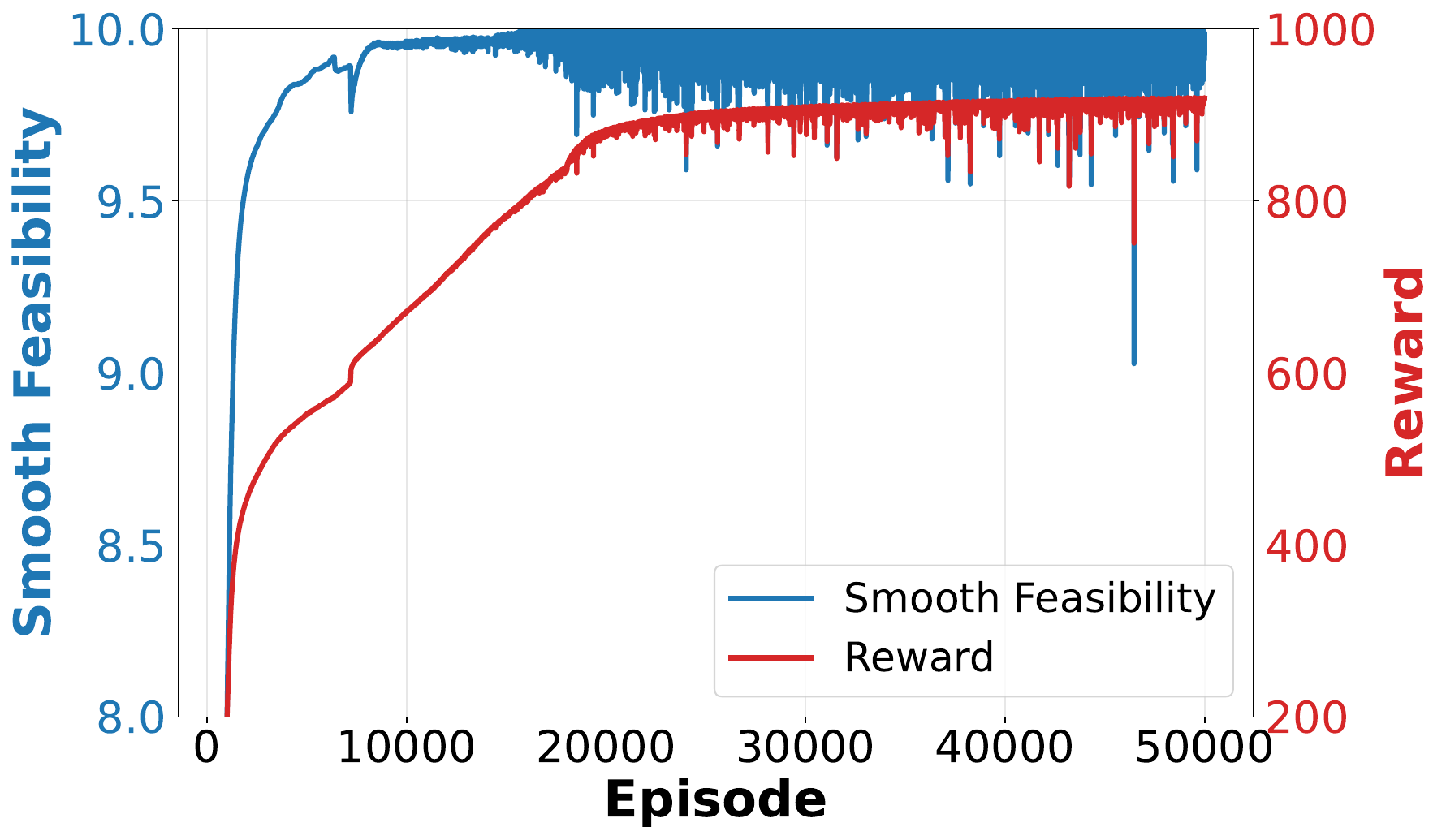}
        \caption{Linear}
    \end{subfigure}
    \hfill
    \begin{subfigure}[t]{0.32\linewidth}
        \centering
        \includegraphics[width=\linewidth]{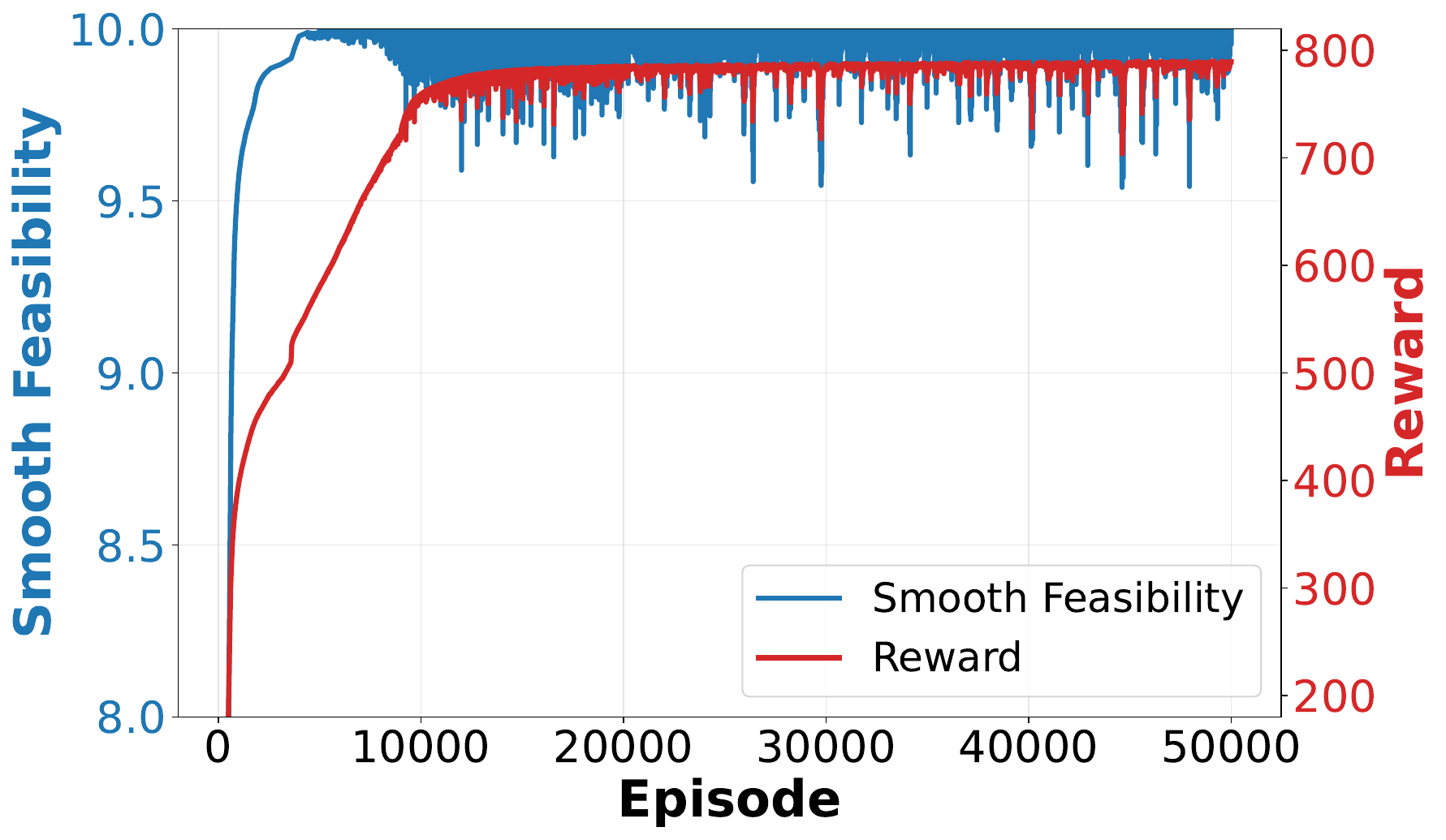}
        \caption{Quadratic Convex}
    \end{subfigure}
    \hfill
    \begin{subfigure}[t]{0.32\linewidth}
        \centering
        \includegraphics[width=\linewidth]{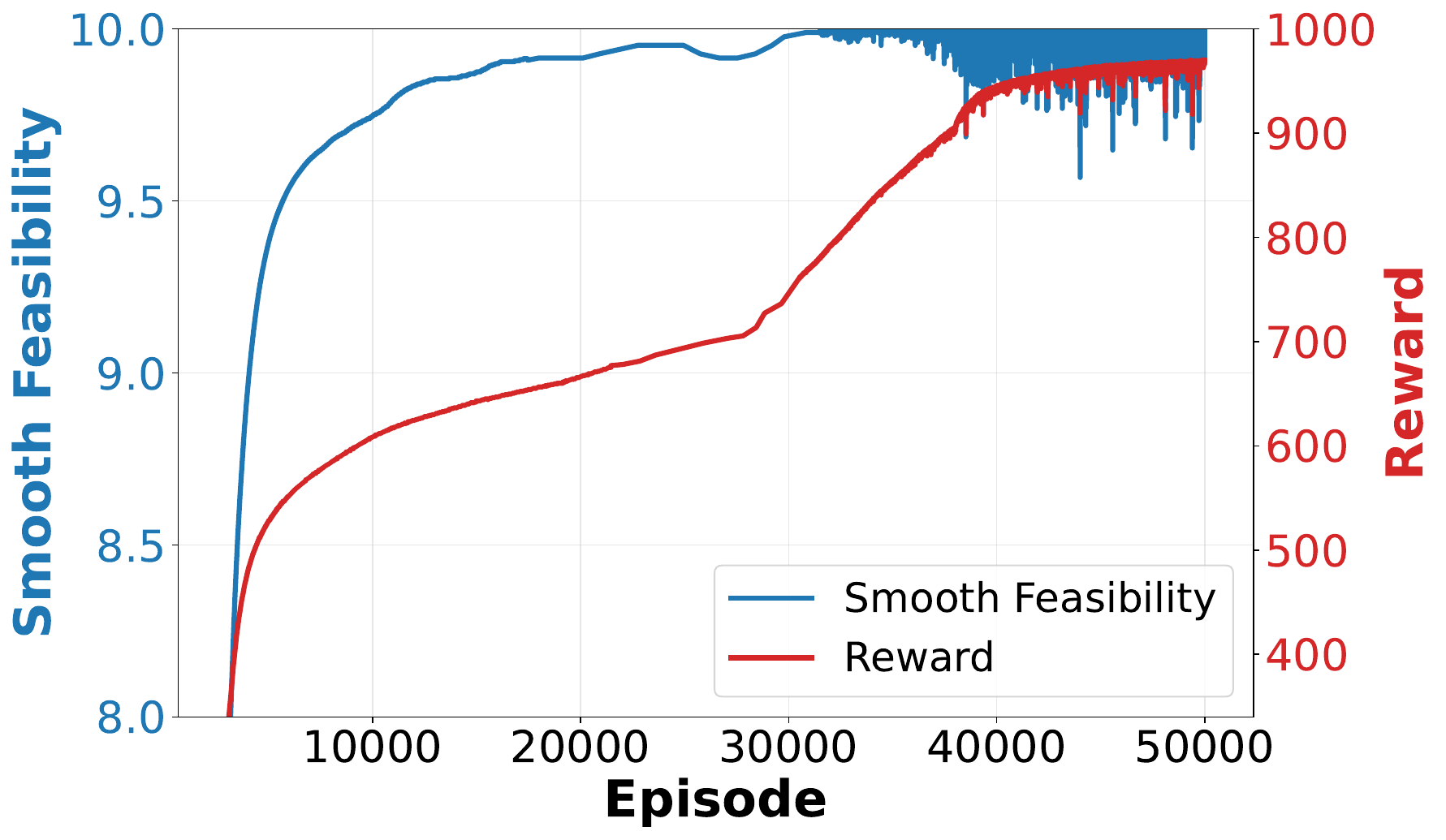}
        \caption{Log Concave}
    \end{subfigure}
    \vspace{-5pt}  
    \caption{Evolution of Feasible Solution quality (blue, left axis) and Agent Reward (red, right axis) over training episodes.}
\label{fig:reward_2}
\end{figure*}

In Figure \ref{fig:reward_1}, we show how the total budget evolves over training episodes across three different edge weight settings: Linear, Quadratic Convex, and Log-Concave. A noteworthy observation across all plots is that the budget initially increases sharply during the early training phase, especially visible in episodes 0–3000. This behavior can be explained as: at this stage, the RL agent has not yet learned effective policies and instead prioritizes feasibility, attempting to push more critical paths above the required threshold $T$, often by over-allocating budget. This is aligned with the feasibility curve in Figure \ref{fig:reward_2} (a–c), where the smooth feasibility skyrockets during the same initial phase, confirming that the agent is successfully raising path costs. As training continues and the feasibility nears saturation (close to a value of 10, meaning that the total cost of the shortest paths for all 10 target pairs exceeds the threshold $T$), we begin to observe a transition: the agent shifts focus toward reducing the overall budget while preserving feasibility. This can be seen in Figure \ref{fig:reward_1}(a), where from episode 3000 onwards, the total budget starts to steadily decline. Correspondingly, Figure \ref{fig:reward_2}(a) shows that feasibility remains stable, indicating that the agent is successfully optimizing the second term of the reward without sacrificing the first. However, a particularly interesting phenomenon occurs around episode 7000. As the agent becomes increasingly focused on minimizing the cost penalty, it temporarily neglects feasibility. This is reflected in Figure \ref{fig:reward_2}(a) as a sharp dip in smooth feasibility around that episode, suggesting some paths have dropped below the threshold. Simultaneously, Figure \ref{fig:reward_1}(a) shows a sudden plunge in total budget, confirming that the policy has aggressively reduced edge weights to cut cost. Eventually, the agent corrects this behavior. Realizing that dropping feasibility lowers the total reward, the policy re-adjusts — it selectively increases the budget on critical edges to recover feasibility, while continuing to reduce less impactful edges. This adaptive dynamic is exactly what the reward function is designed to elicit: the policy explores aggressively, corrects when needed, and eventually converges to a high-feasibility, low-cost solution.

This self-regulating behavior is consistent across Quadratic Convex (Figure 10b/11b) and Log-Concave (Figure 10c/11c) settings, although the trajectory is slightly more gradual in the latter due to the slower cost escalation of log-concave functions. The reward curves (in red, Figure 11) grow monotonically across all cases, confirming that the policy is indeed optimizing the reward function effectively. Overall, these experiments confirm the soundness of the reward design. It provides a useful training signal that balances feasibility and efficiency, and encourages intelligent exploration–exploitation dynamics during learning.


\subsection{Relative optimal gap convergence}
\label{app:Relative optimal gap convergence}

\begin{figure*}[htp]
    \centering
    \includegraphics[width=0.9\linewidth]{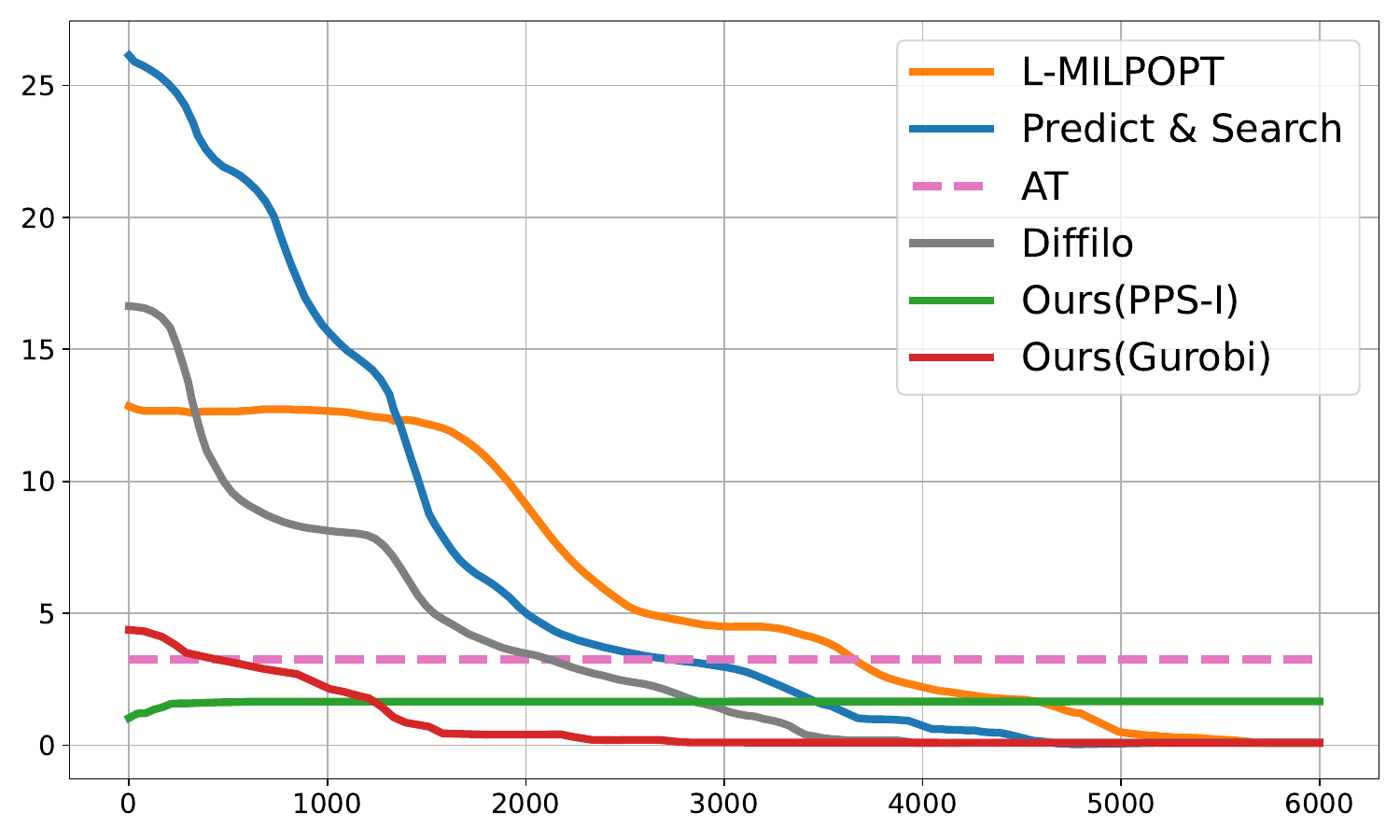}
    \caption{Convergence comparison of the relative optimal gap across \PIMMA and baseline methods under the Linear cost setting. The relative optimal gap measures the percentage deviation from the best-known (optimal) solution, with lower values indicating better performance. Existing ML-based methods (L-MILPOPT, Predict-and-Search, DiffILO) gradually reduce their gaps over time through external solver refinement (e.g., Gurobi), but typically require thousands of seconds. In contrast, \PIMMA with Gurobi (red curve) rapidly converges to zero gap due to its high-quality initial solutions and efficient generative modeling. \PIMMA with PPS-I (green curve), although not exact, achieves low gaps in significantly less time, offering a fast and practical alternative.}
    \label{fig:gap}
\end{figure*}


The goal of this experiment is to evaluate how effectively and efficiently each method approaches the optimal solution in terms of total budget, using relative optimal gap as the key metric. This gap quantifies the percentage deviation from the best-known solution, where lower values indicate better performance and zero denotes exact optimality. As observed in the Figure \ref{fig:gap}, existing learning-based baselines, such as L-MILPOPT, Predict-and-Search, and DiffILO—require post-processing using an exact solver like Gurobi to reduce their optimality gap. While this refinement phase eventually helps those methods converge, it often takes several thousand seconds and incurs substantial computational cost. In contrast, \PIMMA with Gurobi refinement (red curve) starts with high-quality initial solutions, thanks to the expressiveness of Mix-CVAE, and rapidly converges to zero gap much faster. This demonstrates the synergy between generative modeling and reinforcement learning in guiding the solution space effectively. Meanwhile, Hephaestus with PPS-I (green curve) does not reach exact optimality, as PPS-I is an approximation algorithm that does not enforce all path constraints exhaustively. However, it consistently achieves low optimality gaps in a fraction of the time compared to exact methods, offering an effective trade-off between solution quality and efficiency. An interesting observation in the curve is the initial rise in total cost during the first few seconds, after which the value quickly plateaus. This behavior comes from the fact that the initial solution generated by the RL policy $\pi$—may not fully satisfy all feasibility constraints. PPS-I then incrementally increases edge budgets based on marginal benefit-to-cost analysis, just enough to achieve feasibility, and terminates once all critical pairs are satisfied. Thus, this experiment underscores that \PIMMA is capable of adapting to different computational needs. It delivers fast approximate results with PPS-I, or provably optimal solutions with Gurobi refinement—while consistently outperforming existing baselines in both convergence quality and speed.

\subsection{Hephaestus with different feasibility refinement} \label{app:HEPHAESTUS_pps_at}
\begin{table}[htp]
    \centering
    \setlength{\tabcolsep}{4pt}  
    \small  
    \caption{Performance comparison of \PIMMA (Hephaestus) under different refinement strategies (Initial Prediction, PPS-I, and Gurobi) across varying graph densities and benchmarked against baselines. The table reports total budget (lower is better), feasibility rate (higher is better), and running time (in seconds). Results demonstrate that \PIMMA consistently achieves optimal or near-optimal solutions with significantly lower computation time using PPS-I, while Gurobi refinement reaches exact feasibility at higher computational cost.}
    \label{tab:pps_synthetic_final_split}
    \begin{tabular}{ll S[table-format=5.0] S[table-format=3.2,table-space-text-post={\%}] S[table-format=5.2]}
        \toprule
        Graph & Method & {Total Budget $\downarrow$} & {Feasibility Rate $\uparrow$} & {Running Time (s) $\downarrow$} \\
        Density ($p$) & & & & \\
        \midrule
        \multirow{13}{*}{0.2} 
                              & AT & 163 & 100.00 & 44.10 \\
                              & IG & 189 & 100.00 & 43.23 \\
                              & SA & 240 & 100.00 & 508.49 \\
                              & L-MILPOPT (Initial Pred.) & 462 & 95.11 & 15.22 \\ 
                              & L-MILPOPT (Gurobi) & 119 & 100.00 & 240.96 \\      
                              & P\&S (Initial Pred.) & 411 & 95.52 & 34.57 \\
                              & P\&S (Gurobi) & 119 & 100.00 & 716.31 \\
                              & DIFFILO (Initial Pred.) & 386 & 92.78 & 28.31 \\
                              & DIFFILO (Gurobi) & 119 & 100.00 & 412.64 \\
                              & \PIMMA (Initial Pred.) & 119 & 100.00 & 12.98 \\
                              & \textbf{\PIMMA (PPS-I)} & 119 & 100.00 & 13.72 \\
                              & \PIMMA (Gurobi) & 119 & 100.00 & 198.22 \\
                              & Exact Solver & 119 & 100.00 & 921.18 \\
        \midrule
        \multirow{13}{*}{0.5} 
                              & AT & 856 & 100.00 & 147.38 \\
                              & IG & 992 & 100.00 & 144.46 \\
                              & SA & 1260 & 100.00 & 2409.83 \\
                              & L-MILPOPT (Initial Pred.) & 2735 & 84.17 & 37.56 \\ 
                              & L-MILPOPT (Gurobi) & 639 & 100.00 & 1503.01 \\      
                              & P\&S (Initial Pred.) & 2158 & 82.91 & 86.43 \\
                              & P\&S (Gurobi) & 630 & 100.00 & 4476.94 \\
                              & DIFFILO (Initial Pred.) & 2027 & 88.36 & 70.78 \\
                              & DIFFILO (Gurobi) & 630 & 100.00 & 2579.00 \\
                              & \PIMMA (Initial Pred.) & 631 & 99.12 & 21.95 \\
                              & \textbf{\PIMMA (PPS-I)} & 637 & 100.00 & 24.00 \\
                              & \PIMMA (Gurobi) & 630 & 100.00 & 1238.88 \\
                              & Exact Solver & 630 & 100.00 & 5757.38 \\
        \midrule
        \multirow{13}{*}{1.0} 
                              & AT & 2288 & 100.00 & 298.00 \\
                              & IG & 2471 & 100.00 & 350.33 \\
                              & SA & 3238 & 100.00 & 8113.90 \\
                              & L-MILPOPT (Initial Pred.) & 8573 & 78.12 & 175.08 \\ 
                              & L-MILPOPT (Gurobi) & 2237 & 100.00 & 5070.23 \\      
                              & P\&S (Initial Pred.) & 8549 & 78.36 & 172.85 \\
                              & P\&S (Gurobi) & 2204 & 100.00 & 4836.18 \\
                              & DIFFILO (Initial Pred.) & 7095 & 83.01 & 141.55 \\
                              & DIFFILO (Gurobi) & 2204 & 100.00 & 3919.23 \\
                              & \PIMMA (Initial Pred.) & 2215 & 98.18 & 28.77 \\
                              & \textbf{\PIMMA (PPS-I)} & 2236 & 100.00 & 37.12 \\
                              & \PIMMA (Gurobi) & 2204 & 100.00 & 2218.16 \\
                              & Exact Solver & 2204 & 100.00 & 8127.64 \\
        \bottomrule
    \end{tabular}
    
\end{table}

In this section,  we want to show performance of \PIMMA under three refinement modes: initial prediction only (no refinement), refinement via PPS-Inference (PPS-I), and full Gurobi-based refinement. These configurations are compared against classical approximation baselines (AT, IG, SA), several learning-based approaches (L-MILPOPT, Predict-and-Search, DIFFILO), and the exact solver, across synthetic Erdős–Rényi graphs with increasing edge density ($p = 0.2, 0.5, 1.0$). As illustrated in Table \ref{tab:pps_synthetic_final_split}, across all densities, \PIMMA’s initial predictions already yield strong performance, achieving near-optimal total cost and high feasibility—often outperforming other ML-based methods even before refinement. When PPS-I is applied, \PIMMA consistently reaches 100\% feasibility with only minor increases in total cost. For instance, at $p = 1.0$, the PPS-I solution attains 2236 total budget, just 1.45\% above the exact optimum (2204), while completing in only 37 seconds—over 200× faster than the full exact solver. This efficiency comes from the fact that the latent generative model already places solutions very close to the feasible region. PPS-I only needs to make slight local adjustments (i.e., increasing budgets on a few critical edges) to push all violated shortest-path constraints above threshold $T$. Thus, PPS-I acts as a lightweight yet effective refinement method, ideal for scenarios where runtime is critical but small optimality gaps are acceptable. When exact optimality is required, combining \PIMMA with Gurobi refinement yields the best possible solution (zero optimality gap) in all cases. Importantly, due to the strong initialization from Mix-CVAE and RL, Gurobi converges significantly faster than from scratch—demonstrating the value of learning-based warm starts even for traditional solvers. Compared to L-MILPOPT, Predict-and-Search, and DIFFILO, which depend heavily on post-hoc Gurobi refinement and suffer from lower feasibility in their initial outputs, \PIMMA stands out by producing feasible or near-feasible solutions immediately. This results in better cost-feasibility tradeoffs and reduced reliance on expensive solver calls.

\end{document}